\documentclass[times,twocolumn,final,nopreprintline]{elsarticle}

\usepackage{amsmath,amssymb,amsfonts}
\usepackage{graphicx}
\usepackage{booktabs,makecell,longtable}
\usepackage{latexsym}
\usepackage{multirow}
\usepackage{url}
\usepackage{float}
\usepackage{csquotes}
\usepackage{enumitem}
\usepackage{hyperref}
\hypersetup{colorlinks=true, linkcolor=blue, citecolor=blue, pdfauthor=author}

\usepackage[dvipsnames]{xcolor}
\usepackage{subcaption}
\usepackage{caption}

\usepackage{tabularx}
\usepackage{tabulary}
\usepackage{bbm}
\usepackage{xtab,afterpage}
\usepackage[ruled,linesnumbered]{algorithm2e}
\SetArgSty{textnormal}
\usepackage{colortbl}

\usepackage{pdflscape}
\usepackage{flafter}
\usepackage{fancyhdr}
\fancypagestyle{lscape}{
\fancyhf{} 
\fancyfoot{}
\fancyfoot{}
 
}

\journal{Medical Image Analysis}

\begin{document}


\begin{frontmatter}

\title{Self-supervised driven consistency training for annotation efficient histopathology image analysis}%

\author[a,b]{Chetan L. Srinidhi \corref{cor1}}
\author[c]{Seung Wook Kim}
\author[d]{Fu-Der Chen}
\author[a,b]{Anne L. Martel}

\cortext[cor1]{Corresponding author: \\
  \textit{E-mail address:} chetan.srinidhi@utoronto.ca (Chetan L. Srinidhi)}

\address[a]{Physical Sciences, Sunnybrook Research Institute, Toronto, Canada}
\address[b]{Department of Medical Biophysics, University of Toronto, Canada}
\address[c]{Department of Computer Science, University of Toronto, Canada}
\address[d]{Department of Electrical \& Computer Engineering, University of Toronto, Canada}

\begin{abstract}
Training a neural network with a large labeled dataset is still a dominant paradigm in computational histopathology. However, obtaining such exhaustive manual annotations is often expensive, laborious, and prone to inter and intra-observer variability. While recent self-supervised and semi-supervised methods can alleviate this need by learning unsupervised feature representations, they still struggle to generalize well to downstream tasks when the number of labeled instances is small. In this work, we overcome this challenge by leveraging both \textit{task-agnostic} and \textit{task-specific} unlabeled data based on two novel strategies: i) a self-supervised pretext task that harnesses the underlying multi-resolution contextual cues in histology whole-slide images to learn a powerful supervisory signal for unsupervised representation learning; ii) a new \textit{teacher-student} semi-supervised consistency paradigm that learns to effectively transfer the pretrained representations to downstream tasks based on prediction consistency with the task-specific unlabeled data. 

We carry out extensive validation experiments on three histopathology benchmark datasets across two classification and one regression based tasks, \textit{i.e.,} tumor metastasis detection, tissue type classification, and tumor cellularity quantification. Under limited-label data, the proposed method yields tangible improvements, which is close to or even outperforming other state-of-the-art self-supervised and supervised baselines. Furthermore, we empirically show that the idea of bootstrapping the self-supervised pretrained features is an effective way to improve the task-specific semi-supervised learning on standard benchmarks. Code and pretrained models are made available at: \url{https://github.com/srinidhiPY/SSL\_CR\_Histo}
\end{abstract}

\end{frontmatter}

\section{Introduction}
\label{sec:Introduction}
\vspace{-2mm}
Deep neural network models have achieved tremendous success in obtaining state-of-the-art performance on various histology image analysis tasks ranging from disease grading, cancer classification to outcome prediction \citep{srinidhi2019deep, bera2019artificial, litjens2017survey}. The main success of these methods is attributed to the availability of large-scale open datasets with clean manual annotations.  However, collecting such a large corpus of labeled data is often expensive, laborious, and requires skillful domain expertise, notably in the histopathology domain \citep{Madabhushi2016}. Recently, self-supervised and semi-supervised approaches are becoming increasingly popular to alleviate the annotation burden by leveraging the readily available unlabeled data that can be trained with limited supervision. These methods have recently demonstrated promising results on various computer vision \citep{jing2020, laine2016temporal, sohn2020fixmatch} and medical image analysis tasks \citep{chen2019self, tellez2019neural, li2020transformation}. In this paper, we focus on the self-supervised driven semi-supervised learning paradigm for histopathology image analysis by efficiently exploiting the underlying information present in unlabeled data, both in \textit{task-agnostic} and \textit{task-specific} ways.

The existing plethora of self-supervised learning (SSL) methods can be viewed as defining a surrogate task, i.e., a \textit{pretext} task - which is formulated using only unlabeled examples and which requires a high-level semantic understanding of the image to solve the pretext tasks \citep{jing2020}. The neural network model trained to solve this pretext task often learns useful visual representations that can be transferred to any \textit{downstream} task to solve the task-specific problem. On the other hand, another important stream of work is based on semi-supervised learning (SmSL), which seeks to learn from both labeled and unlabeled examples with limited manual annotations \citep{Chapelle2010}. Among SmSL methods, the most recent and popular stream of approaches are based on \textit{consistency} regularization \citep{laine2016temporal, sajjadi2016regularization} and \textit{pseudo labeling} \citep{lee2013pseudo, sohn2020fixmatch} techniques. The consistency based methods aim to constrain network predictions to be invariant to input or model weight perturbations, such as image augmentations \citep{xie2019unsupervised}, network dropout \citep{srivastava2014dropout} and stochastic depth \citep{huang2016deep}. The main idea is that the model should predict similar labels for both the input image and its perturbed (augmented) version of the same image. Alternatively, pseudo labeling based methods impute artificial (pseudo labels) for the unlabeled data obtained from the model class predictions, which is trained using labeled data alone \citep{sohn2020fixmatch}. The success of these SmSL approaches is attributed to the fact that these models implicitly learn to fit decision boundaries by grouping similar images to share similar labels, forming high-density clusters in the input feature space. 
 
Despite significant advancements among SSL and SmSL approaches, they still suffer from some major limitations. Several SSL methods assume that optimizing the pretext objective task will invariably yield suitable downstream representations for the target task. However, many recent studies \citep{zoph2020rethinking, yan2020clusterfit, goyal2019scaling} have shown that SSL methods overfit to the pretraining objective and may not generalize well to the downstream tasks. In contrast, methods based on SmSL approaches generally struggle to learn effectively when the number of labeled instances are scarce and also noisy \citep{rebuffi2020semi}. This is a typical scenario in histopathology, where the number of manually labeled annotations is small and are often noisy  \citep{shi2020graph}. Besides, when the ratio of labeled and unlabeled samples is highly imbalanced, models trained solely based on consistency strategy have very low accuracy and higher entropy, which prevents them from achieving high-confidence scores (i.e., pseudo labels) on unlabeled data \citep{kim2020distribution}. To address these shortcomings, several recent studies \citep{zhai2019s4l, rebuffi2020semi} have explored the feasibility of integrating the merits of both SSL and SmSL approaches to further enhance the performance on downstream tasks by efficiently exploiting the limited labeled data together with abundant unlabeled data. These approaches first aim to initialize a good latent representation of the data using self-supervised pretraining, followed by reinitializing these features on downstream tasks in a task-specific semi-supervised manner. This idea of bootstrapping features trained via SSL algorithm has been shown to improve the SmSL approach by preventing overfitting on the target domain \citep{zhai2019s4l}.

In this paper, we take inspiration from the above observations and propose a novel self-supervised driven semi-supervised learning framework for histopathology image analysis, which leverages the unlabeled data both in a task-agnostic and task-specific manner. To this end, we first present a simple yet effective, self-supervised pretext task, namely, \textbf{\textit{Resolution Sequence Prediction} (RSP)}, which leverages the multi-resolution contextual information present in the pyramidal nature of histology whole-slide images (WSI's). Our design choice is inspired by the way a pathologist searches for cancerous regions in a WSI. Typically, a pathologist zooms in and out into each region, where the tissue is examined at high to low resolution to obtain the details of individual cells and their surroundings. In this work, we show that exploiting such meaningful multi-resolution contextual information provides a powerful surrogate supervisory signal for unsupervised representation learning. Second, we further develop a `\textit{\textbf{teacher-student}}' semi-supervised consistency paradigm by efficiently transferring the self-supervised pretrained representations to downstream tasks. Our approach can be viewed as a knowledge distillation method \citep{hinton2015distilling}, where the self-supervised teacher model learns to generate pseudo labels for the task-specific unlabeled data, which forces the student model to make predictions consistent with the teacher model. We experimentally show that initializing the student model with the SSL pretrained teacher model achieves robustness against noisy input data (i.e., \textit{noise} is injected through various kinds of domain-specific augmentations) and helps learn faster than the teacher in practice. Our whole framework is trained end-to-end to seamlessly integrate the information present in labeled and unlabeled data both in task-specific and task-agnostic ways.     

\vspace{1.5mm}
The major contributions of this paper are:
\begin{itemize}
\vspace{-1mm}
    \item We propose a novel self-supervised pretext task for generating unsupervised visual representations via predicting the resolution sequence ordering in the pyramidal structure of histology WSI.
    \vspace{-1mm}
    \item We present a new `teacher-student' semi-supervised consistency paradigm by efficiently transferring the self-supervised pretrained representations to downstream tasks based on prediction consistency with the task-specific unlabeled data.  
    \vspace{-1mm}
    \item We extensively validate our method on three benchmark datasets across two classification and one regression based histopathology tasks, i.e., tumor metastasis detection, tissue type classification, and tumor cellularity quantification. The proposed self-supervised method, along with consistency training, is shown to improve the performance on all three datasets, especially in the less annotated data regime. 
\end{itemize}

The paper is organized as follows: we first briefly introduce the related works in Section \ref{sec:Related works}. In Section \ref{sec:Methods}, we present the details of our proposed methodology. Datasets and experimental results are described in Section \ref{sec:Experiments}, \ref{sec:Ablation studies}. Finally, we discuss our key findings and limitations of our work in Section \ref{sec:Discussions}, followed by conclusion in Section \ref{sec:Conclusion}.  
\vspace{-2.5mm}

\section{Related works}
\label{sec:Related works}
In this section, for brevity, we review only the recent developments in self-supervised and semi-supervised representation learning literature that are closely relevant to our work.

\subsection{Self-Supervised learning}
\label{ssec:Self-Supervised learning}
Self-supervised learning (SSL) has recently gained momentum in many medical image analysis tasks for reducing the manual annotation burden. These approaches aim to construct different auxiliary pretext tasks, where the supervisory signals are generated within the data. Such self-supervised pretraining of convolutional neural networks (CNN), which are designed to solve these pretext tasks, results in useful visual representations that can be transferred to multiple downstream tasks by fine-tuning with limited labels in a task-specific way. The existing SSL methods in the literature are generally grouped into three categories \citep{jing2020}: context-based, generative-based, and contrastive-based methods. 

Among the \textit{context-based} methods, the design of pretext tasks is generally based on domain-specific knowledge or handcrafted according to the data. Examples with application to medical image analysis include image context restoration \citep{chen2019self}, anatomical position prediction \citep{bai2019self}, 3D distance prediction \citep{spitzer2018improving}, Rubik’s cube recovery \citep{zhuang2019self} and image intrinsic spatial offset prediction \citep{blendowski2019learn}. Many such pretext tasks are often designed based on ad-hoc heuristics, limiting the generalizability of learned representations. An alternative stream of approach is based on \textit{generative} modeling such as VAE \citep{kingma2013auto} or GAN-based models \citep{dumoulin2016adversarially,donahue2016adversarial} which implicitly learn representations by minimizing the reconstruction loss in the pixel space. Compared with the discriminative ones, generative approaches are overly focused on pixel-level details, which limits their ability to model complex structures present in an image. 

Recently, a new class of discriminative methods is proposed based on \textit{contrastive} learning, which learns to enforce similarities in the latent space between similar/dissimilar pairs \citep{he2020momentum, chen2020simple, oord2018representation}. In such methods, the similarity is defined through maximizing mutual information \citep{oord2018representation} or with different data augmentations \citep{chen2020simple}. Notable technique by \cite{chaitanya2020contrastive} extended the contrastive learning approach \citep{chen2020simple} by utilizing domain and problem-specific cues to segment volumetric medical images on three different MRI datasets. Furthermore, \cite{li2020self} proposed a patient feature based softmax embedding to learn multi-modal representations for diagnosing retinal diseases. Finally, the contrastive based approach was also extended to histopathology in \cite{lu2019semi}, by combined attention based multiple instance learning with contrastive predictive coding for weakly supervised histology classification.
\vspace{-5mm}

\subsection{Semi-Supervised learning}
\label{ssec:Semi-Supervised learning}
The existing semi-supervised learning (SmSL) techniques can be broadly categorized into three groups: i) \textit{adversarial training}-based \citep{zhang2017deep, diaz2019retinal, quiros2019pathology}; ii) \textit{graph}-based \citep{shi2020graph, javed2020cellular, aviles2019graphx}; and iii) \textit{consistency}-based \citep{li2020dual, zhou2020deep, li2020transformation, su2019local, liu2020semi} approaches. 

The \textit{adversarial training} based SmSL approaches learn a generative and a discriminate model simultaneously by forcing the discriminator to output class labels instead of estimating the input probability distribution, as in a normal generative adversarial network (GAN). For example, \cite{zhang2017deep} proposed a segmentation and evaluation network, where the segmentation network is encouraged to obtain a segmentation mask for unlabeled images; while, the evaluation network is forced to distinguish the segmentation results with the ground truth by assigning different scores. Meanwhile, \cite{quiros2019pathology} generated pathologically meaningful representations in histopathology to synthesize high-fidelity H\&E breast cancer images resembling that of real ones. On the other hand, \textit{graph} based methods construct a graph that establishes a semantic relationship between its neighbors and utilize the transduction of the graph to assign labels to unlabeled data via label propagation. As a typical example, \cite{aviles2019graphx} proposed a graph-based SSL model for chest X-ray classification, where the pseudo labels for unlabeled data are generated using label propagation. In more recent work, \cite{shi2020graph} utilized a graph-based self-ensembling approach to minimizes the distance between the label prediction and its ensemble target via consistency cost. In general, such self-ensembling based approaches are shown to be robust to noisy labels compared to other graph-based techniques.

The most recent line of work in SmSL is based on \textit{consistency} regularization, which enforces the consistency of predictions to random perturbations such as data augmentations \citep{french2017self}, stochastic regularization \citep{laine2016temporal, sajjadi2016regularization} and adversarial perturbation \citep{miyato2018virtual}. The most notable method by \cite{tarvainen2017mean} proposed the mean-teacher framework that averages the model weights instead of the exponential moving average of the label predictions to enhance the quality of consistency targets. These strategies were recently extended to several medical image analysis tasks. For instance, \cite{li2020transformation} introduced a transformation consistent self-ensembling model for segmenting medical images. Further, several extensions to mean-teacher have also been explored by enforcing prediction consistency in either region-based \citep{zhou2020deep}, relation-based \citep{liu2020semi, su2019local} or cross-domain based \citep{li2020dual}, which is subjected under various domain-specific perturbations.   
\vspace{-2.5mm}

\section{Methods}
\label{sec:Methods}
\vspace{-1.8mm}
\begin{figure*}\centering
\centerline{\includegraphics[width=0.72\linewidth]{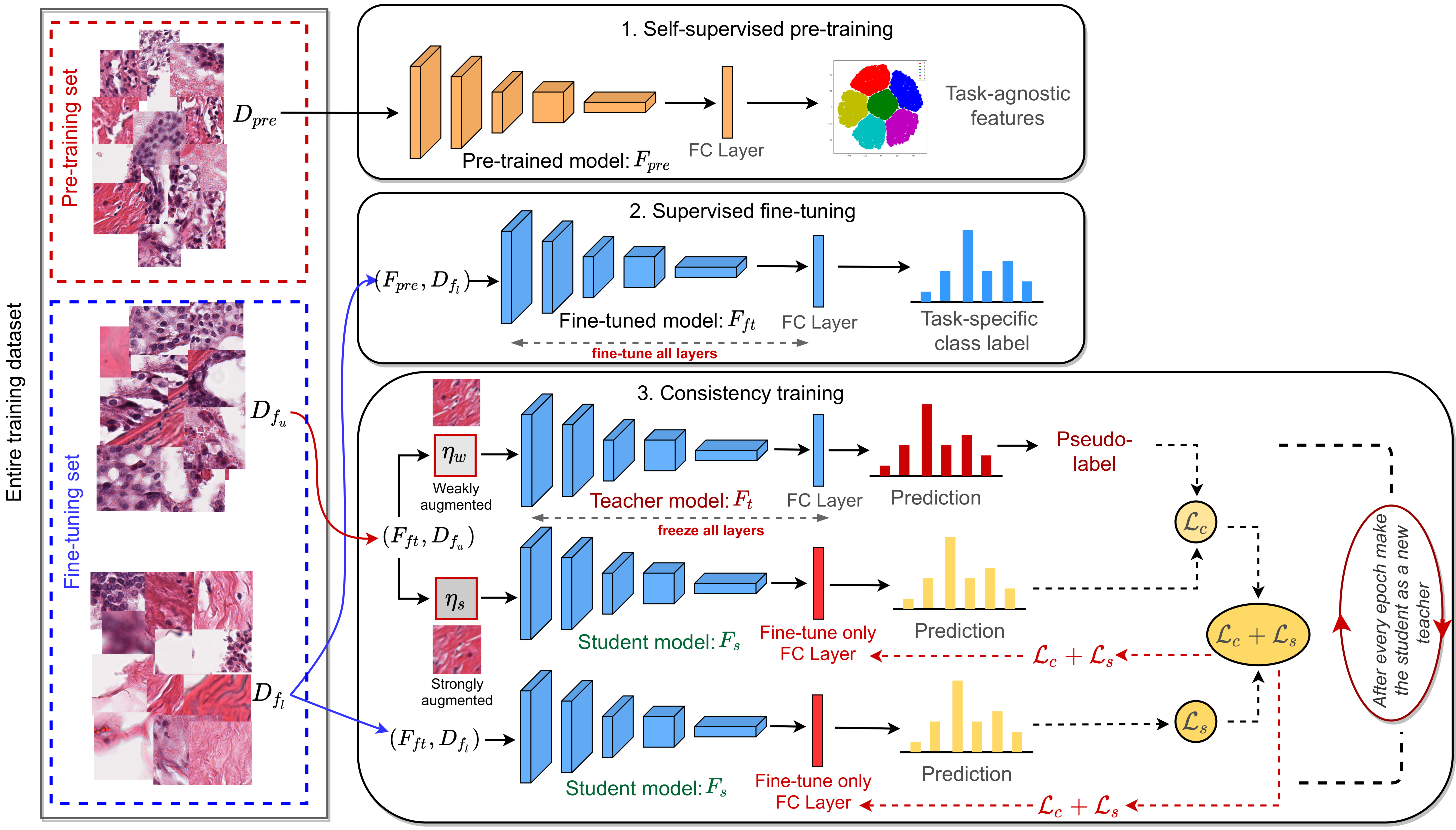}}
\caption{Our self-supervised driven consistency training approach for histopathology image analysis. Our approach consists of three main stages: \textbf{i)} First, we pretrain a self-supervised (SSL) model $F_{pre}$ on the unlabeled set $D_{pre}$ to obtain task-agnostic feature representations; \textbf{ii)} Second, we fine-tune the pretrained SSL model on a limited amount of labeled data $D_{f_l}$ to obtain the task-specific downstream features; and \textbf{iii)} Finally, we further improve the downstream performance on the target task by using both labeled ($D_{f_l}$) and unlabeled ($D_{f_u}$) set in a task-specific semi-supervised manner. Both teacher $F_t$ and student $F_s$ networks are initialized with fine-tuned model $F_{f_t}$ for consistency training on the target task. The main objective is to optimize the student network, which learns to minimize the supervised loss $\mathcal{L}_s$ on labeled set $D_{f_l}$ and consistency loss $\mathcal{L}_c$ on an unlabeled set $D_{f_u}$. The consistency loss is measured between the pseudo labels produced by the teacher network on weakly augmented unlabeled input images with the labels predicted by the student network on a strongly augmented version of the same unlabeled input images. Note: all the above network ($F_{pre}$, $F_{f_t}$, $F_t$, $F_s$) share the same backbone ResNet-18 architecture.}
\label{Fig:Overall_Learning_Scheme}
\vspace{-5mm}
\end{figure*}
An overview of our proposed self-supervised driven consistency training approach is illustrated in Fig. \ref{Fig:Overall_Learning_Scheme}. Our framework consists of three main stages: \textbf{i)} we pretrain a self-supervised model $F_{pre}$ on an unlabeled set $D_{pre}$ to obtain task-agnostic feature representations; \textbf{ii)} we fine-tune the SSL model on a limited amount of labeled data $D_{f_l}$ to obtain the task-specific features; and \textbf{iii)} we further improve the downstream performance on the target task by using both labeled $D_{f_l}$ and unlabeled $D_{f_u}$ data in a task-specific semi-supervised manner. For task-specific semi-supervised training, both teacher $F_t$ and student $F_s$ networks are initialized with the fine-tuned model $F_{f_t}$. The main objective is then to optimize the student network, which learns to minimize the supervised loss on the labeled set ($D_{f_l}$) and consistency loss on the unlabeled set ($D_{f_u}$). During consistency training, the teacher network predicts the pseudo label on a weakly augmented unlabeled image, while the student network tries to match this pseudo label by making its prediction on a strongly augmented version of the same unlabeled image. The details are presented next.

\subsection{Self-supervised pretraining of visual representations}
\label{ssec:Self-supervised pretraining of visual representations}
The goal of self-supervised pretraining is to learn generic visual representations using unlabeled data that can be transferred to many different downstream tasks by fine-tuning with limited labels in a task-specific way. The self-supervised pretraining is performed by solving a pretext task in a task-agnostic manner, where the labels needed to train the network are generated within the data. Before we begin our proposed SSL approach, we first outline some basics of SSL in detail. 

Let us denote the \textbf{pretraining set} as $D_{pre} = \{x_{i}\}_{i=1}^M$ consisting of $M$ \textbf{unlabeled} training samples. The aim of SSL is to train a convolutional neural network (convNet) $F_{pre}$ on this unlabeled set $D_{pre}$ to obtain generalized feature representations in a task-agnostic manner. In histopathology, the input $x_{i} \in \mathbb{R}^{H x W x 3}$ denotes the RGB image patch, sampled from a gigapixel WSI, with height ($H$) and width ($W$); and $y_{i} \in C$ is the class label for $x_{i}$, with $C=\{0, 1\}$ for classification or $\mathbb{R}$ for regression. Our goal is to learn feature embedding $F_{\theta}(.)$ in an unsupervised manner that maps an unlabeled set $F_{\theta}(x_{i})$ to a low-dimensional embedding $F_{\theta}(x_{i}): \mathbb{R}^{H x W x 3} \rightarrow \mathbb{R}^{d}$, with $d$ being the feature dimension, and $F(.)$ denotes the neural network parameterized by $\theta$. Given a set of $M$ training samples $D_{pre} = \{x_{i}\}_{i=1}^M$, the self-supervised pretraining aims to optimize the following objective:
\vspace{-1mm}
\begin{equation}
    L_{pre} = \min_{\theta} \frac{1}{M} \sum_{i=1}^{M} loss ~(x_{i}, p_{i}),
    \label{Eq: SSL}
\end{equation}
where, $p_{i}$ are the pseudo labels generated automatically from the self-supervised pretext tasks. To this end, we propose a context-based resolution sequence prediction as a domain-specific pretext task to learn generic visual representations, which can be transferred to many different downstream tasks in histopathology.   

\subsubsection{Resolution sequence prediction (RSP)}
\label{sssec:Resolution sequence prediction (RSP)}
\begin{figure*}[t]\centering
\centerline{\includegraphics[width=0.70\linewidth]{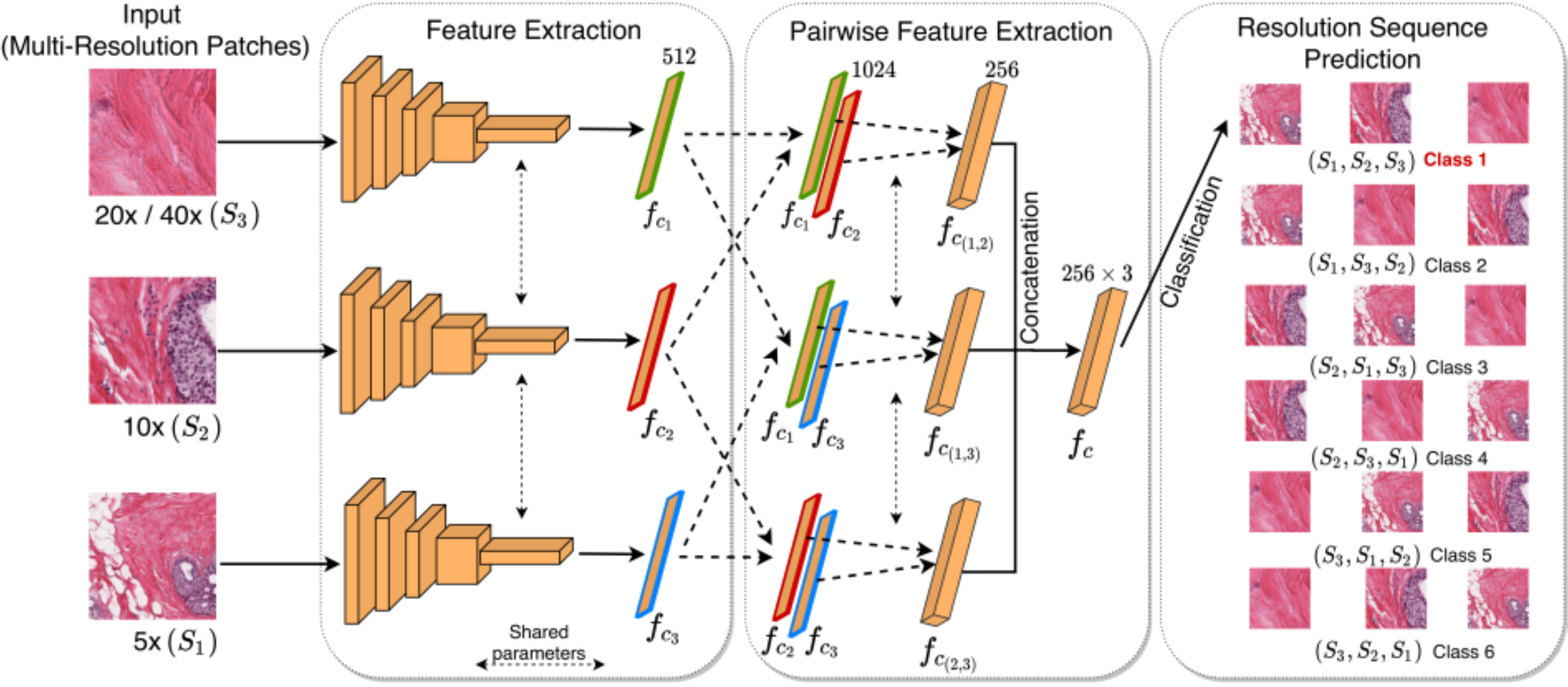}}
\caption{Resolution sequence prediction (RSP) pretext task that we propose for self-supervised representation learning. Given a tuple of three input multi-resolution patches sampled from the 3! = 6 possible permutations of resolution sequences, we train a ConvNet model $F(.)$ to predict the label $y \in \mathbb{R}^{P}$ corresponding to the order of resolution sequence, where, $P \in (1, 2, ..., 6)$. The proposed RSP framework consists of three following stages: \textbf{i)} feature extraction; \textbf{ii)} pair-wise feature extraction; and \textbf{iii)} resolution sequence prediction. Features ($d=512$) for each input multi-resolution patch are extracted from the convNet model, which are later pair-wise concatenated (to obtain $d=1024$), followed by learning a 2-layer multi-layer perceptron (MLP) to obtain pair-wise features ($d=256$). Finally, these pair-wise features are concatenated ($d=768$), which are then fed to another 2-layer MLP with softmax to predict the resolution sequence order $P$.}
\label{Fig:Resolution sequence prediction (RSP)}
\vspace{-4mm}
\end{figure*}
Our self-supervised design choice for the ``Resolution sequence prediction (\textbf{RSP})" task is inspired by how a pathologist examines a WSI during diagnosis for potential abnormalities. Typically, a pathologist switches multiple times between lower magnification levels for context and higher magnification levels for detail. Such multi-resolution multi-field-of-view (FOV) analysis is possible due to the WSI's pyramidal nature, where the multiple downsampled versions of the original image are stored in a pyramidal structure. In this work, we exploit this multi-resolution nature of WSIs by proposing a novel self-supervised pretext task - which learns image representations by training convNets to predict the \textit{order} of all possible sequences of resolution that can be created from the input multi-resolution patches. We argue that solving this resolution prediction task will allow CNN to learn useful visual representations that inherently capture both contextual information (at \textit{lower} magnification) and fine-grained details (at \textit{higher} magnification levels).  

Specifically, we create \textit{6}-tuples of randomly shuffled multi-resolution patches sampled from input WSI. We formulate our resolution sequence prediction task as a \textit{multi-class} classification problem. Formally, we construct a tuple of three concentric multi-resolution RGB image patches $(S_{1}, S_{2}, S_{3}) \in \mathbb{R}^{\mathcal{P} \times \mathcal{P} \times 3}$ extracted at three different magnification levels, such that the spatial resolution of $S_1 << S_2 << S_3$ (measured in $\mu m /px$). By extracting such multiple concentric same size patches ($\mathcal{P} \times \mathcal{P} \times 3$), we ensure that the FOV of one image patch ($S_3$) lies inside the central square region of the other two ($S_2, S_1$) lower magnification patches. A sample set of multi-resolution concentric patches are shown in Fig. \ref{Fig:Resolution sequence prediction (RSP)}. These sets of patches form an input tuple to our self-supervised RSP framework. For brevity, we only consider a tuple of three input patches from a given WSI, for which 3! = 6 possible permutations can be constructed (which is referred to as \textit{resolution sequence ordering}), as illustrated in Fig. \ref{Fig:Resolution sequence prediction (RSP)}. 

To achieve our goal, given an input multi-resolution sequence $x \in (S_{1}, S_{2}, S_{3})^{P} \in \mathbb{R}^{\mathcal{P} \times \mathcal{P} \times 3}$ - among $P$ possible permutations, we aim to train a siamese convNet model \citep{koch2015siamese} $F_{pre}$ to predict the label $y \in \mathbb{R}^{P}$ (i.e., order of resolution sequences over $P \in (1, 2, ..., 6)$ possible classes), which is given by,
\begin{equation}
    F_{pre}(x ~|~ \theta) = \{F^{y}_{pre}(x ~|~ \theta)\}_{y=1}^P, 
\end{equation}
where, $F^{y}_{pre}~(x ~|~ \theta)$ is the predicted class probability for the input sequence $x$, with label $y$, and $\theta$ being the learnable parameter of the model $F(.)$. Therefore, given a set of $M$ training samples from the unlabeled set $D_{pre} = \{x_{i}\}_{i=1}^M$, the convNet model learns to solve the objective function defined in Eq. \ref{Eq: SSL}, by minimizing the categorical cross-entropy (CE) loss defined by,
\begin{equation}
    loss~(x_{i}, y_{i}) = - log ~(F^{y}_{pre}~(x_{i} ~|~ \theta)).
\end{equation}

The proposed RSP framework has \textbf{three} main stages: i) feature extraction; ii) pair-wise feature extraction; and iii) resolution sequence prediction. In the \textit{first} stage, we adopt the siamese based architecture to obtain features for each input multi-resolution patch, where all three network branches share the same parameters. In our work, we adopt the commonly used ResNet-18 model to obtain the features $h_{i} = F(x_{i})$, after the global average pooling layer; where, $h_{i} \in \mathbb{R}^{d}$ is a latent vector of dimension 512. An additional crucial part of self-supervised pretraining is preparing the training data. To prevent the model from picking up on low-level cues and learning only trivial features, we make the sequence prediction task more difficult by applying various geometric transformations to the input data. The details of these geometric transformations are discussed thoroughly in Section \ref{ssec:Implementation details}. In the \textit{second} stage, we perform pair-wise feature extraction on the extracted feature vector $h_{i}$, to capture the intrinsic relationship between the multi-resolution frames. Specifically, we concatenate features of each pair of input patches (i.e., $Concat~(h_1, h_2), Concat~(h_1, h_3), Concat~(h_2, h_3))$ to obtain $h_{ij} \in \mathbb{R}^{d=1024}$ feature vector. Next, we use a 2-layer multi-layer perceptron (MLP) to obtain $z_{i} = g(h_{ij}) = W_{2}\sigma(W_{1}h_{ij})\in\mathbb{R}^{d=256}$; where, $\sigma$ denotes ReLU and the bias is ignored for simplicity. Finally, in the \textit{third} stage, the pair-wise features ($z_i$'s) are concatenated, resulting in $d=256x3$ feature vector. This feature vector is finally fed to another 2-layer MLP with softmax function to predict the order of resolution sequence (i.e., one of 6 possible permutations), as illustrated in Fig. \ref{Fig:Resolution sequence prediction (RSP)}. 

The most common protocol \citep{goyal2019scaling} to evaluate the self-supervised pretrained representations is by training a supervised linear classifier on fixed feature representations, followed by a final evaluation on the test set. In general, most existing SSL approaches utilize the entire task-specific training set $D_f$ to fine-tune the pretrained model on the downstream tasks. However, despite their tremendous success, they still fail to adapt to the new target tasks when the number of labeled training instances is less \citep{zoph2020rethinking, yan2020clusterfit, goyal2019scaling}. A recent study by \cite{zoph2020rethinking} reveals that the value of pretraining diminishes with stronger data augmentation and with the use of more task-specific label data. Further, the authors have shown that the self-supervised pretraining benefits only when fine-tuned with limited amounts of labeled data; whereas, the model performance deteriorates using a more extensive label set.

This raises an important \textit{\textbf{question}} to ``what degree the SSL works and how much amount of labeled data do we need to fine-tune the pretrained SSL model". In this work, we focus on \textit{\textbf{answering}} the above question by performing a set of control experiments by varying the amount of labeled data in both low-data and high-data regimes on three different histopathology datasets. To this end: i) we first supervised fine-tune the pretrained SSL model with varying amounts of labeled data; and ii) we provide an elegant solution based on \textit{teacher-student} consistency paradigm to further improve the downstream performance by exploiting the unlabeled data in a task-specific semi-supervised manner. The details are presented next.

\vspace{-1mm}
\subsection{Supervised fine-tuning}
\label{ssec:Supervised fine-tuning}
The unsupervised learned representations are now transferred to the downstream task using limited labeled data in a task-specific way. We choose to fine-tune all layers in the pretrained network end-to-end to obtain task-specific representations: $F_{ft}(x_i)=W_{ft} F_{pre}(x_i)$; where, $W_{ft}$ are the weights for the task-specific linear layer. Specifically, we add a linear classifier or a regressor onto the end of the pretrained network (i.e., ResNet-18 (until the global average pooling layer) + 2-layer MLP + Fc) and use limited labeled data to fine-tune the entire network. In order to assess how performance of self-supervised pretrained models varies as we increase the fraction of available labeled data $D_{f_{l}} = \{(x_{i}, y_{i})\}_{i=1}^{\alpha \times N}$, we choose to look at the effect of keeping just $\alpha \times N$ of the total $N$ available labels in each dataset, where $\alpha \in  \{10\%, 25\%, 50\%, 100\%\}$. 

\vspace{-0.5mm}
\subsection{Consistency training}
\label{ssec:Consistency traning}
Let us consider the downstream fine-tuning set $D_f$ in each experiment consists of a total of $N$ samples, out of which $\alpha \times N$ are retained as \textbf{labeled} inputs $D_{f_{l}} = \{(x_{i}, y_{i})\}_{i=1}^{\alpha \times N}$, where $\alpha \in  \{10\%, 25\%, 50\%, 100\%\}$. In order to maximise the number of unlabeled instances available for training, we use all $N$ samples as \textbf{unlabeled} inputs $D_{f_u} = \{x_{i}\}_{i=1}^{N}$, i.e., we include all labeled instances $D_{f_{l}}$ as a part of unlabeled set - discarding their labels, when constructing $D_{f_{u}}$. 

\vspace{1mm}
Our teacher-student consistency training (shown in Fig.\ref{Fig:Overall_Learning_Scheme}) has \textbf{three} following steps: 

\vspace{-1mm}
\subsubsection{Initialization}
\label{ssec:Initialization}
First, we start by initializing both teacher $F_t$ and student $F_s$ network with the supervised fine-tuned model $F_{ft}$ (as explained in Section \ref{ssec:Supervised fine-tuning}). Both teacher and student are identical networks consisting of ResNet-18 backbone (until the global average pooling layer) + 2-layer MLP + Fc. We freeze the weights of the teacher network across all layers; while the student network consisting of the last 2-layer MLP (Fc1, ReLU, Fc2) and a linear classifier/regressor (Fc) is being trained, keeping the backbone weights (until the output of global average pooling layer) fixed. 

\vspace{-1mm}
\subsubsection{Training the student model}
\label{ssec:Training the student model}
Second, we use the teacher network $F_t$ to generate pseudo labels on the deliberately noised unlabeled data $D_{f_u}$. Then, a student network $F_s$ is trained via standard supervised loss (on labeled data) and consistency loss (on unlabeled data). i.e., the supervised loss is evaluated by comparing against the ground truth labels (cross-entropy (CE) for classification / mean squared error (MSE) for regression task); while, the consistency loss (CE for classification / MSE for regression task) is obtained by comparing against the pseudo labels (i.e., logits for regression or one-hot labels for classification) produced by the teacher network. More formally, we aim to minimize the following objective (\textbf{total loss}):
\begin{equation}
    \min_{\theta} \sum_{i=1}^{\alpha N} \mathcal{L}_{s}\big(F_{s}(x_{i};\theta_{s}), y_{i}\big) + \lambda ~\mathcal{L}_{c}\big(\{x_i\}_{i=1}^{N}; F(.), \theta_t, \eta_w, \theta_s, \eta_s \big),
    \label{Eq: SmSL}
\end{equation}
where, $\mathcal{L}_{s}$ is the \textbf{supervised loss} measured against the labeled inputs and $\mathcal{L}_{c}$ is the \textbf{consistency loss} evaluated between the same unlabeled inputs with different data augmentations. The term $\lambda$ is the weighting factor which is empirically set as $1$, that controls the trade-off between the supervised and consistency loss. $F(.)$ denotes the ConvNet model parameterized by $\theta$, with $\theta_t$ and $\theta_s$ are the weights of the teacher and student network, respectively; while, $\eta_w$ and $\eta_s$ represents the weak and strong data augmentations applied to teacher and student model, respectively.

In each input minibatch, we sample $L=\big\{(x_i, y_i):i \in (1, ..., B)\big\}$ with $B$ labeled examples drawn from labeled set $D_{f_{l}}$ and $U=\big\{(x_i):i \in (1, ..., \mu B)\big\}$ with $\mu B$ unlabeled examples drawn from unlabeled set $D_{f_{u}}$, such that $\mu B >> B$. The term $\mu$ is a hyperparameter that determines the relative ratio of $U$ and $L$. Next, we define the \textbf{consistency loss} $\mathcal{L}_{c}$ for \textbf{regression} task as the distance between the prediction of teacher network $F_t$ (with weights $\theta_t$ and noise $\eta_w$) with the prediction of student network $F_s$ (with weights $\theta_s$ and noise $\eta_s$)~:
\begin{equation}
    \mathcal{L}_{c}^{\text{~regression}} = \sum_{i=1}^{\mu B} \mathbb{E}_{\eta_w, \eta_s} \big \lVert F_t(x_i, \theta_{t}, \eta_w) - F_s(x_i, \theta_{s}, \eta_s) \big \rVert^{2},
    \label{Eq: rlc}
\end{equation}
where, $x_i$ denotes each unlabeled training instance sampled from the unlabeled training set $D_{f_{u}}$. In contrast, for \textbf{classification} task, the consistency loss is calculated via standard cross-entropy (CE) loss defined by:
\vspace{-2mm}
\begin{equation}
    \mathcal{L}_{c}^{\text{~classification}} = \sum_{i=1}^{\mu B} \mathbbm{1}\big\{ \text{max} (q_i)\big\} ~\text{H} \big(\text{arg\hspace{0.5mm}max}(q_{i}), \hat{q}_{i}\big),
    \label{Eq: clc}
\end{equation}
where, $q_{i} = p_{t}\big(y_i~|~\eta_w(x_i)\big)$ is the predicted class probability by the teacher network $F_t$ for input $x_i$ applied with \textbf{weak augmentation} ($\eta_w$) and $\hat{q}_i = p_s\big(y_i~|~\eta_s(x_i)\big)$ is the predicted class probability by the student network $F_s$ for input $x_i$ applied with \textbf{strong augmentation} ($\eta_s$). The term H(.) denotes the CE between two probability distributions and $arg~max(q_i)$ is the \textbf{pseudo label} produced by the teacher network on weakly augmented unlabeled input image $\eta_w(x_i)$. In this work, we leverage two kinds of augmentations: weak and strong. The weak augmentation includes simple horizontal flip and cropping; while for strong augmentation, we use RandAugment technique \citep{Cubuk2020}. The complete list of data augmentations and their parameter settings are listed in Section \ref{ssec:Implementation details}.

During each epoch, we only update the weights of the student network while keeping the teacher network weights fixed. The student network weights are updated by learning a 2-layer MLP (Fc1, ReLU, Fc2) with a task-specific linear classifier/regressor on the output of global average pooling layer, with the rest of the layer weights being frozen. The idea of fine-tuning the last layers (i.e., 2-layer MLP and a linear classifier/regressor) of the student model improves the task-specific performance by using both labeled and unlabeled examples in a task-specific manner. This is because the effect of pretraining and most feature re-use happens in the lowest layers of the network, while fine-tuning higher layers change representations that are well adapted to the downstream tasks. This observation was also shown to be consistent in a recent study by \cite{Raghu2019}.

\vspace{-0.5mm}
\subsubsection{Updating the teacher model}
\label{ssec:Updating the teacher model}
At the end of each epoch, the teacher is replaced by the newly trained student $F_t \leftarrow F_s$  and this process is iterated until the model converges. In this way, our teacher-student consistency approach propagates the label information to the unlabeled data by constraining the model predictions to be consistent with the unlabeled data under different data augmentations. The pseudocode for our proposed consistency training is illustrated in Algorithm \ref{Algo:Consistency training pseudocode}. 
\begin{algorithm}
\footnotesize
\DontPrintSemicolon
\vspace{1mm}
\SetKwInOut{Parameter}{Inputs}
\Parameter{
$L = \big\{(x_i, y_i) : i \in (1, ..., B)\big\}$ with $B$ labeled examples drawn from labeled set $D_{f_{l}}$ \\
$U = \big\{(x_i) : i \in (1, ..., \mu B)\big\}$ with $\mu B$ unlabeled examples \\drawn from unlabeled set $D_{f_{u}}$ \\
$\mu$ = ratio of unlabeled data\\
$\lambda$ = weighting factor for consistency loss\\
$F_{ft}$ = fine-tuned model\\
$F_t$ = teacher model with parameter $\theta_t$\\
$F_s$ = student model with parameter $\theta_s$\\
$\eta_w$ = set of weak augmentations\\
$\eta_s$ = set of strong augmentations\\
$\eta$ = set of augmentations applied to labeled data}
\vspace{1mm}
\SetKwInOut{Parameter}{Initialize}
\Parameter{$F_t \leftarrow F_{ft}$, with weights frozen across entire network \\
\vspace{0.75mm}
$F_s \leftarrow F_{ft}$, with weights frozen until the output of \\ global average pooling layer; and training a 2-layer \\ MLP (Fc1, ReLU, Fc2) + a linear classifier/regressor}
\vspace{1mm}
\For{$t$ in [1, num\_epochs]}{\vspace{0.75mm}
\For{each minibatch $B$}{
\vspace{0.25mm}
$z_{i \in B}^{S} \leftarrow F_{s} \big(~\eta (L_{i \in B})~\big)$ \\
\vspace{0.25mm}
$z_{i \in \mu B}^{U} \leftarrow ~F_{t} \big(~\eta_{w} (U_{i \in \mu B})~\big)$ \\
\vspace{0.25mm}
$\hat{z}_{i \in \mu B}^{U} \leftarrow ~F_{s} \big(~\eta_{s} (U_{i \in \mu B})~\big)$ \\
\vspace{0.25mm}
$q_{i \in \mu B} = p_t\big(y_i~|~\eta_w(U_{i \in \mu B}); \theta_t\big)$ ~// prediction computed by $F_t$ \\
\vspace{0.25mm}
$\hat{q}_{i \in \mu B} = p_s\big(y_i~|~\eta_s(U_{i \in \mu B}); \theta_s\big)$ ~// prediction computed by $F_s$ \\
\vspace{0.75mm}
\colorbox{Gray!40}{supervised loss}\\
\vspace{0.5mm}
$\mathcal{L}_{s}^{classification} = \frac{-1}{|B|} \sum_{i \in B}$ log $z_{i}^{S}[y_i]$ \\
$\mathcal{L}_{s}^{regression} = \frac{1}{|B|} \sum_{i \in B} \lVert z_{i}^{S} - y_i \rVert^{2}$ \\
\vspace{0.75mm}
\colorbox{Gray!40}{consistency loss}\\
\vspace{0.5mm}
$\mathcal{L}_{c}^{classification} = \frac{1}{|\mu B|} \sum_{i \in \mu B} \mathbbm{1}~\{ \text{max} (q_i)\} ~\text{H}$(arg max$(q_i), \hat{q}_{i})$ \\ 
$\mathcal{L}_{c}^{regression} = \frac{1}{|\mu B|} \sum_{i \in \mu B} \lVert z_{i}^{U} - \hat{z}_{i}^{U} \rVert^{2}$ \\
\vspace{0.75mm}
\colorbox{Gray!40}{total loss}\\
\vspace{0.5mm}
loss $\leftarrow \mathcal{L}_{s} + \lambda \mathcal{L}_{c}$ \\
update $\theta_s$ using optimizer}
$F_t \leftarrow F_s$ ~// make student as the new teacher and go back to step 2}
\textbf{return} $\theta_s$;
\caption{Consistency training pseudocode}
\label{Algo:Consistency training pseudocode}
\end{algorithm}

\vspace{-3mm}
\section{Experiments}
\label{sec:Experiments}
We evaluate the efficacy of our method on one regression and two classification tasks on histopathology benchmark datasets, including BreastPathQ \citep{Martel2019tcia}, Camelyon16 \citep{bejnordi2017diagnostic} and Kather multi-class \citep{kather2019predicting}. Further, we also show extensive ablation experiments and compare them with state-of-the-art SSL methods by varying different percentages of labeled data.    

For baselines, we compare our SSL approach (i.e., RSP) with two other popular SSL methods, including the supervised one: VAE \citep{kingma2013auto} (refer, \ref{sec:Variational autoencoder}), MoCo \citep{he2020momentum} (refer, \ref{sec:Momentum Contrast (MoCo)}), and the random weight initialized (supervised). To further evaluate our approach on task-specific consistency training, we fine-tune the same self-supervised pretrained models for the second time using different percentages of task-specific labeled data. In our experiments, we first initialize the teacher-student model with the fine-tuned SSL model trained on different percentages of labeled data ($\alpha$): 10\%, 25\%, 50\%, and 100\% (depicted as ``self-supervised pretraining and supervised fine-tuning" in Table \ref{tab:BreastPathQ}, \ref{tab:Camelyon16}, \ref{tab:Kather}). Next, we train each of these fine-tuned models again for the second time using labeled and unlabeled samples, again by varying percentages of labeled data ($\alpha$) and report the final results (depicted as ``consistency training (CR)" as shown in Table \ref{tab:BreastPathQ}, \ref{tab:Camelyon16}, \ref{tab:Kather}). Note: this experimental setting is kept standard across all three datasets for a fair evaluation. 
\vspace{-1mm}

\subsection{Datasets}
\label{ssec:Datasets}
The distribution of the number of WSI's and their corresponding patches in all three datasets used for our experiments is shown in Table \ref{tab:dist_images}. In this section, we briefly describe all three publicly available datasets, whereas the data-specific implementations such as pretraining, fine-tuning, and test splits adopted in our experiments are explained in their respective subsections. 
\begin{table}[t]
\caption{The total number of WSI's and the patches used in each dataset to perform experiments.}
\label{tab:dist_images}
\resizebox{\linewidth}{!}{
\begin{tabular}{@{}lcccccc@{}}
\toprule[0.50pt]
\multicolumn{2}{l}{\multirow{2}{*}{\textbf{Datasets}}} & \multicolumn{2}{c}{\textbf{Pretrain}} & \multicolumn{2}{c}{\textbf{Fine-tune}} & \multirow{2}{*}{\textbf{Test}} \\ \cmidrule(lr){3-4} \cmidrule(lr){5-6}
\multicolumn{2}{l}{} & Train & Validation & Train & Validation &  \\ \midrule \midrule
\multirow{2}{*}{\textbf{BreastPathQ}} & WSIs & \multicolumn{2}{c}{69} & \multicolumn{2}{c}{69} & 25 \\ \cmidrule(l){2-7} 
 & Patches & 10000 & 3000 & 2063 & 516 & 1121 \\ \midrule
\multirow{2}{*}{\textbf{Camelyon16}} & WSIs & \multicolumn{2}{c}{60} & \multicolumn{2}{c}{210} & 129 \\ \cmidrule(l){2-7} 
 & Patches & 62156 & 10000 & 306303 & 40000 & -- \\ \midrule
\multirow{2}{*}{\begin{tabular}[c]{@{}l@{}}\textbf{Kather} \\ \textbf{Multi-class}\end{tabular}} & WSIs & \multicolumn{2}{c}{--} & \multicolumn{2}{c}{--} & -- \\ \cmidrule(l){2-7} 
 & Patches & -- & -- & 80K & 20K & 7180 \\ \bottomrule[0.50pt]
\end{tabular}}
\vspace{-1mm}
\end{table}

\vspace{1mm}
\textbf{BreastPathQ dataset:} This is a publicly available dataset consisting of hematoxylin and eosin (H\&E) stained 96 WSI's of post-NAT-BRCA specimens \citep{Martel2019tcia, Peikari2017}, which are scanned at $20\times$ magnification level ($0.5~lm/pixel$). A set of 2579 patches each with dimension $512 \times 512$ are extracted from 69 WSI's for training, and the remaining 1121 patches are extracted from 25 WSI's, which are reserved for testing. Two expert pathologists label the images in this dataset according to the percentage of cancer cellularity in each image patch. 

\vspace{1mm}
\textbf{Camelyon16 dataset:} We performed classification of breast cancer metastases at slide-level on the dataset from Camelyon16 challenge \citep{bejnordi2017diagnostic}. This dataset contains 399 H\&E stained WSI's of lymph nodes in the breast, which is split into 270 for training and 129 for testing. The images are acquired from two different scanners with specimen level pixel sizes of $0.226 ~\mu m/pixel$ and $0.243 ~\mu m/pixel$ spatial resolution and are exhaustively annotated by pathologists.

\vspace{1mm}
\textbf{Kather multiclass dataset:}
This dataset contains two subsets of patches containing nine tissue classes: adipose, background, debris, lymphocytes, mucus, smooth muscle, normal colon mucosa, cancer-associated stroma, and colorectal cancer epithelium \citep{kather2019predicting}. Out of the two subsets, the training set consists of 100K image patches of H\&E stained colorectal cancer images of $224 \times 224$ pixels scanned at $0.5 ~\mu m/pixel$ spatial resolution. In contrast, the test set contains 7180 image patches. In this dataset, only patches are made available without access to WSIs. 
\vspace{-1mm}

\subsection{Implementation details}
\label{ssec:Implementation details}
\begin{table}
\caption{List of hyperparameters used in our experiments across all three datasets.}
\label{tab:List of hyperparameters}
\resizebox{\linewidth}{!}{
\begin{tabular}{@{}clcccc@{}}
\toprule[1pt]
\multicolumn{2}{c}{\textbf{Hyperparameters}} & \textbf{BreastPathQ} & \textbf{Camelyon16} & \begin{tabular}[c]{@{}c@{}}\textbf{Kather}\\ \textbf{Multiclass}\end{tabular} \\ \midrule \midrule
\raisebox{-10.0\normalbaselineskip}[1pt][1pt]{\rotatebox[origin=l]{90}{\colorbox{Gray!10}{\textbf{Supervised fine-tuning}}}} & Epochs & 90 & 90 & 90 \\ \cmidrule(l){2-5} 
 & Batch size & 4 & 16 & 64 \\ \cmidrule(l){2-5} 
 & \begin{tabular}[c]{@{}l@{}}Learning \\ rate $(lr)$\end{tabular} & 0.0001 & 0.0005 & 0.00001 \\ \cmidrule(l){2-5} 
 & Optimizer & \begin{tabular}[c]{@{}c@{}}Adam \\ $\beta_1$, $\beta_2=(0.9, 0.999)$ \\ $w_d=1e-^{-4}$\end{tabular} & \begin{tabular}[c]{@{}c@{}}SGD with \\ Nesterov momentum \\ of 0.9, $w_d=1e-^{-4}$\end{tabular} & \begin{tabular}[c]{@{}c@{}}Adam \\ $\beta_1$, $\beta_2=(0.9, 0.999)$ \\ $w_d=1e-^{-4}$\end{tabular} \\ \cmidrule(l){2-5} 
 & Scheduler & \begin{tabular}[c]{@{}c@{}}MultiStep with $lr$ \\ decay at [30, 60] \\ epochs by 0.1\end{tabular} & \begin{tabular}[c]{@{}c@{}}MultiStep with $lr$ \\ decay at [30, 60] \\ epochs by 0.1\end{tabular} & \begin{tabular}[c]{@{}c@{}}MultiStep with $lr$ \\ decay at [30, 60] \\ epochs by 0.1\end{tabular} \\ \cmidrule(l){2-5} 
 & \begin{tabular}[c]{@{}c@{}}Selection of\\best model\end{tabular} & \begin{tabular}[c]{@{}c@{}}Lowest validation\\ loss\end{tabular} & \begin{tabular}[c]{@{}c@{}}Highest validation \\ accuracy\end{tabular} & \begin{tabular}[c]{@{}c@{}}Highest validation \\ accuracy\end{tabular} \\ \midrule[1.2pt]
\raisebox{-7.0\normalbaselineskip}[1pt][1pt]{\rotatebox[origin=c]{90}{\colorbox{Gray!10}{\textbf{Consistency training}}}} & Epochs & 90 & 90 & 90 \\ \cmidrule(l){2-5} 
 & Batch size & 4 & 8 & 8 \\ \cmidrule(l){2-5} 
 & \begin{tabular}[c]{@{}l@{}}Ratio of\\unlabeled data \\ ($\mu$)\end{tabular} & 7 & 7 & 7 \\ \cmidrule(l){2-5} 
 & \begin{tabular}[c]{@{}l@{}}Learning \\ rate ($lr$)\end{tabular} & 0.0001 & 0.0005 & 0.00001 \\ \cmidrule(l){2-5} 
 & Optimizer & \begin{tabular}[c]{@{}c@{}}Adam \\ $\beta_1$, $\beta_2=(0.9, 0.999)$ \\ $w_d=1e-^{-4}$\end{tabular} & \begin{tabular}[c]{@{}c@{}}SGD with \\ Nesterov momentum \\ of 0.9, $w_d=1e-^{-4}$\end{tabular} & \begin{tabular}[c]{@{}c@{}}Adam \\ $\beta_1$, $\beta_2=(0.9, 0.999)$ \\ $w_d=1e-^{-4}$\end{tabular} \\ \cmidrule(l){2-5} 
 & Scheduler & \begin{tabular}[c]{@{}c@{}}MultiStep with $lr$ \\ decay at [30, 60] \\ epochs by 0.1\end{tabular} & \begin{tabular}[c]{@{}c@{}}MultiStep with $lr$ \\ decay at [30, 60] \\ epochs by 0.1\end{tabular} & \begin{tabular}[c]{@{}c@{}}MultiStep with $lr$ \\ decay at [30, 60] \\ epochs by 0.1\end{tabular} \\ \cmidrule(l){2-5} 
 & \begin{tabular}[c]{@{}c@{}}Selection of\\best model\end{tabular} & \begin{tabular}[c]{@{}c@{}}Lowest validation\\ loss\end{tabular} & \begin{tabular}[c]{@{}c@{}}Highest validation\\ accuracy\end{tabular} & \begin{tabular}[c]{@{}c@{}}Highest validation\\ accuracy\end{tabular} \\ \bottomrule[1pt]
\end{tabular}}
\vspace{-4.5mm}
\end{table}
We perform all our experiments by selecting ResNet-18 as the backbone feature embedding network on all three datasets. All the experiments were performed on 4 Tesla NVIDIA V100 GPUs, and the entire framework is implemented in PyTorch. We first specify the implementation details common to all datasets, and data-specific implementations are provided in Table \ref{tab:List of hyperparameters}.

\vspace{1mm}
For \textbf{self-supervised pretraining}. The model is trained for 250 epochs with a batch size of 64. We employ (SGD with Nesterov momentum + Lookahead) optimizer \citep{zhang2019}, with a momentum of 0.9, weight decay of $1e^{-4}$ and a constant learning rate of 0.01. For Lookahead, we set $k = 5$ and slow weights step size $\alpha = 0.5$. The details of the MoCo and VAE pretraining is provided in \ref{sec:Momentum Contrast (MoCo)} and \ref{sec:Variational autoencoder}. The best pretrained model is chosen based on the lowest validation loss for BreastPathQ and Camelyon16 datasets. 

We adopt domain-specific data augmentations recommended by \cite{tellez2019quantifying}, including rotations, horizontal flips, scaling, additive Gaussian noise, brightness and contrast perturbations, shifting hue and saturation values in HSV color space, and perturbations in H\&E color space. We also add random resized crops, blur, and affine transformations to the previous list. Specifically, we use a rotation factor between [$-90 ^{\circ}, +90 ^{\circ}$], scaling factor between [0.8, 1.2], additive Gaussian noise with [$\mu=0$ and $\sigma=(0, 0.1)$], affine transformation with translation, scale and rotation limit of [$0.0625, 0.5, 45 ^{\circ}$], respectively, hue and saturation intensity ratio between [-0.1, 0.1] and [-1, 1], respectively, brightness and contrast intensity ratios between [-0.2, 0.2], blurring the input image using a random-sized kernel in the range [3, 7], and randomly resizing and cropping the image patch to its original image size. Finally, we perturb the intensity of hematoxylin and eosin (HED) color channels with a factor of [-0.035, 0.035]. We apply all these transformations in sequence by randomly selecting them in each mini-batch to obtain a diverse set of training images. 

\vspace{1mm}
For \textbf{supervised fine-tuning}. We fine-tune the entire pretrained SSL model (all layers) with a linear classifier or a regressor trained on top of the learned representations with limited labeled examples $\alpha = \{10\%, 25\%, 50\%, 100\%\}$ to directly evaluate the performance of RSP, VAE, and MoCo models. In particular, for RSP, we fine-tune the pretrained network (i.e., ResNet-18 (until global average pooling layer) + 2-layer MLP) with a linear layer on the top of $d=(256 \times 3) = 768$ dimensional ($d$) feature embedding, followed by softmax to obtain task-specific predictions. However, for VAE and MoCo, we fine-tune with a linear layer on the $512-d$ feature vector. For fine-tuning, we use different sets of hyperparameters for all three datasets, which are provided in Table \ref{tab:List of hyperparameters}. Further, we include a simple set of augmentations ($\eta$ as depicted in Algorithm \ref{Algo:Consistency training pseudocode}), such as rotation, scaling, and random resized crops. For rotation and scaling, we use a factor of [$-90 ^{\circ}, +90 ^{\circ}$] and [0.8, 1.2], respectively, and we randomly resize and crop the image patch to its original image size. 

\vspace{1mm}
For \textbf{consistency training}. We use a semi-supervised approach for consistency training by using labeled and unlabeled examples in a task-specific manner. We adopt the same task-specific fine-tuned model to initialize both teacher and student network, with teacher network weights frozen across all layers; while training a student network with 2-layer MLP (Fc1, ReLU, Fc2) and a task-specific linear layer (classifier/regressor) on the output of global average pooling layer (with rest of the layer weights fixed). All the hyperparameters related to consistency training are shown in Table \ref{tab:List of hyperparameters}. In our experiments, we initialize the teacher-student model with the fine-tuned SSL model trained on different percentages of labeled data $\alpha = \{10\%, 25\%, 50\%, 100\%\}$. Next, we train each of these fine-tuned models again for the second time using labeled and unlabeled samples, again by varying percentages of labeled data ($\alpha$), and report the final results.

In our work, we use two kinds of augmentations for consistency training: ``\textbf{weak}" and ``\textbf{strong}" augmentation for teacher and student network, respectively. We employ simple transformations such as horizontal flip and random cropping to its original image size as weak augmentations for the teacher network. Whereas for the student network, we adopt a similar set of transformations to the pretraining stage, but with different hyperparameters to strengthen the augmentation severity, which we refer to as strong augmentations. The following are the list of augmentations with different parameters to the pretraining stage: an affine transformation with translation limit of [0.01, 0.1], scale limit of [0.51, 0.60] and rotation of $90^{\circ}$, HSV intensity ratio between [-1, 1] and blurring the input image using a random-sized kernel in the range [5, 7]. We apply these augmentations in sequence by randomly selecting them in each mini-batch using the RandAugment technique \citep{Cubuk2020}. In our experiments, we use $N_{Aug}=7, M_{g}=[1, 10]$ in RandAugment; where, $N_{Aug}$ denotes the number of augmentations to apply sequentially in each mini-batch, and $M_g$ is the magnitude that is sampled within a pre-defined range [1, 10] that controls the severity of distortion in each mini-batch. 

\subsection{Experiments for BreastPathQ}
\label{ssec:Experiments for BreastPathQ}
In this experiment, we train our approach to automatically quantify tumor cellularity (TC) scores in digitized slides of breast cancer images for tumor burden assessment. TC \textit{score} is defined as the percentage of the total area occupied by the malignant tumor cells in a given image patch \citep{Peikari2017}. For pretraining the SSL approach, we adopted 69 WSI's of the training set, from which we randomly extract patches of size ($256 \times 256$) at $20\times$, $10\times$, and $5\times$ magnification for RSP, while for VAE and MoCo, patches are extracted at $20\times$ magnification. We perform fine-tuning by resizing the image patches to ($256 \times 256$) on 2579 training image patches (out of which 80\% (2063) are reserved for training and 20\% (516) for validation), and testing is done on 1121 image patches (as shown in Table \ref{tab:dist_images}). To experiment with limited data on the downstream task, we divide the fine-tuning set (i.e., 2063 patches) into four incremental training subsets: $\alpha = \{10\%, 25\%, 50\%, 100\%\}$. Two pathologists annotated each image patch in the test set according to the percentage of cancer cellularity in each image patch. We report the intra-class correlation coefficient (ICC) values between the proposed methods and the two pathologists A and B.

\begin{table*}[t]
\scriptsize
\centering
\caption{Results on BreastPathQ dataset. Predicting the percentage of tumor cellularity (TC) at patch-level (intra-class correlation (ICC) coefficients between two pathologists A and B). The 95\% confidence intervals (CI) are shown in square brackets. We bold the best results.}
\label{tab:BreastPathQ}
\resizebox{\linewidth}{!}{
\begin{tabular}{@{}lllllllll@{}}
\toprule[0.75pt]
\multirow{3}{*}{\begin{tabular}[c]{@{}c@{}}\\\textbf{Methods} \\ \\ \% Training Data ($\alpha$)\end{tabular}}  & \multicolumn{8}{c}{ICC Coefficient (95\% CI)} \\ \cmidrule{2-9} 
& \multicolumn{1}{c}{Pathologist A} & \multicolumn{1}{c}{Pathologist B} & \multicolumn{1}{c}{Pathologist A} & \multicolumn{1}{c}{Pathologist B} & \multicolumn{1}{c}{Pathologist A} & \multicolumn{1}{c}{Pathologist B} & \multicolumn{1}{c}{Pathologist A} & \multicolumn{1}{c}{Pathologist B} \\ \cmidrule(lr){2-3} \cmidrule(lr){4-5} \cmidrule(lr){6-7} \cmidrule(lr){8-9} 
& \multicolumn{2}{c}{10\% (206 labels)} & \multicolumn{2}{c}{25\% (516 labels)} & \multicolumn{2}{c}{50\% (1031 labels)} & \multicolumn{2}{c}{100\% (2063 labels)} \\ \midrule \midrule
\rowcolor[gray]{0.90}
\multicolumn{9}{l}{\textbf{Self-supervised pretraining + Supervised fine-tuning}} \\ \midrule 
Random & \multicolumn{1}{c}{0.697 {[}0.67, 0.73{]}} & \multicolumn{1}{c}{0.637 {[}0.60, 0.67{]}} & \multicolumn{1}{c}{0.786 {[}0.76, 0.81{]}} & \multicolumn{1}{c}{0.727 {[}0.70, 0.75{]}} & \multicolumn{1}{c}{0.812 {[}0.79, 0.83{]}} & \multicolumn{1}{c}{0.797 {[}0.77, 0.82{]}} & \multicolumn{1}{c}{0.863 {[}0.85, 0.88{]}} & \multicolumn{1}{c}{0.843 {[}0.83, 0.86{]}} \\ 
VAE & \multicolumn{1}{c}{\textbf{0.733 {[}0.70, 0.76{]}}} & \multicolumn{1}{c}{\textbf{0.693 {[}0.66, 0.72{]}}} & \multicolumn{1}{c}{0.767 {[}0.74, 0.79{]}} & \multicolumn{1}{c}{\textbf{0.756 {[}0.73, 0.78{]}}} & \multicolumn{1}{c}{0.790 {[}0.77, 0.81{]}} & \multicolumn{1}{c}{0.775 {[}0.75, 0.80{]}} & \multicolumn{1}{c}{0.853 {[}0.84, 0.87{]}} & \multicolumn{1}{c}{0.824 {[}0.80, 0.84{]}} \\ 
MoCo & \multicolumn{1}{c}{0.675 {[}0.64, 0.71{]}} & \multicolumn{1}{c}{0.648 {[}0.61, 0.68{]}} & \multicolumn{1}{c}{0.718 {[}0.69, 0.75{]}} & \multicolumn{1}{c}{0.651 {[}0.62, 0.68{]}} & \multicolumn{1}{c}{0.746 {[}0.72, 0.77{]}} & \multicolumn{1}{c}{0.711 {[}0.68, 0.74{]}} & \multicolumn{1}{c}{0.757 {[}0.73, 0.78{]}} & \multicolumn{1}{c}{0.718 {[}0.69, 0.75{]}} \\
\textbf{RSP (ours)} & \multicolumn{1}{c}{0.701 {[}0.67, 0.73{]}} & \multicolumn{1}{c}{0.667 {[}0.63, 0.70{]}} & \multicolumn{1}{c}{\textbf{0.796 {[}0.77, 0.82{]}}} & \multicolumn{1}{c}{0.734 {[}0.71, 0.76{]}} & \multicolumn{1}{c}{\textbf{0.842 {[}0.82, 0.86{]}}} & \multicolumn{1}{c}{\textbf{0.834 {[}0.82, 0.85{]}}} & \multicolumn{1}{c}{\textbf{0.884 {[}0.87, 0.90{]}}} & \multicolumn{1}{c}{\textbf{0.872 {[}0.86, 0.89{]}}} \\ \midrule[0.001pt]
\rowcolor[gray]{0.90}
\multicolumn{9}{l}{\textbf{Consistency training (CR)}} \\ \midrule 
Random + CR              & \multicolumn{1}{c}{0.658 {[}0.62, 0.69{]}}                     & \multicolumn{1}{c}{0.630 {[}0.59, 0.66{]}}  &  \multicolumn{1}{c}{0.818 {[}0.80, 0.84{]}}                     & \multicolumn{1}{c}{0.802 {[}0.78, 0.82{]}} & \multicolumn{1}{c}{0.847 {[}0.83, 0.86{]}}   & \multicolumn{1}{c}{0.839 {[}0.82, 0.86{]}}   & \multicolumn{1}{c}{0.891 {[}0.88, 0.90{]}} & \multicolumn{1}{c}{0.891 {[}0.88, 0.90{]}} \\  
VAE + CR & \multicolumn{1}{c}{0.771 {[}0.75, 0.79{]}} & \multicolumn{1}{c}{0.727 {[}0.70, 0.75{]}} & \multicolumn{1}{c}{0.842 {[}0.82, 0.86{]}} & \multicolumn{1}{c}{0.826 {[}0.81, 0.84{]}} & \multicolumn{1}{c}{0.866 {[}0.85, 0.88{]}} & \multicolumn{1}{c}{0.857 {[}0.84, 0.87{]}} & \multicolumn{1}{c}{0.884 {[}0.87, 0.90{]}} & \multicolumn{1}{c}{0.864 {[}0.85, 0.88{]}}\\ 
MoCo + CR  & \multicolumn{1}{c}{0.808 {[}0.79, 0.83{]}} & \multicolumn{1}{c}{0.803 {[}0.78, 0.82{]}} & \multicolumn{1}{c}{0.872 {[}0.86, 0.89{]}}  & \multicolumn{1}{c}{\textcolor{red}{\textbf{0.863 {[}0.85, 0.88{]}}}}  & \multicolumn{1}{c}{0.848 {[}0.83, 0.86{]}} & \multicolumn{1}{c}{0.850 {[}0.83, 0.87{]}} & \multicolumn{1}{c}{0.895 {[}0.88, 0.91{]}} & \multicolumn{1}{c}{0.902 {[}0.89, 0.91{]}} \\      
\textbf{RSP + CR (ours)} & \multicolumn{1}{c}{\textcolor{red}{\textbf{0.876 {[}0.86, 0.89{]}}}}                     & \multicolumn{1}{c}{\textcolor{red}{\textbf{0.846 {[}0.83, 0.86{]}}}} & \multicolumn{1}{c}{\textcolor{red}{\textbf{0.873 {[}0.86, 0.89{]}}}}                     & \multicolumn{1}{c}{0.854 {[}0.84, 0.87{]}}  & \multicolumn{1}{c}{\textcolor{red}{\textbf{0.870 {[}0.86, 0.88{]}}}} & \multicolumn{1}{c}{\textcolor{red}{\textbf{0.861 {[}0.84, 0.88{]}}}}  & \multicolumn{1}{c}{\textcolor{red}{\textbf{0.910 {[}0.90, 0.92{]}}}} & \multicolumn{1}{c}{\textcolor{red}{\textbf{0.907 {[}0.90, 0.92{]}}}} \\ \bottomrule[0.75pt]
\end{tabular}}
\end{table*}
Table \ref{tab:BreastPathQ} presents the ICC values for different methodologies, and the corresponding TC scores produced by each method on sample WSIs of the BreastPathQ test set are shown in Fig. \ref{Fig:Heat-maps on BreastPathQ} (shown in Appendix). The consistency training (CR) improved the results of self-supervised pretrained models (VAE, MoCo, and RSP) by a 3\% increase in ICC values. Further, all SSL and CR methods (VAE, MoCo, and RSP) seem to exhibit optimal performance, which is close or even outperforming that of supervised baseline (random) on all training subsets. Among all the methods, the RSP+CR achieves the best score of greater than $0.90$, which even surpassed the intra-rater agreement score of $0.89$ \citep{akbar2019automated}. Besides, our obtained TC score of $0.90$ on the BreastPathQ test set is superior to state-of-the-art (SOTA) methods \citep{akbar2019automated, rakhlin2019breast}, with a maximum score of 0.883. Specifically, our RSP+CR approach achieves a minimum of $4\%$ greater ICC value than VAE+CR and MoCo+CR, and at least 17\% improvement in ICC value to the supervised baseline, trained on 10\% labeled set ($\approx206$ image patches). In contrast, on a complete training set, all CR methods exhibit competitive/similar performance. This indicates that the consistency training improves upon self-supervised pretraining predominantly in the low-data regime. 

\vspace{-0.5mm}
\subsection{Experiments for Camelyon16}
\label{ssec:Experiments for Camelyon16}
This experiment is a slide-based binary classification task to identify the presence of lymph node metastasis in WSIs using only slide-level labels. To experiment with the limited annotations, we first perform self-supervised pretraining on 60 WSI's (35 normal and 25 tumor), which are set aside from the original training set. For pretraining, we randomly extract patches of size ($256 \times 256$) at $40\times$, $20\times$, and $10\times$ magnification for RSP, while for VAE and MoCo, patches are extracted at $40\times$ magnification. Further, the downstream fine-tuning is performed on the randomly extracted patches of size ($256 \times 256$) from the remaining 210 WSI's (125 normal and 85 tumor) of the training set, out of which 80\% (306.3K patches - 150K tumor + 156.3K normal) are reserved for training and 20\% (40K patches - 20K tumor + 20K normal) for validation. We finally evaluate the methods on 129 WSI's of the test set (as shown in Table \ref{tab:dist_images}). We divide the fine-tuning set containing 306.3K patches into four incremental subsets of $\alpha = \{10\%, 25\%, 50\%, 100\%\}$ containing [30.6K, 76.5K, 153.1K, 306.3K] image patches, respectively.

We follow the same post-processing steps as \cite{wang2016deep} to obtain slide-level predictions. We first train our proposed models to discriminate patch-level tumor vs. normal patches. We then aggregate these patch-level predictions to create a heat-map of tumor probability over the slide. Next, we extract several features similar to \cite{wang2016deep} from the heat map and train a slide-level support vector machine (SVM) classifier to make the slide-level prediction. We compare and evaluate all three SSL pretrained and CR methods with the corresponding supervised baseline. The method's performance is evaluated in terms of area under the receiver operating characteristic curve (AUC) on a test set containing 129 WSIs. In addition, we also evaluate the binary classification performance (accuracy (Acc)) on the patch-level data containing 40K patches (20K tumor + 20K normal) of the validation set. Further, we perform the statistical significance test by comparing the pairs of AUCs between consistency training and SSL methods using the two-tailed Delong's test \citep{sun2014fast}. All differences in AUC value with a $p$-value $<0.05$ were considered significant.

\begin{table*}
\scriptsize
\centering
\caption{Results on Camelyon16 dataset. Predicting the presence of tumor metastasis at WSI level (AUC) and patch-level classification performance (accuracy). The DeLong method \citep{sun2014fast} was used to construct 95\% CIs, which are shown in square brackets. The best scores are shown in bold. Note: the patch-level accuracy (Acc) is reported on 40K patches of the validation set.}
\label{tab:Camelyon16}
\resizebox{\linewidth}{!}{
\begin{tabular}{@{}lllllllll@{}}
\toprule[0.75pt]
\multirow{3}{*}{\begin{tabular}[c]{@{}c@{}}\textbf{Methods} \vspace{2mm}\\\vspace{2mm} \% Training Data ($\alpha$)\end{tabular}} & \multicolumn{1}{c}{AUC} & \multicolumn{1}{c}{Acc} & \multicolumn{1}{c}{AUC} & \multicolumn{1}{c}{Acc} & \multicolumn{1}{c}{AUC} & \multicolumn{1}{c}{Acc} & \multicolumn{1}{c}{AUC} & \multicolumn{1}{c}{Acc} \\ \cmidrule(lr){2-3} \cmidrule(lr){4-5} \cmidrule(lr){6-7} \cmidrule(lr){8-9}  
& \multicolumn{2}{c}{10\% (30630 labels)~~~(4000 labels)} & \multicolumn{2}{c}{25\% (76576 labels)~~~(10000 labels)} & \multicolumn{2}{c}{50\% (153151 labels)~~~(20000 labels)} & \multicolumn{2}{c}{100\% (306303 labels)~~~(40000 labels)} \\ \midrule \midrule
\rowcolor[gray]{0.90}
\multicolumn{9}{l}{\textbf{Self-supervised pretraining + Supervised fine-tuning}} \\ \midrule
Random                   & \multicolumn{1}{c}{0.804 {[}0.72 - 0.89{]}}          & \multicolumn{1}{c}{\textbf{0.904}} & \multicolumn{1}{c}{0.861 {[}0.79 - 0.93{]}}          & \multicolumn{1}{c}{\textbf{0.936}} & \multicolumn{1}{c}{0.847 {[}0.77 - 0.92{]}}          & \multicolumn{1}{c}{0.946} & \multicolumn{1}{c}{0.865 {[}0.79 - 0.93{]}}          & \multicolumn{1}{c}{\textcolor{red}{\textbf{0.968}}} \\ 
VAE                      & \multicolumn{1}{c}{0.737 {[}0.64 - 0.83{]}}          & \multicolumn{1}{c}{0.827}    & \multicolumn{1}{c}{0.814 {[}0.73 - 0.89{]}}          & \multicolumn{1}{c}{0.864} & \multicolumn{1}{c}{0.830 {[}0.75 - 0.91{]}}          & \multicolumn{1}{c}{0.906} & \multicolumn{1}{c}{0.818 {[}0.73 - 0.90{]}}          & \multicolumn{1}{c}{0.907}          \\ 
MoCo                     & \multicolumn{1}{c}{\textcolor{red}{\textbf{0.895 {[}0.84 - 0.95{]}}}} & \multicolumn{1}{c}{0.837}           & \multicolumn{1}{c}{0.867 {[}0.80 - 0.93{]}}          & \multicolumn{1}{c}{0.895}    & \multicolumn{1}{c}{\textcolor{red}{\textbf{0.877 {[}0.81 - 0.94{]}}}} & \multicolumn{1}{c}{0.904}      & \multicolumn{1}{c}{0.857 {[}0.78 - 0.93{]}}          & \multicolumn{1}{c}{0.921}          \\ 
\textbf{RSP (ours)}      & \multicolumn{1}{c}{0.836 {[}0.76 - 0.91{]}}          & \multicolumn{1}{c}{0.898}   & \multicolumn{1}{c}{\textbf{0.886 {[}0.83 - 0.94{]}}} & \multicolumn{1}{c}{0.928}  & \multicolumn{1}{c}{0.861 {[}0.79 - 0.93{]}}  & \multicolumn{1}{c}{\textbf{0.946}} & \multicolumn{1}{c}{\textbf{0.878 {[}0.81 - 0.95{]}}} & \multicolumn{1}{c}{0.953} \\ \midrule[0.001pt]
\rowcolor[gray]{0.90}
\multicolumn{9}{l}{\textbf{Consistency training (CR)}} \\ \midrule 
Random + CR              & \multicolumn{1}{c}{0.659 {[}0.54 - 0.77{]}}          & \multicolumn{1}{c}{\textcolor{red}{\textbf{0.911}}}            & \multicolumn{1}{c}{0.782 {[}0.69 - 0.87{]}}         & \multicolumn{1}{c}{\textcolor{red}{\textbf{0.948}}}   & \multicolumn{1}{c}{0.783 {[}0.69 - 0.87{]}}         & \multicolumn{1}{c}{\textcolor{red}{\textbf{0.955}}}   & \multicolumn{1}{c}{0.870 {[}0.80 - 0.94{]}}         & \multicolumn{1}{c}{\textbf{0.964}} \\ 
VAE + CR                 & \multicolumn{1}{c}{0.633 {[}0.55 - 0.72{]}}          & \multicolumn{1}{c}{0.828}            & \multicolumn{1}{c}{0.719 {[}0.63 - 0.81{]}}          & \multicolumn{1}{c}{0.863}    & \multicolumn{1}{c}{0.741 {[}0.64 - 0.84{]}}         & \multicolumn{1}{c}{0.918}   & \multicolumn{1}{c}{0.779 {[}0.69 - 0.87{]}}          & \multicolumn{1}{c}{0.928} \\
MoCo + CR                 & \multicolumn{1}{c}{0.728 [0.63 - 0.82]}          & \multicolumn{1}{c}{0.835}            & \multicolumn{1}{c}{0.742 [0.64 - 0.84]}          & \multicolumn{1}{c}{0.902}    & \multicolumn{1}{c}{0.766 [0.67 - 0.86]
}         & \multicolumn{1}{c}{0.929}   & \multicolumn{1}{c}{0.825 [0.75 - 0.90]}          & \multicolumn{1}{c}{0.946}\\ 

\textbf{RSP + CR (ours)}                 & \multicolumn{1}{c}{\textbf{0.855 [0.78 - 0.92]}}          & \multicolumn{1}{c}{0.907}            & \multicolumn{1}{c}{\textcolor{red}{\textbf{0.917 [0.86 - 0.97]}}}          & \multicolumn{1}{c}{0.935}    & \multicolumn{1}{c}{\textbf{0.848 [0.77 - 0.93]}}         & \multicolumn{1}{c}{0.949
}   & \multicolumn{1}{c}{\textcolor{red}{\textbf{0.882 [0.80 - 0.96]}}
}          & \multicolumn{1}{c}{0.959}\\ \bottomrule[0.75pt]
\end{tabular}}
\end{table*}
Table \ref{tab:Camelyon16} presents the AUC scores for predicting slide-level tumor metastasis using different methodologies. On the 10\% label regime, RSP and MoCo methods outperformed the supervised baseline, whereas the performance of VAE is significantly decreased compared to other methods. Further, the RSP+CR approach significantly outperforms the RSP by a margin of 2\% on 10\% and 25\% labeled set. The proposed RSP+CR achieves the best score of 0.917 using a 25\% labeled set ($\approx 76$K patches) compared to the winning method in Camelyon16 challenge \citep{wang2016deep}, which obtained an AUC of 0.925 using the fully supervised model trained on millions of image patches. Compared with the unsupervised representation learning methods proposed in \cite{tellez2019neural}, our RSP+CR approach trained on 10\% labels ($\approx 30$K patches) outperforms their top-performing BiGAN method by 13\% higher AUC trained on 50K labeled samples. We also evaluated our methods performance on the validation set containing 40K patches (20K tumor + 20K normal). Surprisingly, the supervised baseline (Random, Random+CR) outperformed the RSP, RSP+CR methods by a slight margin difference of 0.5\% Acc on all percentages of training subsets. 

Most importantly, from our experiments on the Camelyon16 dataset, we draw several insights on the generality of our approach on \textbf{low-} and \textbf{high-labels} training scenarios. On a low-label data regime, \textit{i.e.,} the patch-wise classification task on the validation set, which has training labels ranging from 4K to 40K, we observe that adding consistency training improved the SSL model performance up to a 2\% increase in Acc values. In addition, the AUCs of consistency trained models are statistically higher than AUCs of SSL pretrained models, with $p$-value $<0.02$ across 10\% and 25\% labeled sets. Further, as we increase the number of labeled samples (50\% to 100\%), adding the consistency training to the Random, VAE, and MoCo pretrained models resulted in a noticeable drop in AUC values. However, the results for the RSP model still improved after consistency training in the high-label data regime, but these differences were not statistically significant. Thus, in general, our approach has been shown to work well in a limited annotation setting, which is highly beneficial in the histopathology domain. 

In our experiments, we also observe that the pretraining performance slightly diminishes with an increase in the amount of labeled data (from 10\% (30K) to 100\% (306K) labels), which essentially deteriorates the value of pretrained representations, which is consistent with the recent study by \cite{zoph2020rethinking}. Overall, our consistency training approach continues to improve the task-specific performance only when trained with less label data, and it is additive to pretraining. Fig. \ref{Fig:Heat-maps on Camelyon16} (shown in Appendix) highlight the tumor probability heat-maps produced by different methodologies. Visually all self-supervised pretrained methods (VAE, MoCo, and RSP) were shown to focus on tumor areas with high probability, while the supervised baseline exhibits slightly lower probability values for the same tumor regions. We observe that most methods successfully identify the macro-metastases (Row 1-3), with a tumor diameter larger than 2\textit{mm}, with an excellent agreement with the ground truth annotation. However, the same methods struggle to precisely identify the micro-metastases (Row 4), with tumor diameter smaller than 2\textit{mm}, which is generally challenging even for the fully-supervised models.
\vspace{-1mm}

\subsection{Experiments for Kather Multiclass}
\label{ssec:Experiments for Kather Multiclass}
Due to the unavailability of access to WSIs in this dataset, we could not perform self-supervised pretraining on this dataset. However, we used the SSL pretrained model of Camelyon16 to fine-tune and evaluate the patch-level performance for \textbf{feature transferability} between datasets with different tissue types/organs and resolution protocols. In our experiments, the downstream fine-tuning is performed on 100k image patches of the training set and tested on 7180 test set images by resizing the patches to $(256 \times 256)$ pixels. 

\begin{table*}[t]
\scriptsize
\centering
\caption{Results on Kather Multiclass dataset. Classification of nine tissue types at patch-level (accuracy (Acc), weighted $F_1$ score ($F_1$)). This experiment is performed to assess the generalizability of pretrained features between different tissue types and resolutions. Pretraining is performed on Camelyon16 (Breast) and tested on Kather Multiclass (Colon). We bold the best results.}
\label{tab:Kather}
\resizebox{0.65\linewidth}{!}{
\begin{tabular}{@{}lllllllll@{}}
\toprule[0.75pt]
\multirow{3}{*}{\begin{tabular}[c]{@{}c@{}}\textbf{Methods} \vspace{2mm}\\\vspace{2mm} \% Training Data ($\alpha$)\end{tabular}} & \multicolumn{1}{c}{Acc} & \multicolumn{1}{c}{$F_{1}$} & \multicolumn{1}{c}{Acc} & \multicolumn{1}{c}{$F_{1}$} & \multicolumn{1}{c}{Acc} & \multicolumn{1}{c}{$F_{1}$} & \multicolumn{1}{c}{Acc} & \multicolumn{1}{c}{$F_{1}$} \\ \cmidrule(lr){2-3} \cmidrule(lr){4-5} \cmidrule(lr){6-7} \cmidrule(lr){8-9}  
& \multicolumn{2}{c}{10\% (8000 labels)} & \multicolumn{2}{c}{25\% (20000 labels)} & \multicolumn{2}{c}{50\% (40000 labels)} & \multicolumn{2}{c}{100\% (80000 labels)} \\ \midrule \midrule
\rowcolor[gray]{0.90}
\multicolumn{9}{l}{\textbf{Self-supervised pretraining + Supervised fine-tuning}} \\ \midrule
Random                   & \multicolumn{1}{c}{0.972}          & \multicolumn{1}{c}{0.873} & \multicolumn{1}{c}{0.974}          & \multicolumn{1}{c}{0.885} & \multicolumn{1}{c}{0.979}          & \multicolumn{1}{c}{0.905} & \multicolumn{1}{c}{0.983}          & \multicolumn{1}{c}{0.920} \\ 
VAE                      & \multicolumn{1}{c}{0.963}          & \multicolumn{1}{c}{0.835}    & \multicolumn{1}{c}{0.972}          & \multicolumn{1}{c}{0.885} & \multicolumn{1}{c}{0.980}          & \multicolumn{1}{c}{0.908} & \multicolumn{1}{c}{\textbf{0.986}}          & \multicolumn{1}{c}{\textbf{0.934}}          \\ 
MoCo                     & \multicolumn{1}{c}{\textbf{0.982}} & \multicolumn{1}{c}{\textbf{0.919}}           & \multicolumn{1}{c}{\textbf{0.982}}          & \multicolumn{1}{c}{\textbf{0.919}}    & \multicolumn{1}{c}{\textbf{0.985}} & \multicolumn{1}{c}{\textbf{0.930}}      & \multicolumn{1}{c}{0.975}          & \multicolumn{1}{c}{0.849}          \\ 
\textbf{RSP (ours)}      & \multicolumn{1}{c}{0.976}          & \multicolumn{1}{c}{0.893}   & \multicolumn{1}{c}{0.975} & \multicolumn{1}{c}{0.888}  & \multicolumn{1}{c}{0.979}  & \multicolumn{1}{c}{0.907} & \multicolumn{1}{c}{0.979} & \multicolumn{1}{c}{0.911} \\ \midrule[0.001pt]
\rowcolor[gray]{0.90}
\multicolumn{9}{l}{\textbf{Consistency training (CR)}} \\ \midrule 
Random + CR              & \multicolumn{1}{c}{0.938}          & \multicolumn{1}{c}{0.670}            & \multicolumn{1}{c}{0.943}         & \multicolumn{1}{c}{0.735}   & \multicolumn{1}{c}{0.941}         & \multicolumn{1}{c}{0.723}   & \multicolumn{1}{c}{0.939}         & \multicolumn{1}{c}{0.707} \\ 
VAE + CR                 & \multicolumn{1}{c}{0.972}          & \multicolumn{1}{c}{0.876}            & \multicolumn{1}{c}{0.979}          & \multicolumn{1}{c}{0.906}    & \multicolumn{1}{c}{0.978}         & \multicolumn{1}{c}{0.903}   & \multicolumn{1}{c}{0.982}          & \multicolumn{1}{c}{0.915} \\
MoCo + CR                 & \multicolumn{1}{c}{\textcolor{red}{\textbf{0.987}}}          & \multicolumn{1}{c}{\textcolor{red}{\textbf{0.939}}}            & \multicolumn{1}{c}{\textcolor{red}{\textbf{0.990}}}          & \multicolumn{1}{c}{\textcolor{red}{\textbf{0.953}}}    & \multicolumn{1}{c}{\textcolor{red}{\textbf{0.987}}}         & \multicolumn{1}{c}{\textcolor{red}{\textbf{0.944}}}   & \multicolumn{1}{c}{0.983}          & \multicolumn{1}{c}{0.921}\\ 

\textbf{RSP + CR (ours)}                 & \multicolumn{1}{c}{0.982}          & \multicolumn{1}{c}{0.918}            & \multicolumn{1}{c}{0.982}          & \multicolumn{1}{c}{0.913}    & \multicolumn{1}{c}{0.985}         & \multicolumn{1}{c}{0.930}   & \multicolumn{1}{c}{\textcolor{red}{\textbf{0.986}}
}          & \multicolumn{1}{c}{\textcolor{red}{\textbf{0.934}}}\\ \bottomrule[0.75pt]
\end{tabular}}
\vspace{-3mm}
\end{table*}
Table \ref{tab:Kather} presents the overall Acc and weighted $F_{1}$ score ($F_{1}$) for classification of 9 colorectal tissue classes using different methodologies. On this dataset, the MoCo+CR approach obtains a new state-of-the-art result with an Acc of 0.990, weighted $F_{1}$ score of 0.953 and a macro AUC of 0.997, compared to the previous method \citep{kather2019predicting} which obtained an Acc of 0.943. All the consistency trained methods marginally outperform the SSL pretrained models on all the labeled subsets. Further, the CR methods (RSP+CR, MoCo+CR, VAE+CR) outperform the supervised baseline by 3\% and 17\% increase in Acc and $F_{1}$ score, respectively. Finally, our approach has shown 3\% improvement in Acc by training on just 10\% labels, compared to the previous method \citep{pati2020reducing} trained using 100\% labels (Acc of 0.951). This underscores that our pretrained approaches are more generalizable to unseen domains with different organs, tissue types, staining, and resolution protocols.

Further, we also observed that the generalizability of RSP is not as good as MoCo for feature transferability between datasets (Camelyon16 $\rightarrow$ Kather Multiclass), as seen in Table \ref{tab:Kather}. We hypothesize that this could be for the following reason: the idea of SSL is to learn generic visual representations that are useful for many downstream tasks across domains. However, in many recent studies \citep{misra2020self, purushwalkam2020demystifying, xiao2020should}, it is shown that the generalizability of these pretrained representations is largely dependent on their ability to learn representations that are \textbf{\textit{invariant}} to different image transformations. This invariance to data transformations is inherently learned in MoCo via instance discrimination strategy, which involves treating an image and its transformed versions as one single class. In contrast, the RSP method learns representations that are \textbf{\textit{covariant}} to the pretext task's transformation (i.e., resolution sequence ordering) rather than invariance to data augmentations (such as in MoCo and SimCLR). In histopathology, these invariances to data augmentations play a vital role in transferring representations to a new target domain, which is consistent with the recent study in \citep{tellez2019quantifying}. One future work in improving the RSP could be learning representations that are invariant to both data augmentations and pretext image transformations. Some recent methods \citep{misra2020self, xiao2020should} have been proposed along these lines to combine domain-specific pretext tasks with contrastive learning to improve the quality of pretrained representations. Nevertheless, a more fundamental understanding of these pretrained representations is still essential to bridge the gap between the handcrafted-based pretext tasks vs. contrastive based SSL methods.  
\vspace{-4mm}

\section{Ablation studies}
\label{sec:Ablation studies}
\vspace{-2mm}
In this section, we perform the ablation experiments to study the importance of two components of our method: (i) ratio of unlabeled data; and (ii) impact of strong augmentations on student network. We choose to perform these ablation studies on $\alpha = 10\%$ labeled data on BreastPathQ and Camelyon16 datasets due to time constraints. Further, we exclude the Kather Multiclass dataset, as it was used to evaluate the feature transferability between datasets, thus making it less suitable for this extensive study.
\vspace{-2mm}

\subsection{Impact of ratio of unlabeled data}
\label{ssec:Impact of ratio of unlabeled data}
\begin{table}[t]
\centering
\caption{Impact of the ratio of unlabeled data ($\mu$). These experiments were performed with $\alpha = 10\%$ labeled data. Note: the intra-class correlation (ICC) coefficient is evaluated between two pathologists A and B for BreastPathQ.}
\label{tab:Impact of ratio of unlabeled data}
\resizebox{0.37\textwidth}{!}{
\begin{tabular}{@{}ccccccccc@{}}
\toprule
\multirow{2}{*}{\shortstack[c]{\textbf{Ratio of} \\ \textbf{Unlabeled Data} \\ ($\mu$)}} & \multicolumn{2}{c}{BreastPathQ} & \multicolumn{2}{c}{Camelyon16} \\ \cmidrule(lr){2-3}  \cmidrule(lr){4-5}  
 & \multicolumn{1}{c}{ICC (\textit{$P_A$})} & \multicolumn{1}{c}{ICC (\textit{$P_B$})} & \multicolumn{1}{c}{AUC} & \multicolumn{1}{c}{Acc} \\ \midrule \midrule
1 & 0.871 & 0.851 & 0.738 & 0.904 \\ 
2 & 0.871 & 0.851 & 0.785 & 0.903 \\ 
3 & 0.876 & 0.846 & 0.797 & 0.907 \\ 
4 & 0.876 & 0.856 & 0.803 & 0.911 \\  
5 & 0.880 & 0.861 & 0.810 & 0.914  \\ 
6 & \textbf{0.882} & \textbf{0.862} & 0.853 & 0.907  \\
7 & 0.876 & 0.846 & \textbf{0.855} & \textbf{0.907} \\ \bottomrule[0.75pt]
\end{tabular}}
\vspace{-4mm}
\end{table}
The success of consistency training is mainly attributed to the amount of unlabeled data. From Table \ref{tab:Impact of ratio of unlabeled data}, we observe marginal to a noticeable improvement in performance, as we increase the ratio of unlabeled ($\mu B$) to labeled batch size ($B$). This is consistent with the recent studies in \cite{xie2019unsupervised} and \cite{sohn2020fixmatch}. For each fold increase in the ratio between unlabeled and labeled samples, the performance improves by at least 2\% on BreastPathQ and Camelyon16. However, the performance in BreastPathQ is quite negligible since the number of training samples ($2063$ patches) is substantially less than Camelyon16 ($306$K patches). On the other hand, increasing the ratio of unlabeled data while fine-tuning the pretrained model converges faster than training the model from scratch. In essence, a large amount of unlabeled data is always beneficial for better performance during consistency training.
\vspace{-1mm}

\subsection{Impact of strong augmentations on student network}
\label{ssec:Impact of augmentation}
\begin{table}[t]
\centering
\caption{Impact of strong augmentation policies applied to the student network. These number of possible transformations are applied sequentially by randomly selecting them in each mini-batch.}
\label{tab:Impact of augmentation}
\resizebox{0.37\textwidth}{!}{
\begin{tabular}{@{}ccccccccc@{}}
\toprule
\multirow{2}{*}{\shortstack[c]{\textbf{No of Possible} \\ \textbf{Transformations} \\ $(N_{Aug})$}} & \multicolumn{2}{c}{BreastPathQ} & \multicolumn{2}{c}{Camelyon16} \\ \cmidrule(lr){2-3}  \cmidrule(lr){4-5}
 & \multicolumn{1}{c}{ICC (\textit{$P_{A}$})} & \multicolumn{1}{c}{ICC\textit{($P_{B}$)}} & \multicolumn{1}{c}{AUC} & \multicolumn{1}{c}{Acc} \\ \midrule \midrule
1 & 0.883 & 0.863 & 0.569 & 0.895 \\ 
2 & \textbf{0.883} & \textbf{0.863} & 0.699 & 0.898 \\ 
3 & 0.882 & 0.861 & 0.742 & 0.899 \\ 
4 & 0.878 & 0.859 & 0.772 & 0.903 \\ 
5 & 0.881 & 0.861 & 0.802 & 0.901  \\
6 & 0.880 & 0.861 & 0.798 & 0.903 \\ 
7 & 0.876 & 0.846 & \textbf{0.855} & \textbf{0.907} \\ \bottomrule[0.75pt]
\end{tabular}}
\vspace{-4mm}
\end{table}
The success of teacher-student consistency training is crucially dependent on the different strong augmentation policies applied to the student network. Table \ref{tab:Impact of augmentation} depicts the analysis of the impact of augmentation policies on final performance. In our experiments, we apply each of these augmentations in sequence by randomly selecting them in each mini-batch using the RandAugment \citep{Cubuk2020} technique. We vary the total number of augmentations ($N_{Aug}$) from value 1 to 7 and examine the effect of strong augmentation policies (applied to the student network) during consistency training. From Table \ref{tab:Impact of augmentation}, we observe that as we gradually increase the severity of augmentation policies in the student model, there are marginal to noticeable improvements in the performance gain. This improvement is mainly visible when trained on large amounts of unlabeled data (such as Camelyon16), where there is a minimum $3\%$ improvement in AUC as we increase the augmentation strength. This suggests that adding strong augmentations to the student network is essential to avoid the model being learned just the teacher's knowledge and gain further improvements in task-specific performance.
\vspace{-2mm}

\section{Discussions}
\label{sec:Discussions}
\vspace{-1mm}
With the advancements in deep learning techniques, current histopathology image analysis methods have shown excellent human-level performance on various tasks such as tumor detection \citep{campanella2019clinical}, cancer grading \citep{bulten2020automated}, and survival prediction \citep{wulczyn2020deep}, etc. However, to achieve these satisfactory results, these methods require a large amount of labeled data for training. Acquiring such massive annotations is laborious and tedious in clinical practice. Thus, there is a great potential to explore self/semi-supervised approaches that can alleviate the annotation burden by effectively exploiting the unlabeled data. 

Drawing on this spirit, in this work, we propose a self-supervised driven consistency training method for histopathology image analysis by leveraging the unlabeled data in both a task-agnostic and task-specific manner. We first formulate the self-supervised pretraining as the resolution sequence prediction task that learns meaningful visual representations across multiple resolutions in WSI. Next, a teacher-student consistency training is employed to improve the task-specific performance based on prediction consistency with the unlabeled data.  Our method is validated on three histology datasets, i.e., BreastPathQ, Camelyon16, and Kather Multiclass, in which our method consistently outperforms other self-supervised methods and also with the supervised baseline under a limited-label regime. Our method has also proven efficacy in transferring pretrained features across different datasets with different tissue types/organs and resolution protocols.

Our work differs from previous consistency based methods in several aspects. Earlier works on consistency training \citep{sohn2020fixmatch, xie2019unsupervised, tarvainen2017mean, liu2020semi, li2020transformation} mainly focused on improving the quality of consistency targets (pseudo labels) by using either of the two strategies: \textbf{i)} careful selection of domain-specific data augmentations; or \textbf{ii)} selection of better teacher model rather than the simple replication of student network. However, there exist some limitations with the above approaches: First, the predicted pseudo labels for the unlabeled data might be incorrect since the model itself is used to generate them. Suppose, if a higher weight is assigned to these, the quality of learning may be hampered due to misclassification, and the model may suffer from confirmation bias \citep{arazo2020pseudo}. Second, instead of using a converged model (such as pretrained) to generate pseudo labels with high confidence scores, the models are trained from scratch leading to lower accuracy and high entropy. In this work, we overcome these limitations by providing a solution that leverages the advantage of both the above solutions in a simple, efficient manner. The main key difference between our approach and the other existing consistency training methods are two-fold: \textbf{i)} we make use of the task-specific fine-tuned model to generate high-confidence (i.e., low-entropy) consistency targets instead of relying on the model being trained; \textbf{ii)} we experimentally show that by aggressively injecting noise through various domain-specific data augmentations, the student model is forced to work harder to maintain consistency with the pseudo label produced by teacher model. This ensures that the student network does not merely replicate the teacher's knowledge. 

Despite the excellent performance of our method, there is one main limitation: \textit{i.e.,} if the pseudo labels produced by the teacher network are inaccurate, then the student network is forced to learn from incorrect labels - leading to confirmation bias \citep{arazo2020pseudo}. As a result, the student may not become better than the teacher during consistency training. We solved this issue with RandAugment \citep{Cubuk2020}, a strong data augmentation technique, which we combine with label smoothing (soft pseudo labels) to deal with confirmation bias. This is also consistent with the recent study \citep{arazo2020pseudo} that showed soft pseudo labels outperform hard pseudo labels when dealing with label noise. However, the bias issue still persists with soft pseudo labels in our application. This is prominently visible in our method, where, compared to self-supervised pretraining (see Fig. \ref{Fig:Heat-maps on Camelyon16}, column (c) - (f); Fig. \ref{Fig:Heat-maps on BreastPathQ}, column (b) - (e)), the consistency trained approaches (see Fig. \ref{Fig:Heat-maps on Camelyon16}, column (g) - (j); Fig. \ref{Fig:Heat-maps on BreastPathQ}, column (f) - (i)) exhibits some low probability ($<0.5$) spurious pixels outside the malignant cell boundaries. This happens because of the naive pseudo labeling produced by the teacher network, which sometimes overfits to incorrect pseudo labels. Further, this issue is reinforced when we attempt to train the student network on unlabeled samples with incorrect pseudo labels leading to confirmation bias. One solution to mitigate this issue is to make the teacher network constantly adapt to the feedback of the student model instead of the teacher model being fixed. This technique has been shown to work well in a recent meta pseudo label method \citep{pham2020meta}, where both teacher and student are trained in parallel by making the teacher learn from the reward signal of the student performance on a labeled set. Exploring this idea is beyond the scope of this work, and we will leave this to the practitioner to explore more along these lines.     

In general, our proposed self-supervised driven consistency training framework has a great potential to solve the majority of both classification and regression tasks in computational histopathology, where annotation scarcity is a significant issue. Further, our pretrained representations are more generic and easily extended to other downstream multi-tasks, such as segmentation and survival prediction. It is worth further investigating to develop a universal feature encoder in histopathology that can solve many tasks without excessive labeled annotations.
\vspace{-3mm}

\section{Conclusion}
\label{sec:Conclusion}
\vspace{-1mm}
In this paper, we present an annotation efficient framework by introducing a novel self-supervised driven consistency training paradigm for histopathology image analysis. The proposed framework utilizes the unlabeled data both in a task-agnostic and task-specific manner to significantly advance the accuracy and robustness of the state-of-the-art self-supervised (SSL) methods. To this end, we first propose a novel task-agnostic self-supervised pretext task by efficiently harnessing the multi-resolution contextual cues present in the histology whole-slide images. We further develop a task-specific teacher-student semi-supervised consistency paradigm to effectively distill the SSL pretrained representations to downstream tasks. This synergistic harness of unlabeled data has been shown to significantly improve the SSL pretrained performance over its supervised baseline under a limited-label regime.

Extensive experiments on three public benchmark datasets across two classification and one regression based histopathology tasks, \textit{i.e.,} tumor metastasis detection, tissue type classification, and tumor cellularity quantification demonstrate the effectiveness of our proposed approach. Our experiments also showed that our method's performance is significantly outperforming or even comparable to that of the supervised baseline when trained under limited annotation settings. Furthermore, our approach is more generic and has been shown to generate universal pretrained representations that can be easily adapted to other histopathology tasks and also to other domains without any modifications. 

\vspace{2mm}
\begin{flushleft}
\textbf{Conflict of interest} 
\end{flushleft} 
ALM is co-founder and CSO of Pathcore. CS, SK and FC have no financial or non-financial conflict of interests.

\begin{flushleft}
\textbf{Acknowledgment} 
\end{flushleft} 
This work was funded by Canadian Cancer Society and Canadian Institutes of Health Research (CIHR). It was also enabled in part by support provided by Compute Canada (www.computecanada.ca).

\appendix
\section{Supplementary material}
\label{sec:appendix}

\begin{itemize}
    \item Fig. \ref{Fig:Heat-maps on BreastPathQ} Tumor cellularity scores produced on WSIs of the BreastPathQ test set for 10\% labeled data.
    \item Fig. \ref{Fig:Heat-maps on Camelyon16} Tumor probability heat-maps overlaid on original WSIs from Camelyon16 test set predicted from 10\% labeled data.
\end{itemize}

\section{Momentum contrast (MoCo)}
\label{sec:Momentum Contrast (MoCo)}
\vspace{-1mm}
Momentum Contrast model (MoCo) \citep{he2020momentum} is one of the most popular self-supervised models that even outperforms supervised baseline models. Given a data point $x$ in a dataset, MoCo samples a positive pair $k_+$ and $N$ negative pairs $k_-^1, ..., k_-^N$. MoCo is trained with infoNCE loss \citep{oord2018representation}, defined as 
\begin{equation}
    L_{infoNCE} = -\mathbb{E}_{x\sim p(x)} \Big[\frac{exp(F_q(x)\cdot F_k(k_+) / \tau)}{\sum^N_{i=1} exp(F_q(x)\cdot F_k(k_-^i) / \tau)}\Big],
\end{equation}
where, $F_q$ and $F_k$ are neural networks, and $\tau$ is a hyperparameter for temperature. This is a log loss of a softmax classifier which  minimizes the difference between the representations $F_q(x)$ and its positive pair $F_k(k_+)$ while maximizing the differences between $F_q(x)$ and negative pairs $F_k(k_-^{1,...,N})$. Note that minimizing $L_{infoNCE}$ maximizes the lower bound for mutual information between $x$ and $k_+$ \citep{oord2018representation}. However, the bound is not tight for a small number of $N$; therefore, in practice, we need to use a large number of negative samples for each iteration. However, as this is not practical for computational efficiency, MoCo maintains a large queue of encoded data. At each training iteration, the entire mini-batch consisting of a positive sample and negative samples are inserted into the queue.  Therefore, we use the entire queue (except the positive sample) as the set of negatives for the infoNCE loss. One of the key observation made by MoCo is that this can be problematic if the encoder $F$ changes too quickly, as this would cause the discrepancy between the distribution of the samples in the queue and the new samples to be quite different, and the classifier can easily decrease the loss. To solve this problem, MoCo uses two networks: the encoder $F_q$ with parameters $\theta_q$ and the momentum encoder $F_k$ with parameters $\theta_k$. $F_k$ is not trained with the infoNCE loss but is updated with momentum parameter $m$:
\begin{equation}
    \theta_k = m~\theta_k + (1-m)~\theta_q,
\end{equation}
after each training iteration. We use the queue size of 8192 and $m$ of 0.999, and adopt multiple augmentation schemes. 
In each training iteration, for each data $x$, we \textit{randomly} i) jitter the brightness, contrast, saturation, and hue by 0.6 $\sim$ 1.4, ii) rotate it by 0 $\sim$ 360 degrees, iii) flip vertically \& horizontally, and iv) crop with an area in the range 0.7 $\sim$ 1 and stretch to the original size.
\vspace{-3mm}

\section{Variational autoencoder (VAE)}
\label{sec:Variational autoencoder}
Variational autoencoder (VAE) \citep{kingma2013auto} is an unsupervised machine learning model that is often used for dimensionality reduction and image generation. The model contains an encoder and a decoder, with a latent space that has a dimension smaller than the input data. The reduction in dimension on the latent space helps extract the prominent information in the original data. Unlike the vanilla autoencoders, VAE assumes that input data comes from some latent distribution $z \sim N(0,I)$. The encoder estimates the mean ($\mu$) and variance ($\sigma^{2}$) of the data in the latent space, and the decoder samples a point from the distribution for data reconstruction. The assumption of $z$ following a normal distribution and the stochastic property of the latent vector force the model to create a continuous latent space with similar data closer in the space. This resolves the model overfitting due to irregularities in the latent space often observed in the conventional autoencoder. The learning rule of VAE is to maximize the evidence lower bound ($ELBO$),
\begin{equation}
      ELBO=E_{z \sim q(z|x)}\log p(x|z)-KL~(q(z|x)~||~p(z)) \leq \log p(x),
    \label{eq::ELBO}
\end{equation}%
where, $q$ is the approximate posterior distribution of $p~(z|x)$. The \textit{first} term describes the reconstruction loss of the autoencoder model. The \textit{second} term is can be seen as a regularizer that forces the approximate latent distribution to be close to $N(0,I)$.  Standard stochastic gradient descent methods cannot directly apply to the model because of the stochastic property of the latent vector. The solution, called the reparameterization trick, is to introduce a new random variable $\epsilon \sim N(0,I)$ as the model input and set the latent vector to $z = \mu(x) + \sigma^{1/2}(x) \cdot \epsilon$. This allows all model parameters to be deterministic. 

For our VAE model, we use a ResNet-18 model to encode input image of size $256 \times 256$ to a latent vector of size 512. Then, we use the generator from the BigGAN model \citep{brock2018large} to reconstruct the latent vector back to the original image.
\vspace{-2mm}

\afterpage{
\clearpage
\begin{landscape}
\thispagestyle{lscape}
\pagestyle{lscape}

\begin{figure}
 	\begin{minipage}{0.1\linewidth}
 	\centerline{\textbf{Image}}\medskip
    \centerline{\includegraphics[width=2.1cm, height=2.1cm]{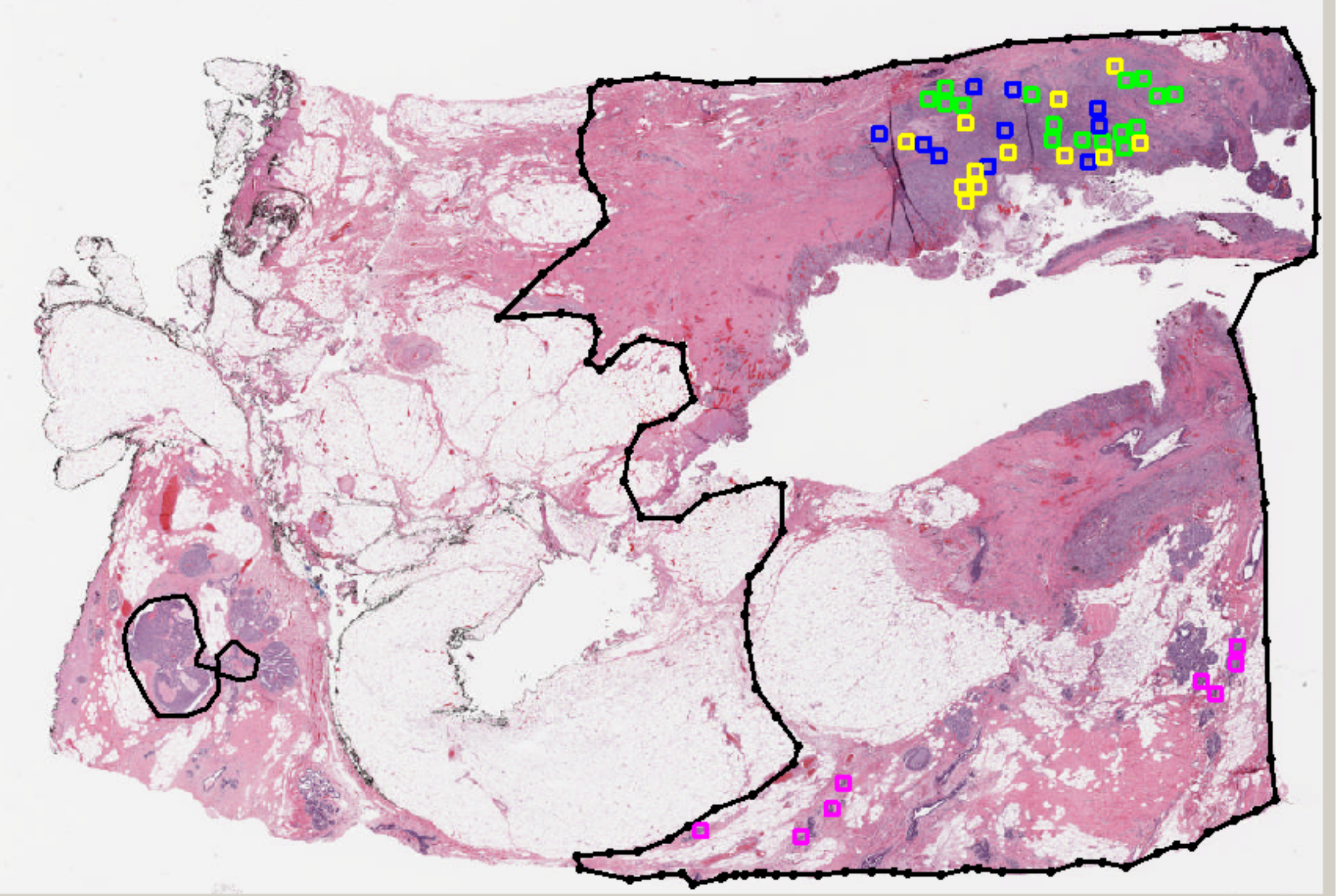}} 
    \end{minipage}
\hfill
	\begin{minipage}{0.1\linewidth}
	\centerline{\textbf{Random}}\medskip
    \centerline{\includegraphics[width=2.1cm, height=2.1cm]{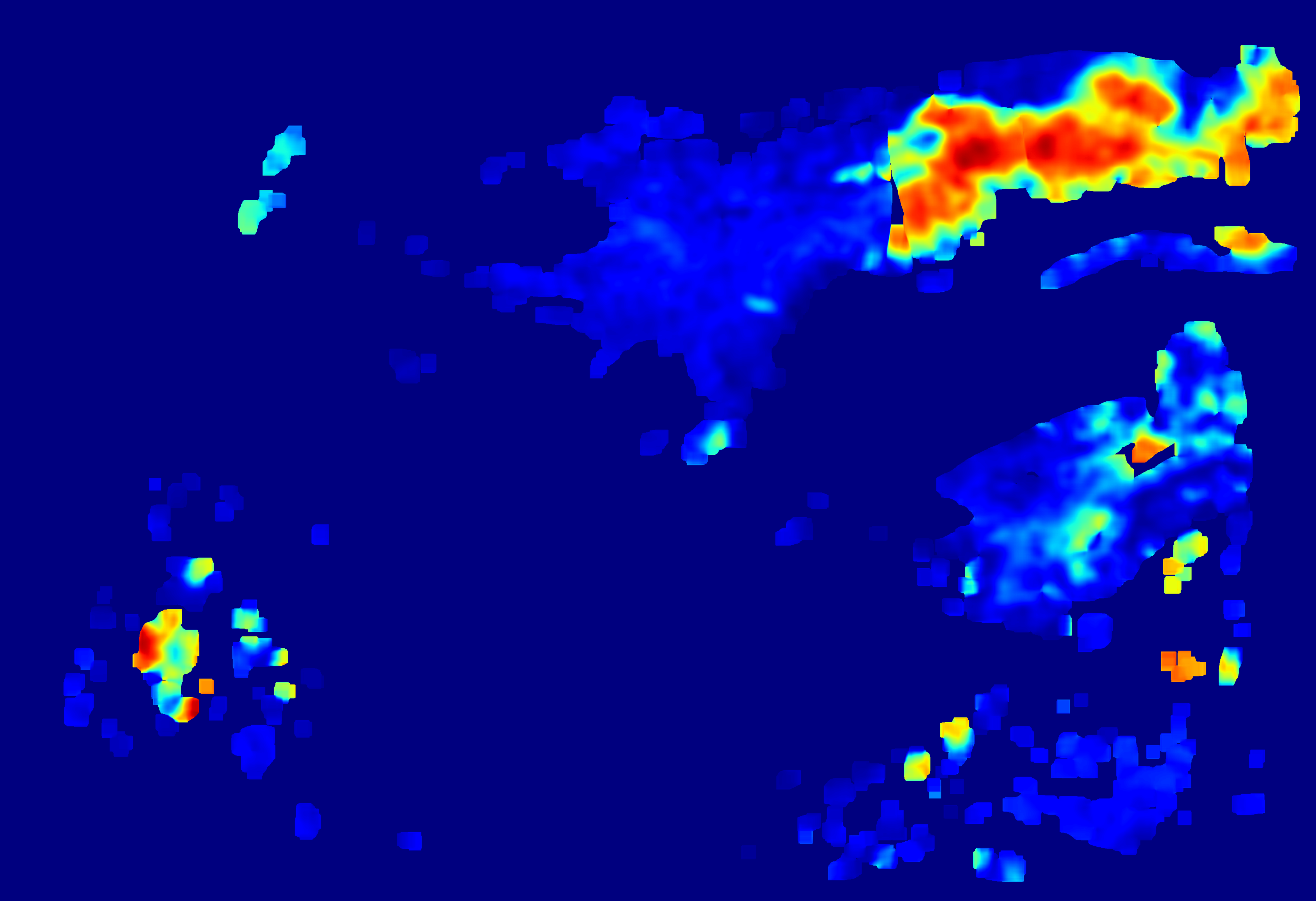}} 
    \end{minipage}
\hfill
	\begin{minipage}{0.1\linewidth}
	\centerline{\textbf{VAE}}\medskip
    \centerline{\includegraphics[width=2.1cm, height=2.1cm]{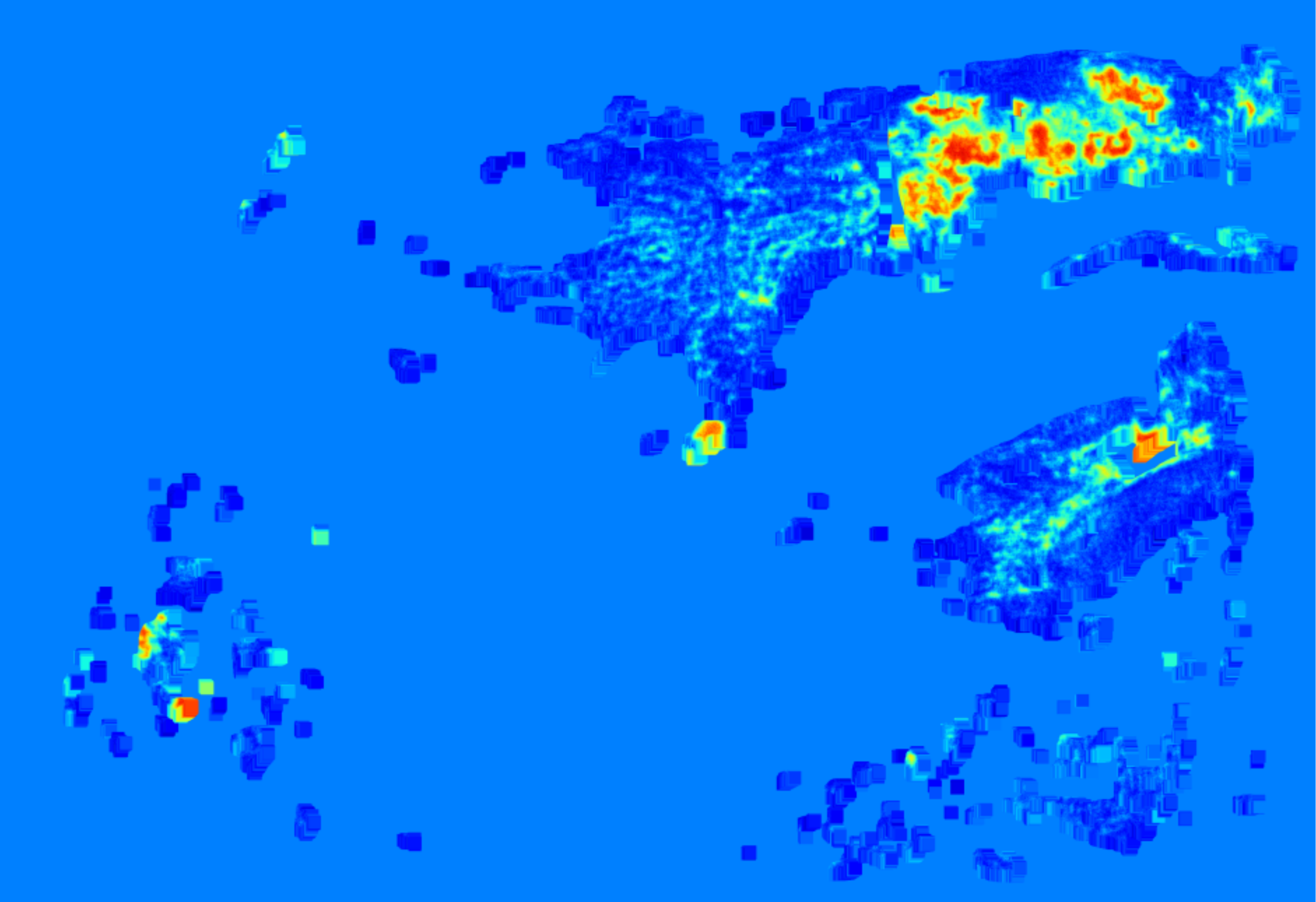}} 
    \end{minipage}
\hfill
	\begin{minipage}{0.1\linewidth}
	\centerline{\textbf{MoCo}}\medskip
    \centerline{\includegraphics[width=2.1cm, height=2.1cm]{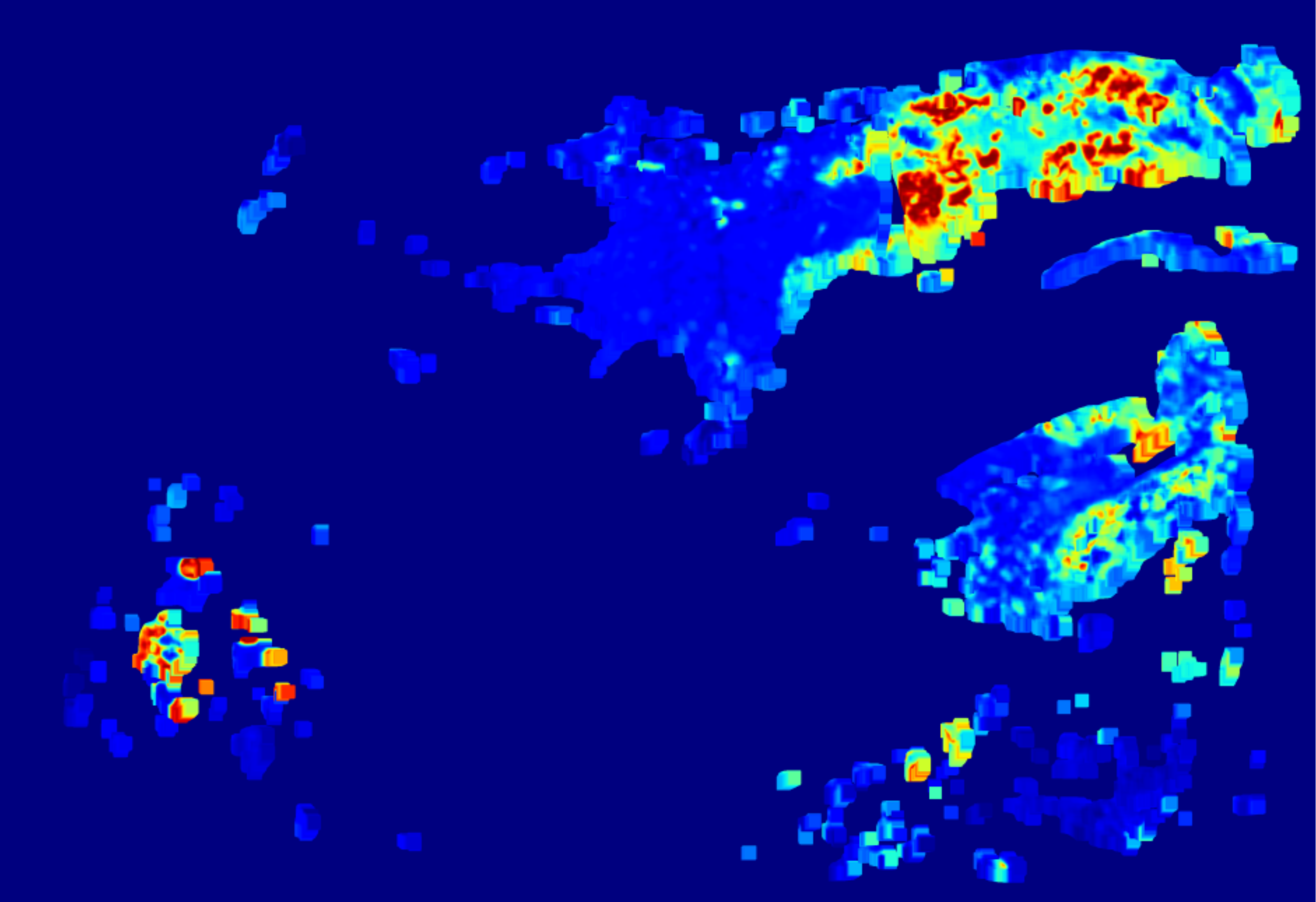}} 
    \end{minipage}
\hfill
	\begin{minipage}{0.1\linewidth}
	\centerline{\textbf{RSP}}\medskip	
    \centerline{\includegraphics[width=2.1cm, height=2.1cm]{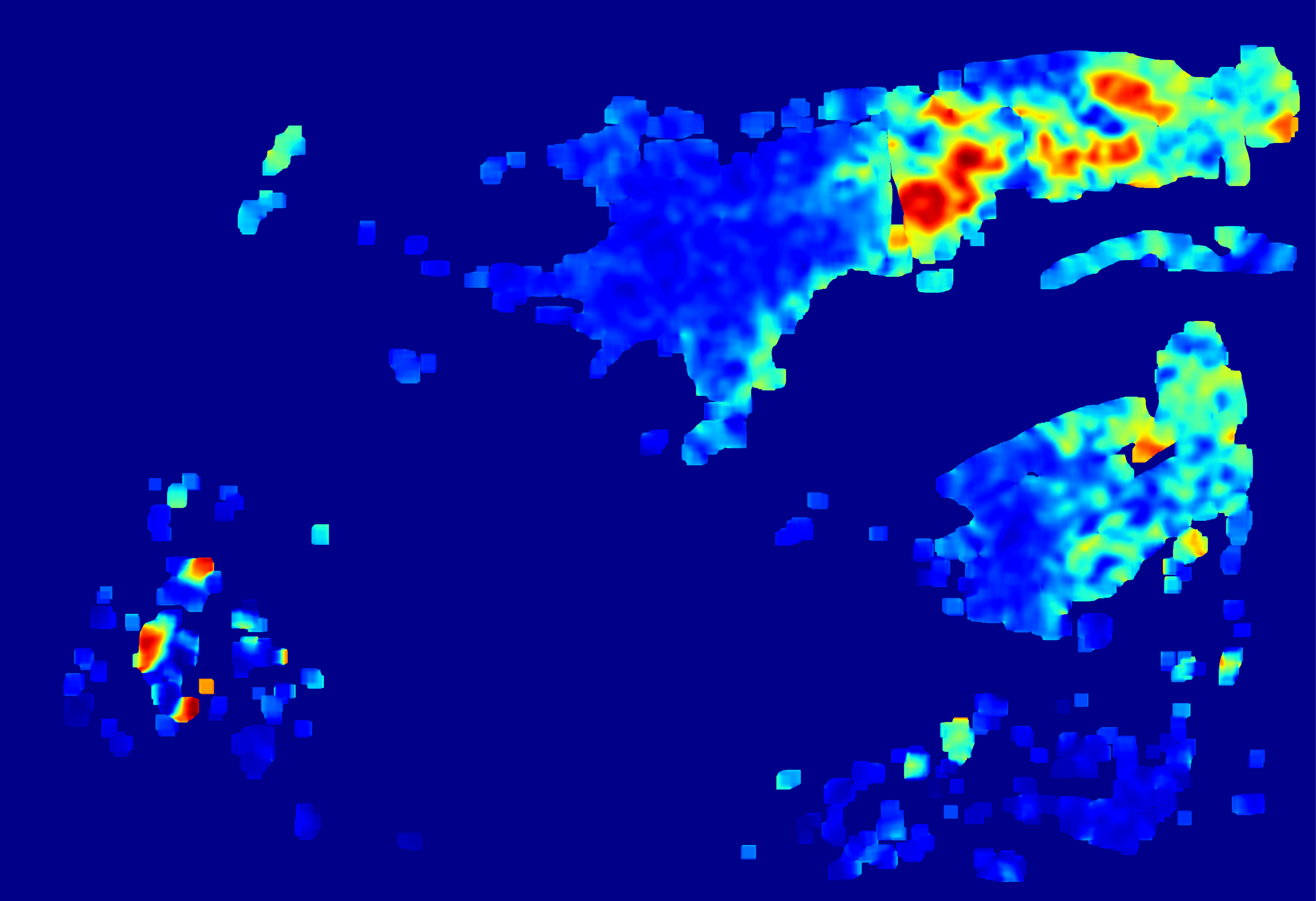}} 
    \end{minipage}    
\hfill
	\begin{minipage}{0.1\linewidth}
	\centerline{\textbf{Random+CR}}\medskip
    \centerline{\includegraphics[width=2.1cm, height=2.1cm]{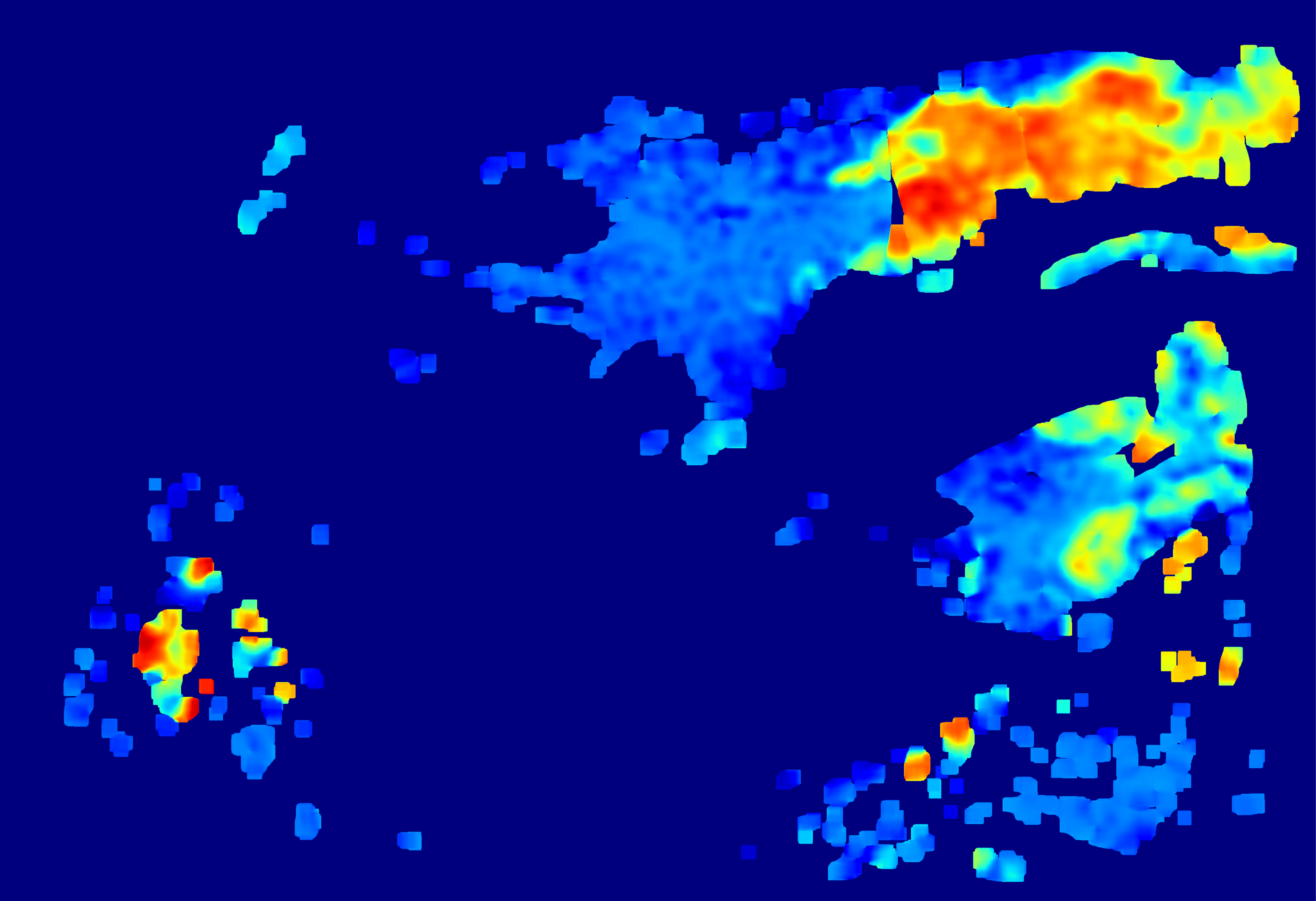}} 
    \end{minipage}
\hfill
	\begin{minipage}{0.1\linewidth}
	\centerline{\textbf{VAE+CR}}\medskip
    \centerline{\includegraphics[width=2.1cm, height=2.1cm]{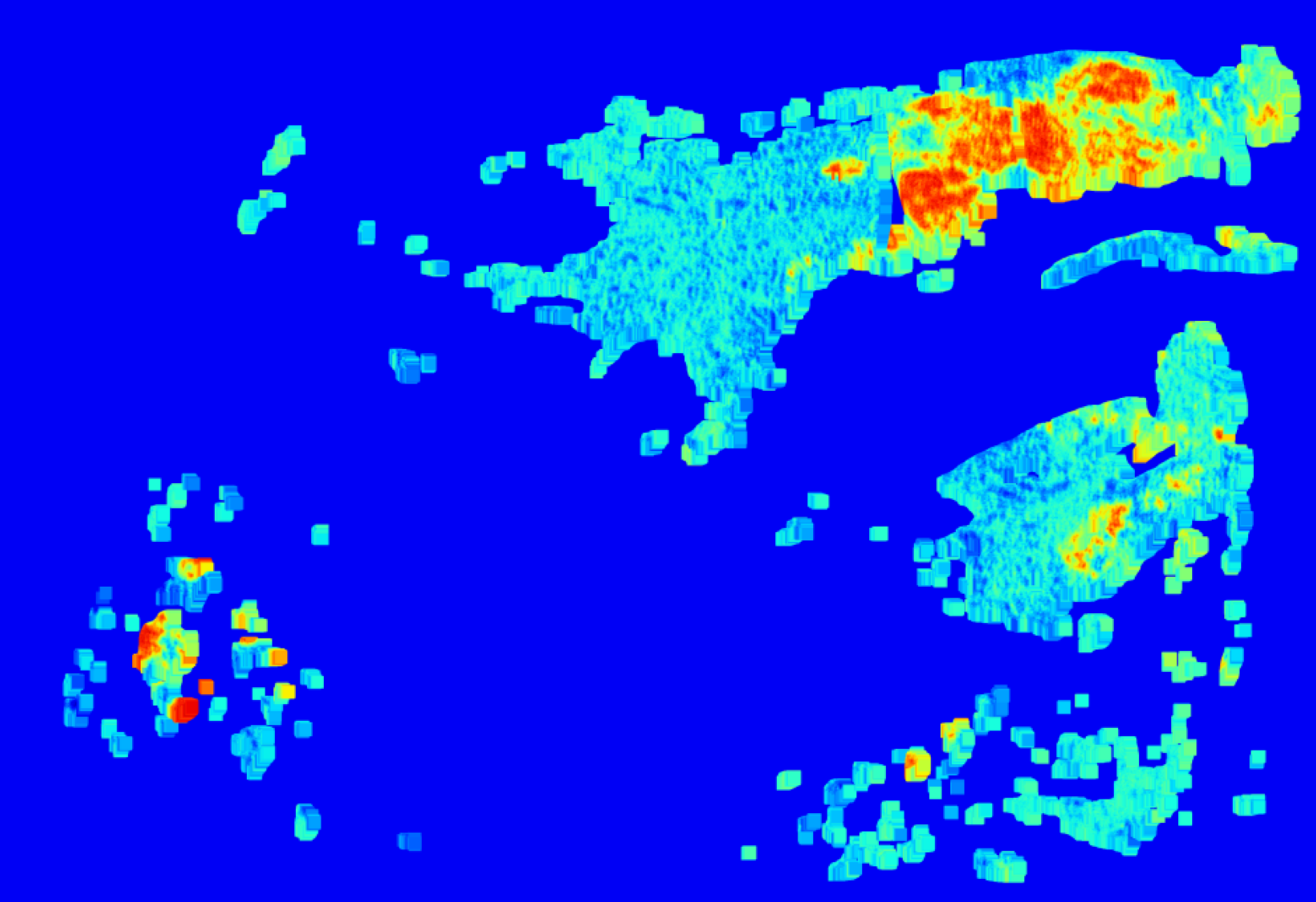}} 
    \end{minipage}
\hfill
	\begin{minipage}{0.1\linewidth}
	\centerline{\textbf{MoCo+CR}}\medskip
    \centerline{\includegraphics[width=2.1cm, height=2.1cm]{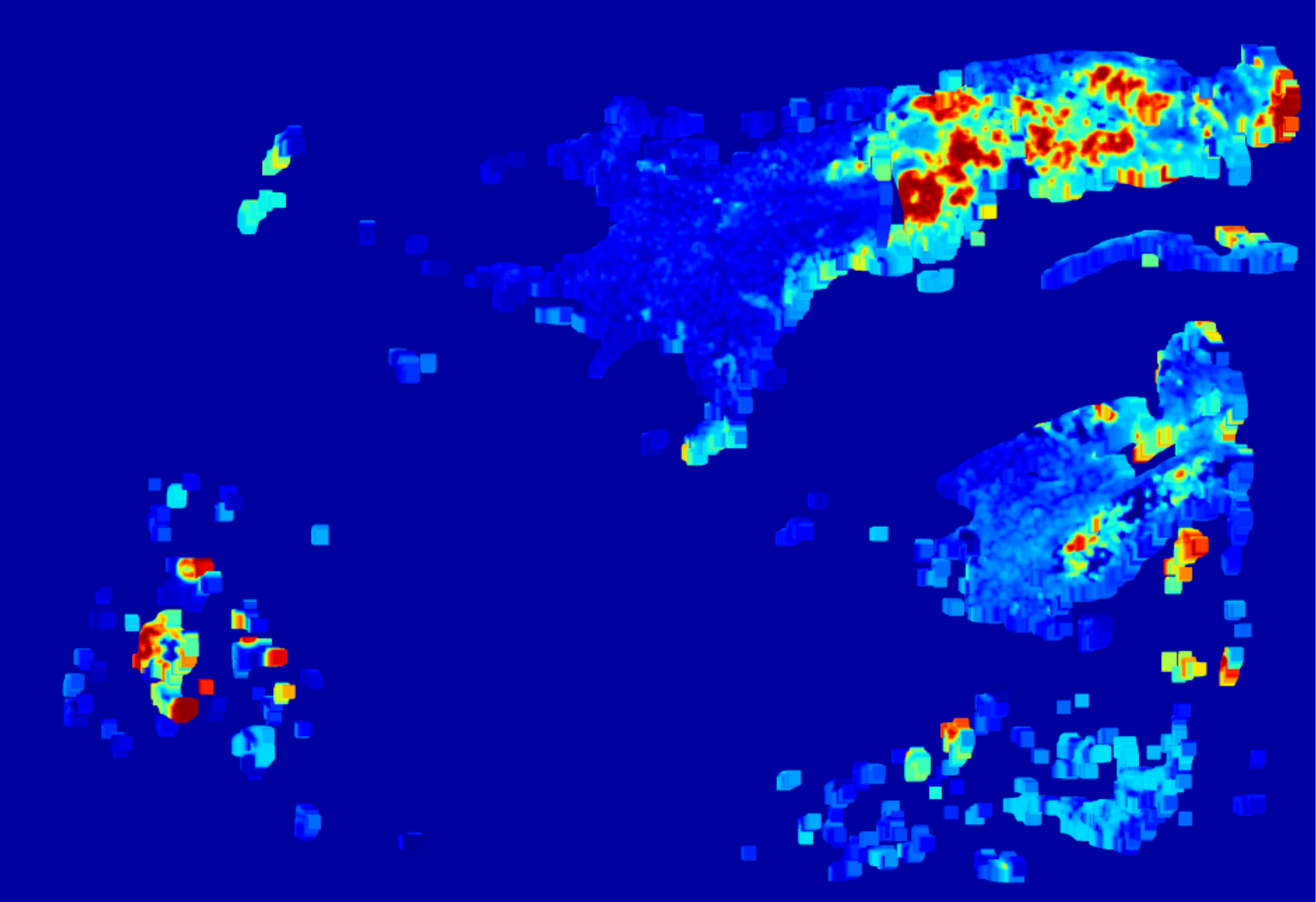}} 
    \end{minipage}
\hfill
	\begin{minipage}{0.1\linewidth}
	\centerline{\textbf{RSP+CR}}\medskip
    \centerline{\includegraphics[width=2.1cm, height=2.1cm]{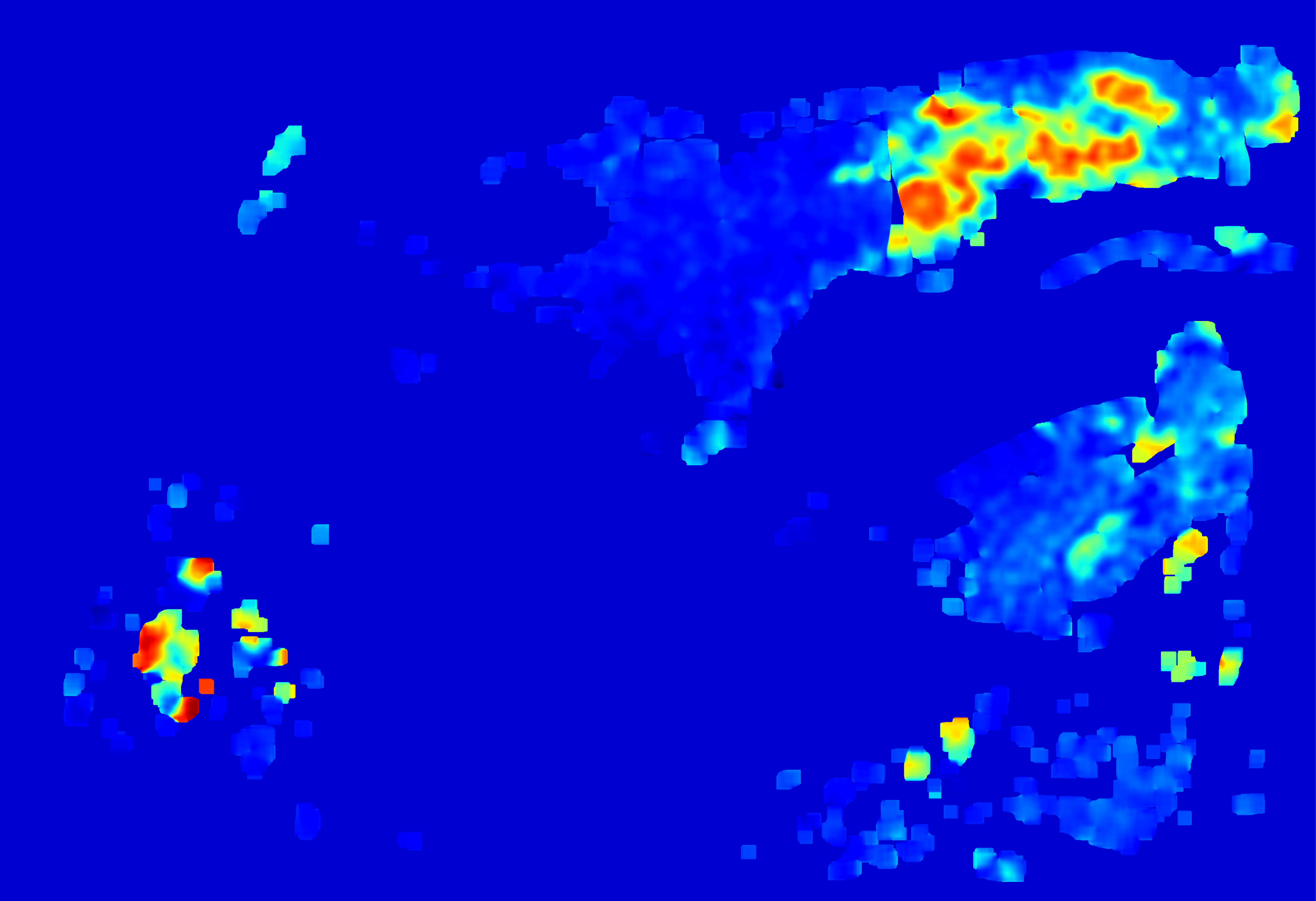}} 
    \end{minipage}   
    
\vfill	

 	\begin{minipage}{0.1\linewidth}
    \centerline{\includegraphics[width=2.1cm, height=2.1cm]{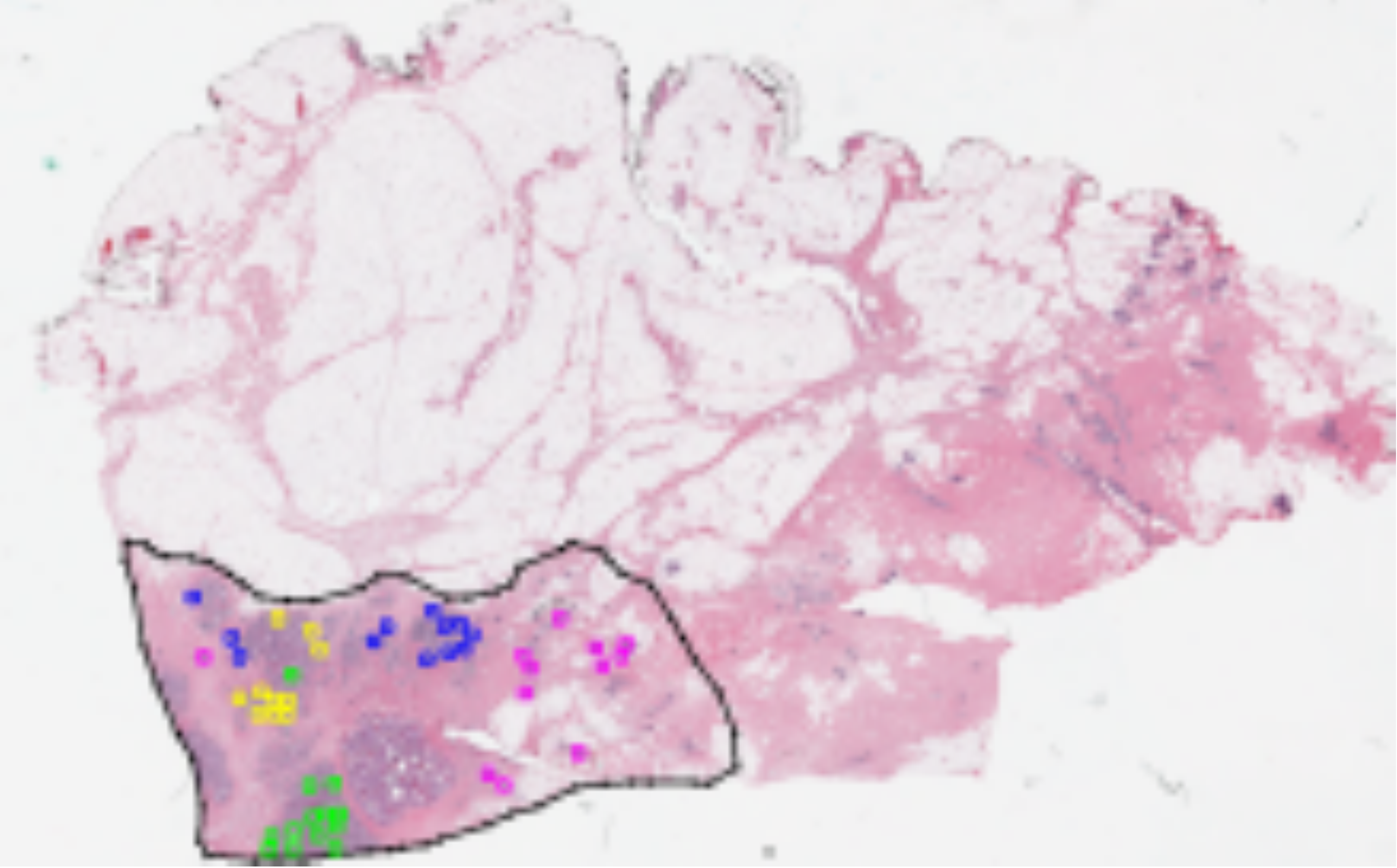}} 
    \end{minipage}
\hfill
	\begin{minipage}{0.1\linewidth}
    \centerline{\includegraphics[width=2.1cm, height=2.1cm]{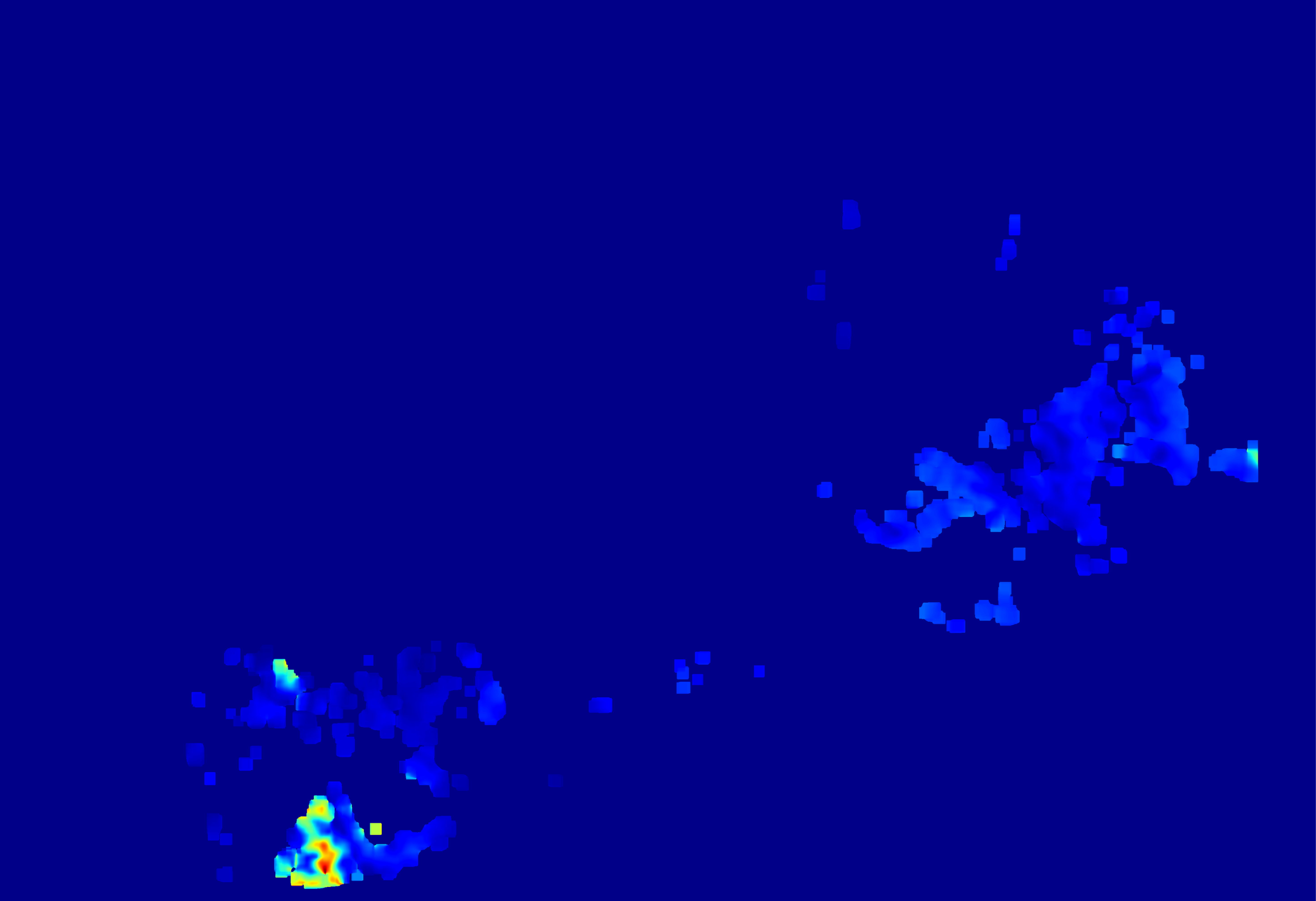}} 
    \end{minipage}
\hfill
	\begin{minipage}{0.1\linewidth}
    \centerline{\includegraphics[width=2.1cm, height=2.1cm]{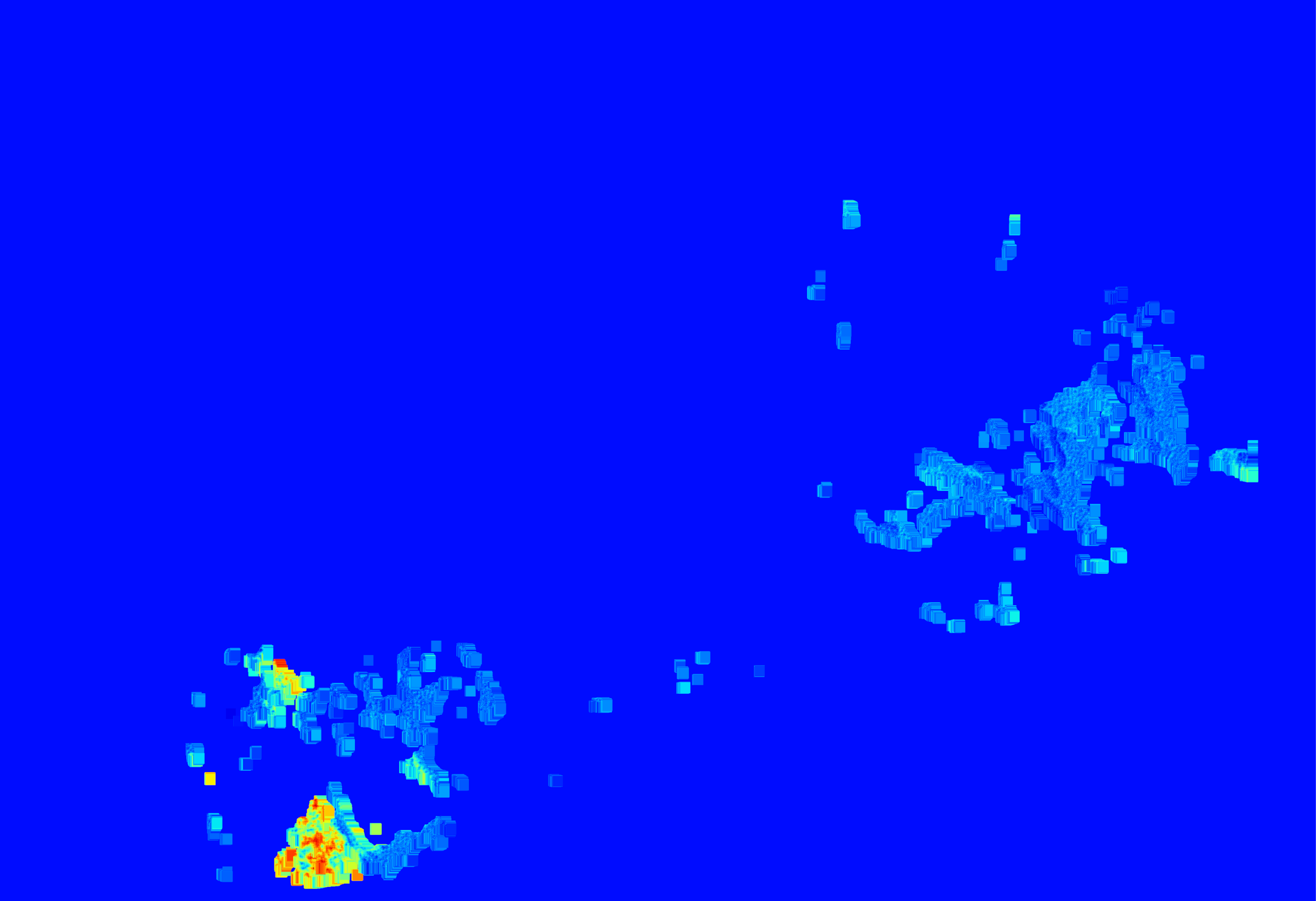}} 
    \end{minipage}
\hfill
	\begin{minipage}{0.1\linewidth}
    \centerline{\includegraphics[width=2.1cm, height=2.1cm]{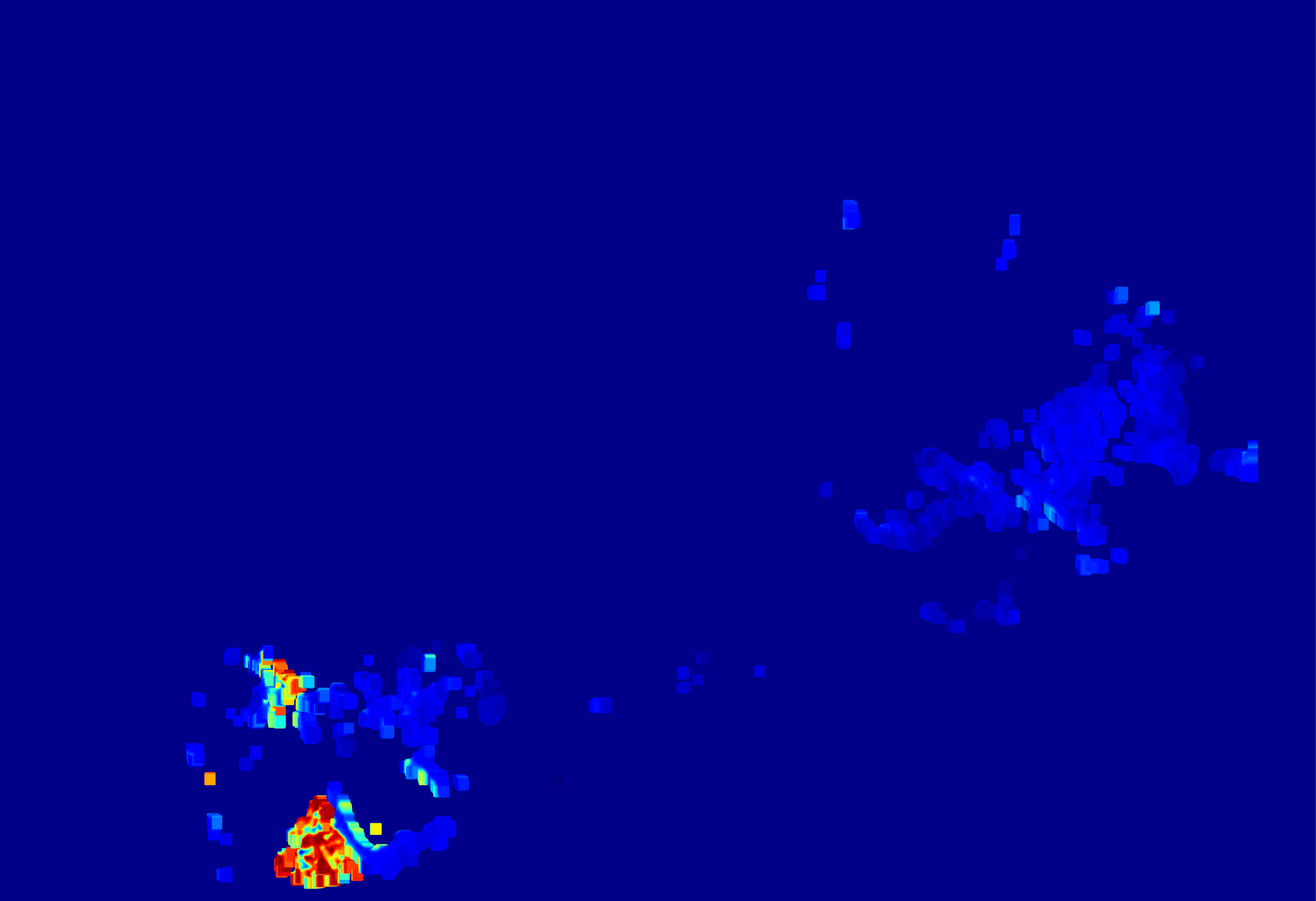}} 
    \end{minipage}
\hfill
	\begin{minipage}{0.1\linewidth}
    \centerline{\includegraphics[width=2.1cm, height=2.1cm]{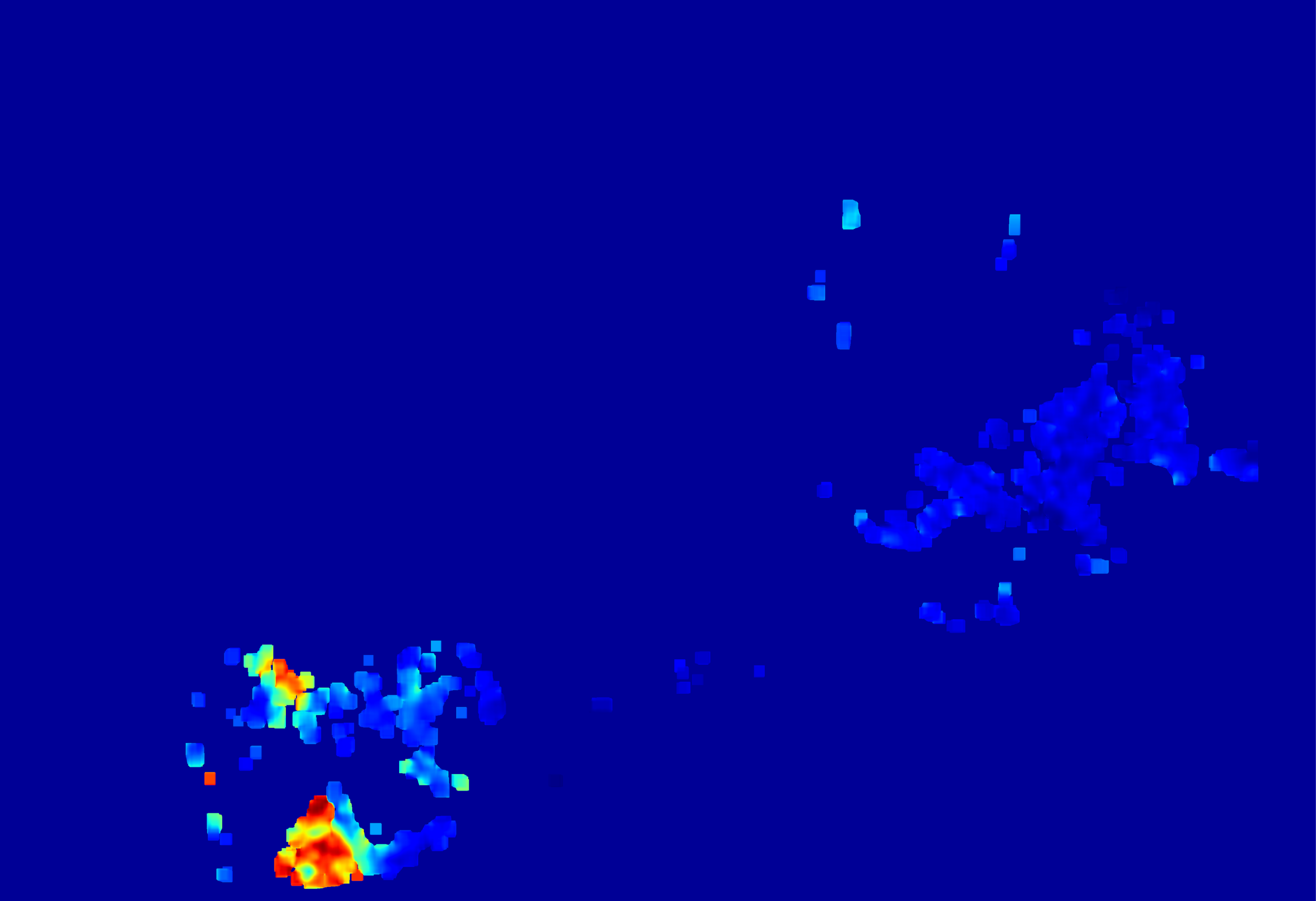}} 
    \end{minipage}    
\hfill
	\begin{minipage}{0.1\linewidth}
    \centerline{\includegraphics[width=2.1cm, height=2.1cm]{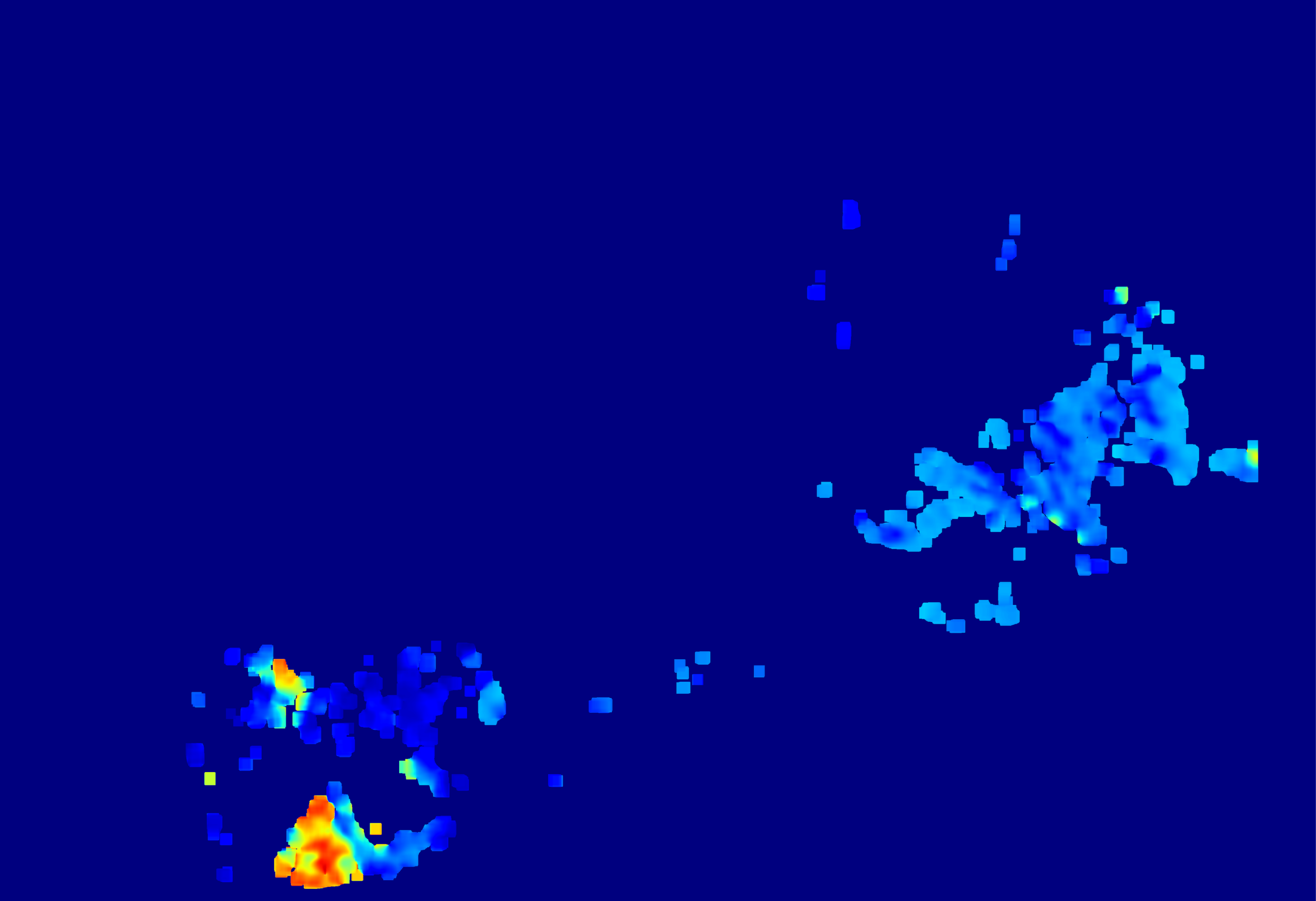}} 
    \end{minipage}
\hfill
	\begin{minipage}{0.1\linewidth}
    \centerline{\includegraphics[width=2.1cm, height=2.1cm]{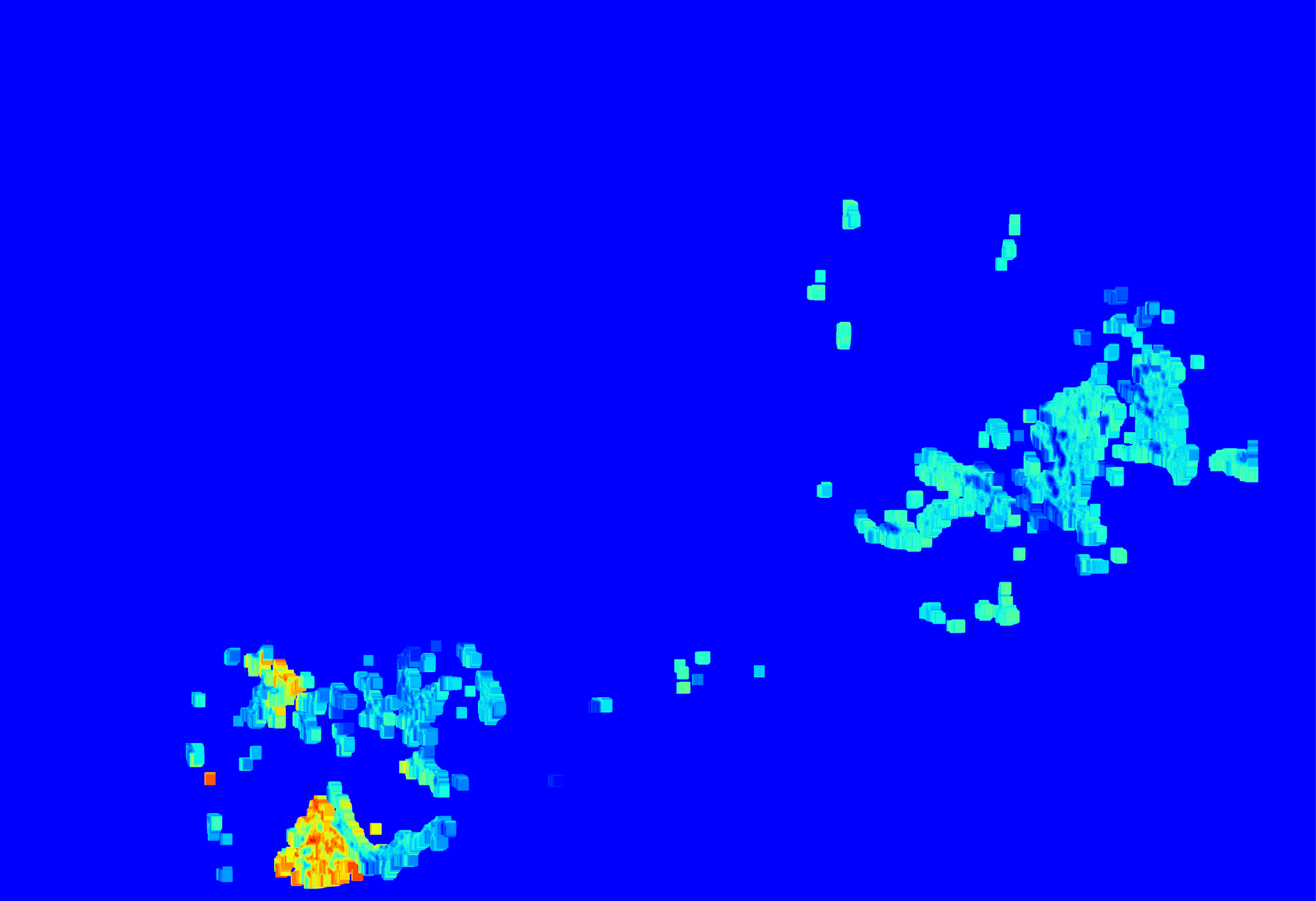}} 
    \end{minipage}
\hfill
	\begin{minipage}{0.1\linewidth}
    \centerline{\includegraphics[width=2.1cm, height=2.1cm]{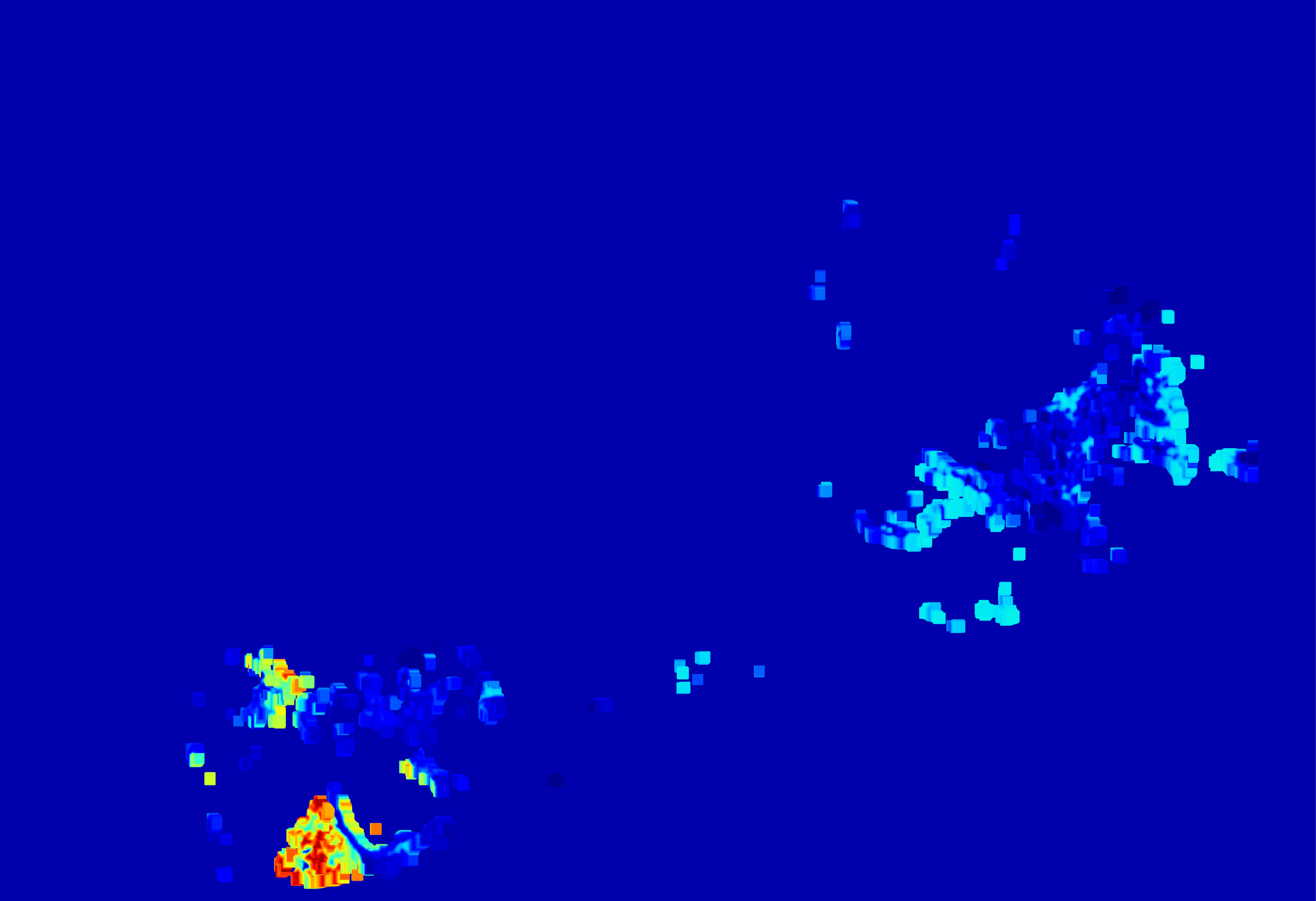}} 
    \end{minipage}
\hfill
	\begin{minipage}{0.1\linewidth}
    \centerline{\includegraphics[width=2.1cm, height=2.1cm]{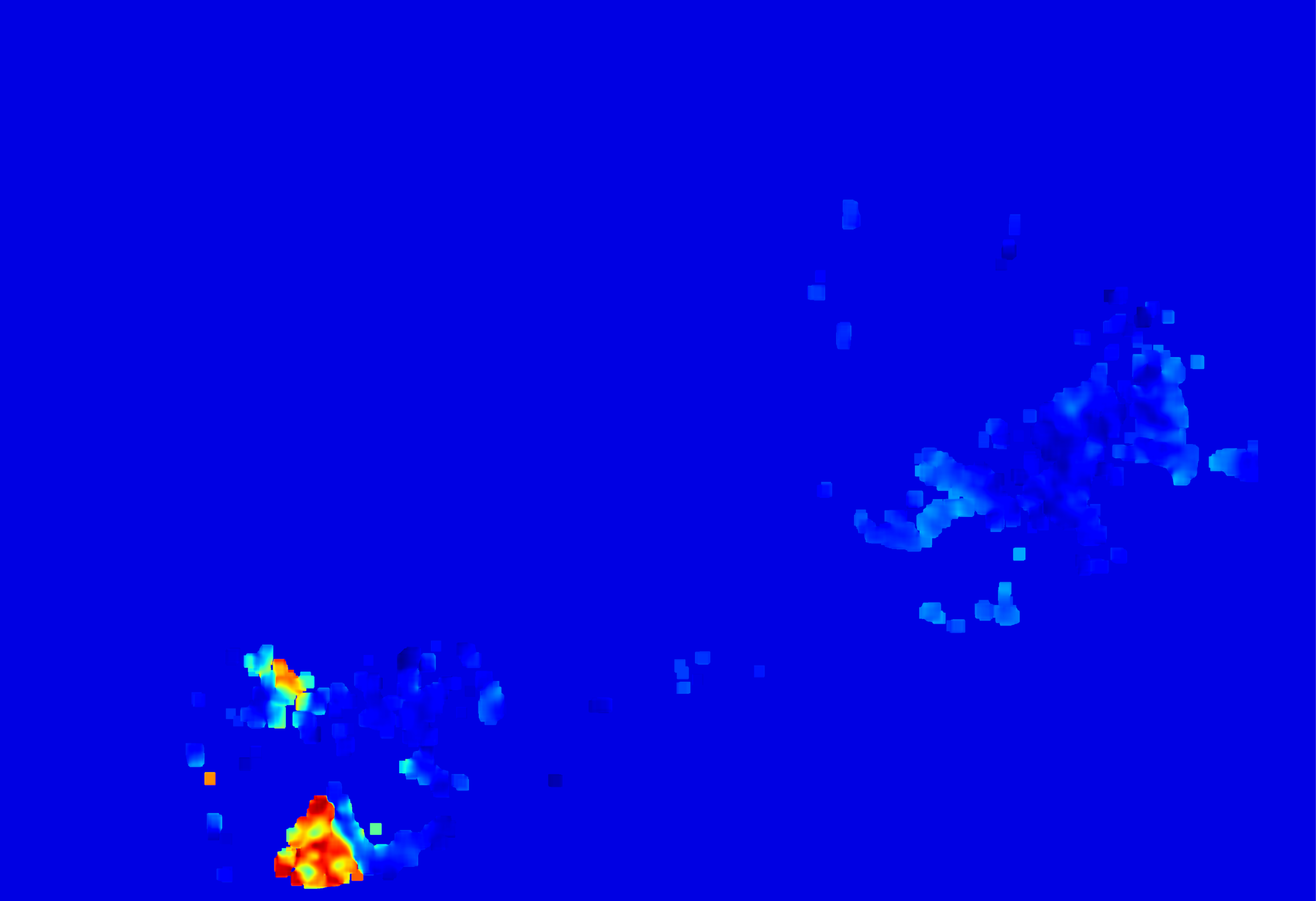}} 
    \end{minipage}

\vfill

 	\begin{minipage}{0.1\linewidth}
    \centerline{\includegraphics[width=2.1cm, height=2.1cm]{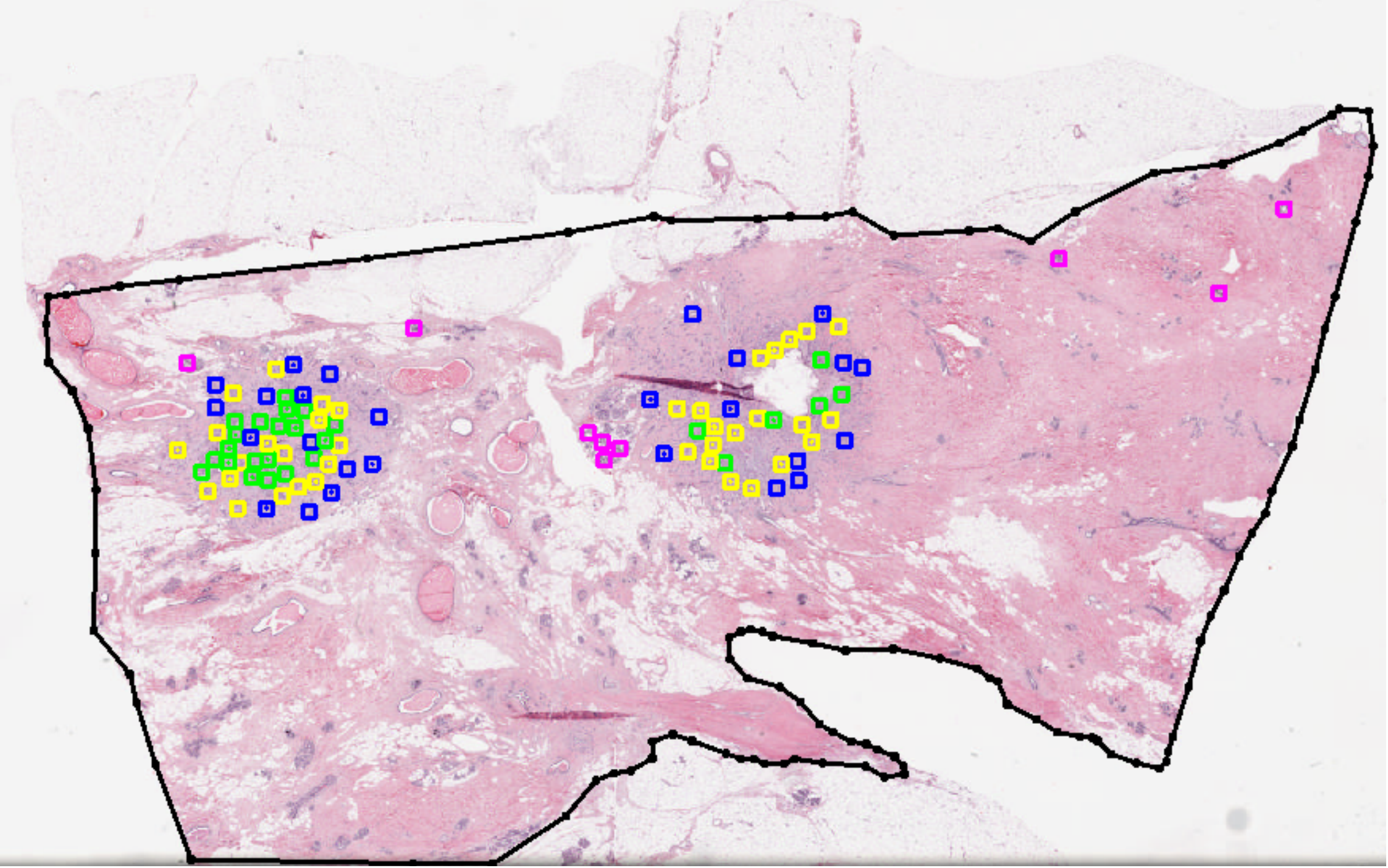}} 
    \centerline{(a)}\medskip
    \end{minipage}
\hfill
	\begin{minipage}{0.1\linewidth}
    \centerline{\includegraphics[width=2.1cm, height=2.1cm]{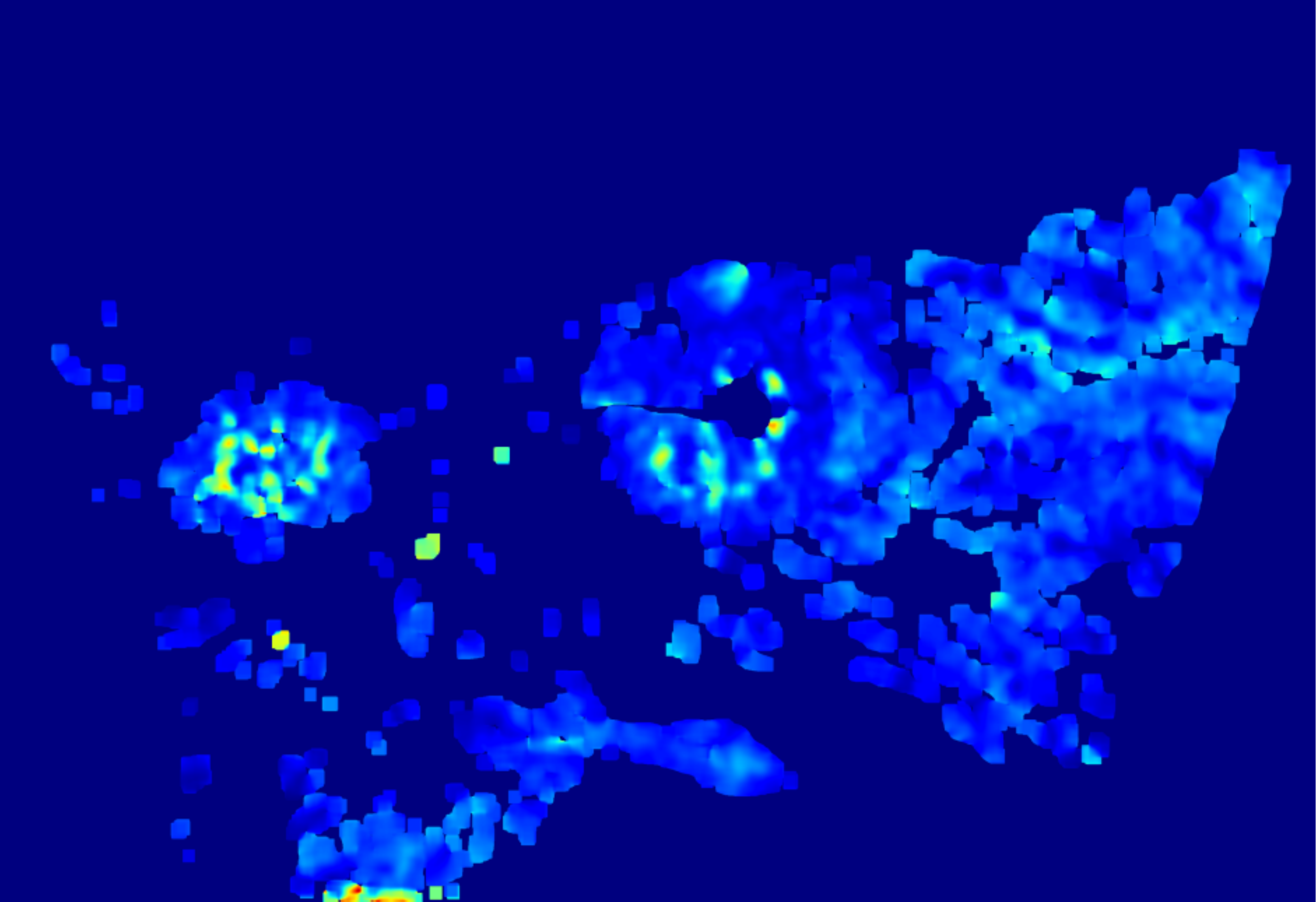}} 
    \centerline{(b)}\medskip
    \end{minipage}
\hfill
	\begin{minipage}{0.1\linewidth}
    \centerline{\includegraphics[width=2.1cm, height=2.1cm]{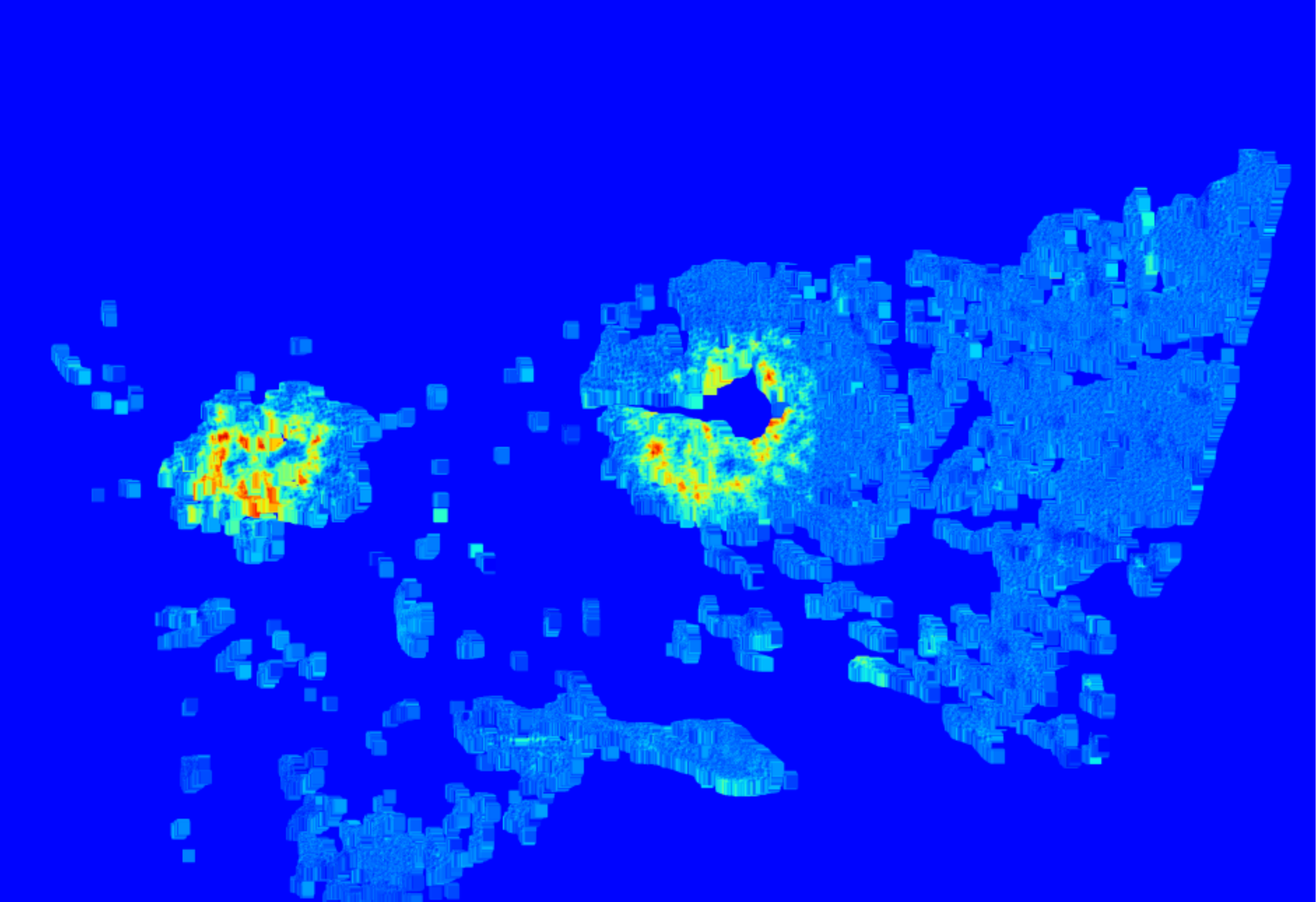}} 
    \centerline{(c)}\medskip
    \end{minipage}
\hfill
	\begin{minipage}{0.1\linewidth}
    \centerline{\includegraphics[width=2.1cm, height=2.1cm]{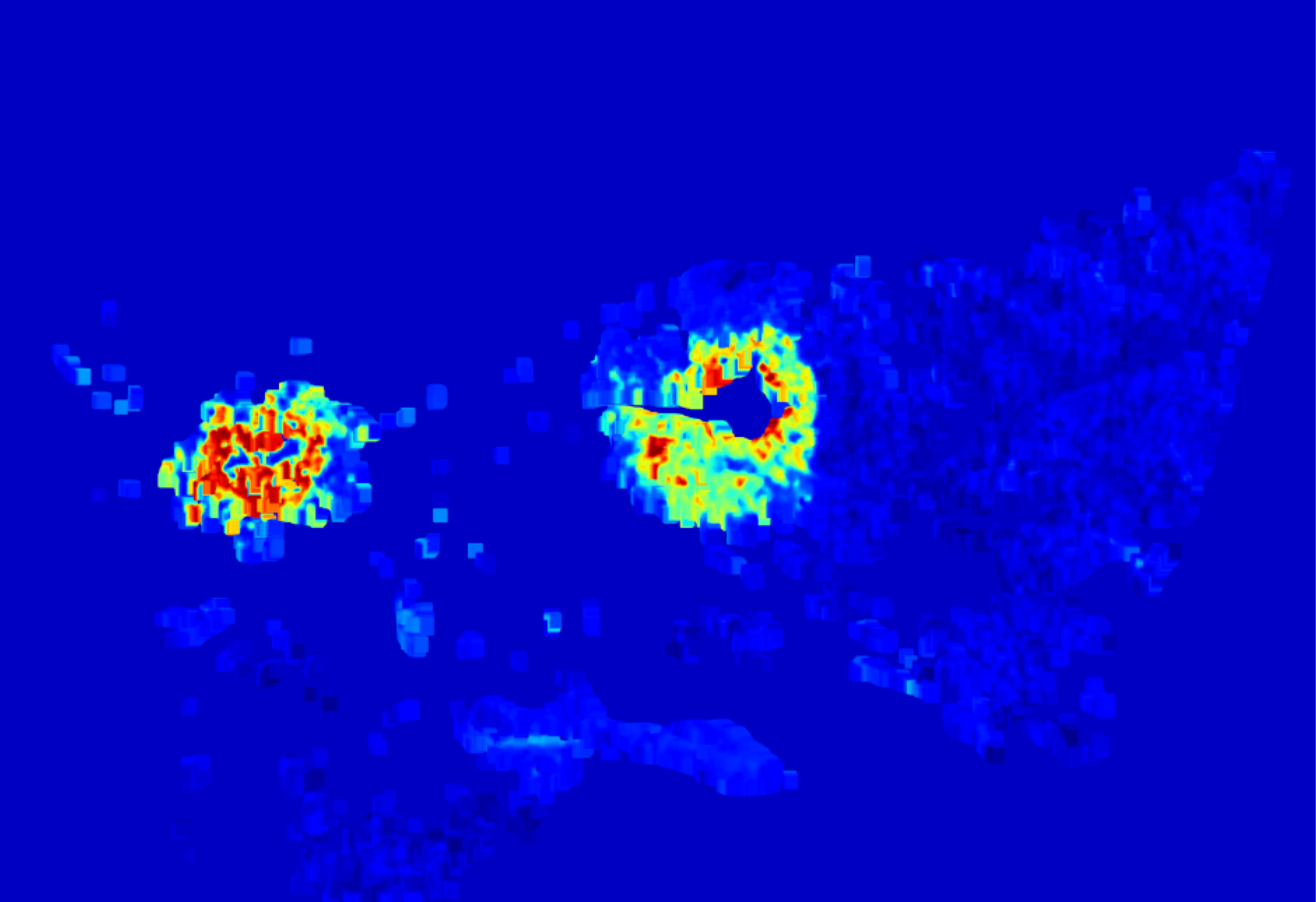}} 
    \centerline{(d)}\medskip
    \end{minipage}
\hfill
	\begin{minipage}{0.1\linewidth}
    \centerline{\includegraphics[width=2.1cm, height=2.1cm]{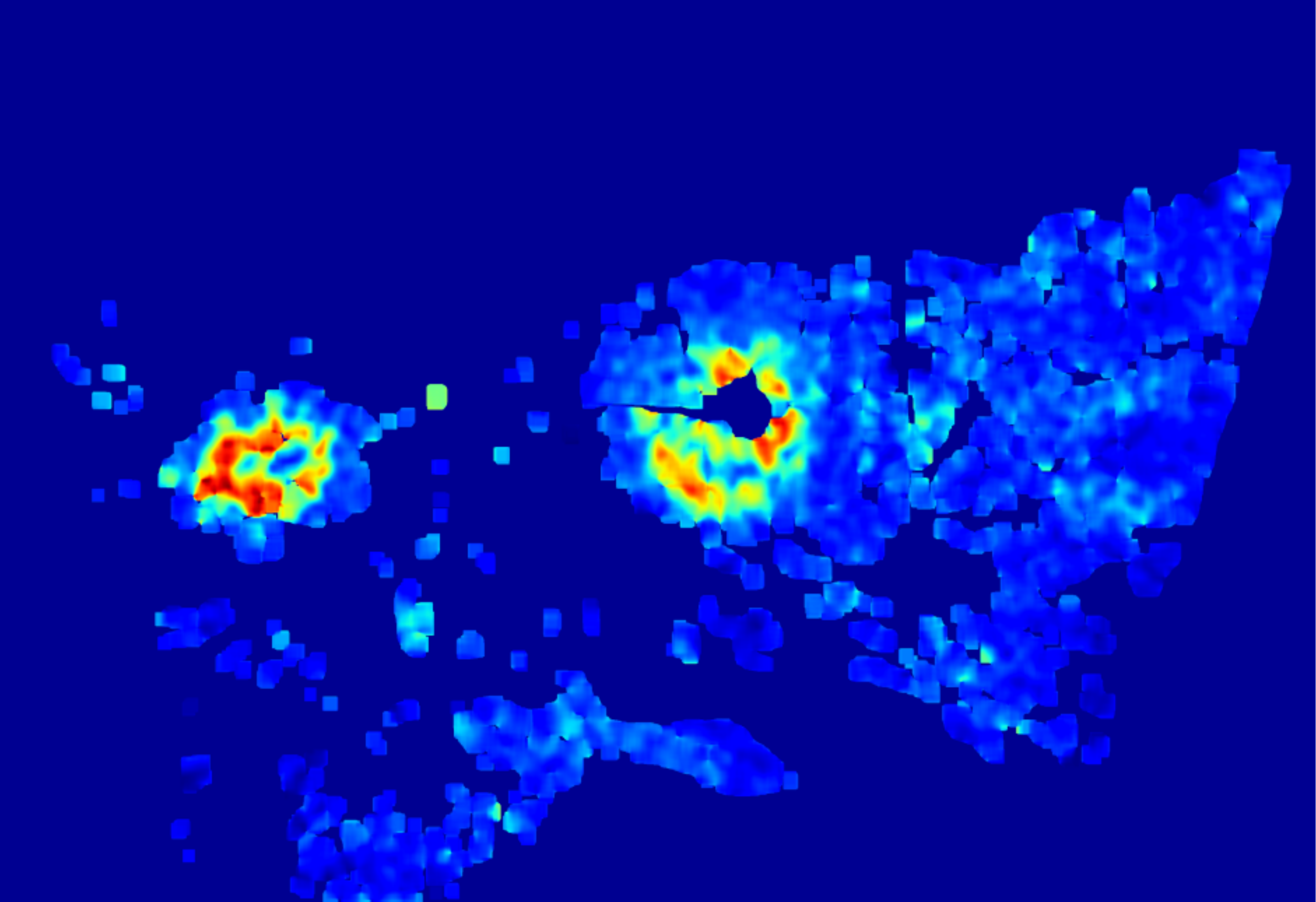}} 
    \centerline{(e)}\medskip    
    \end{minipage}    
\hfill
	\begin{minipage}{0.1\linewidth}
    \centerline{\includegraphics[width=2.1cm, height=2.1cm]{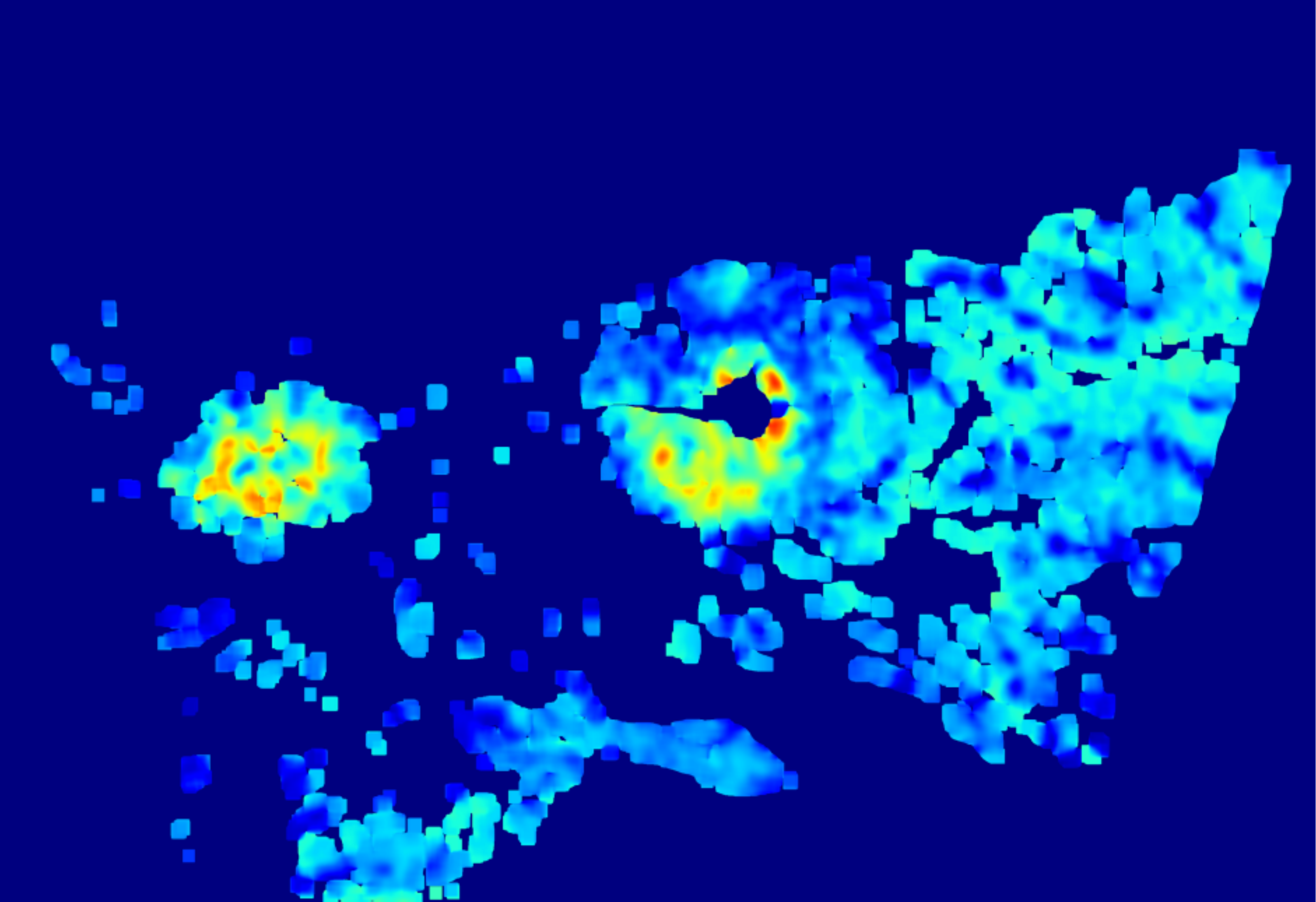}} 
    \centerline{(f)}\medskip    
    \end{minipage}
\hfill
	\begin{minipage}{0.1\linewidth}
    \centerline{\includegraphics[width=2.1cm, height=2.1cm]{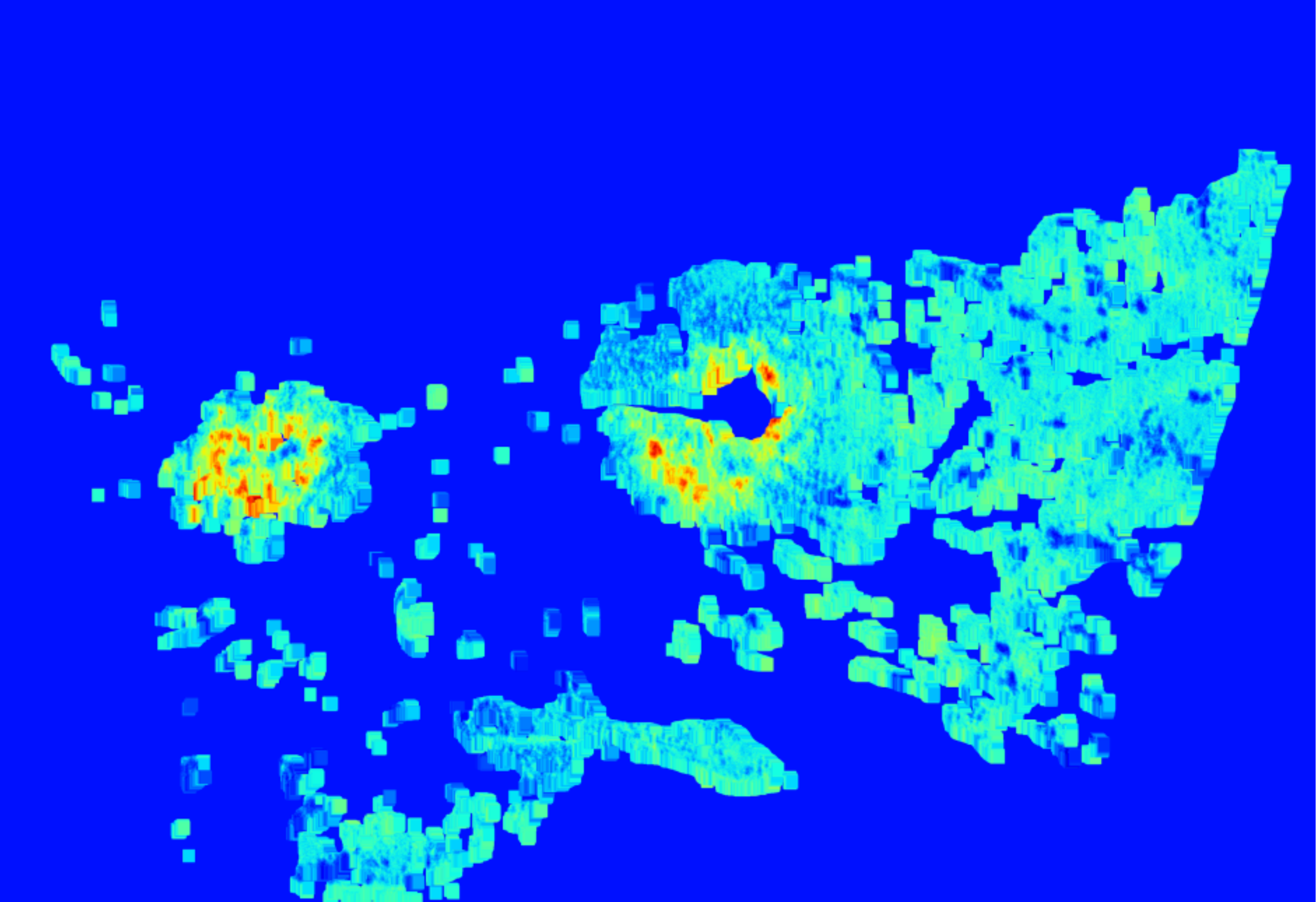}} 
    \centerline{(g)}\medskip    
    \end{minipage}
\hfill
	\begin{minipage}{0.1\linewidth}
    \centerline{\includegraphics[width=2.1cm, height=2.1cm]{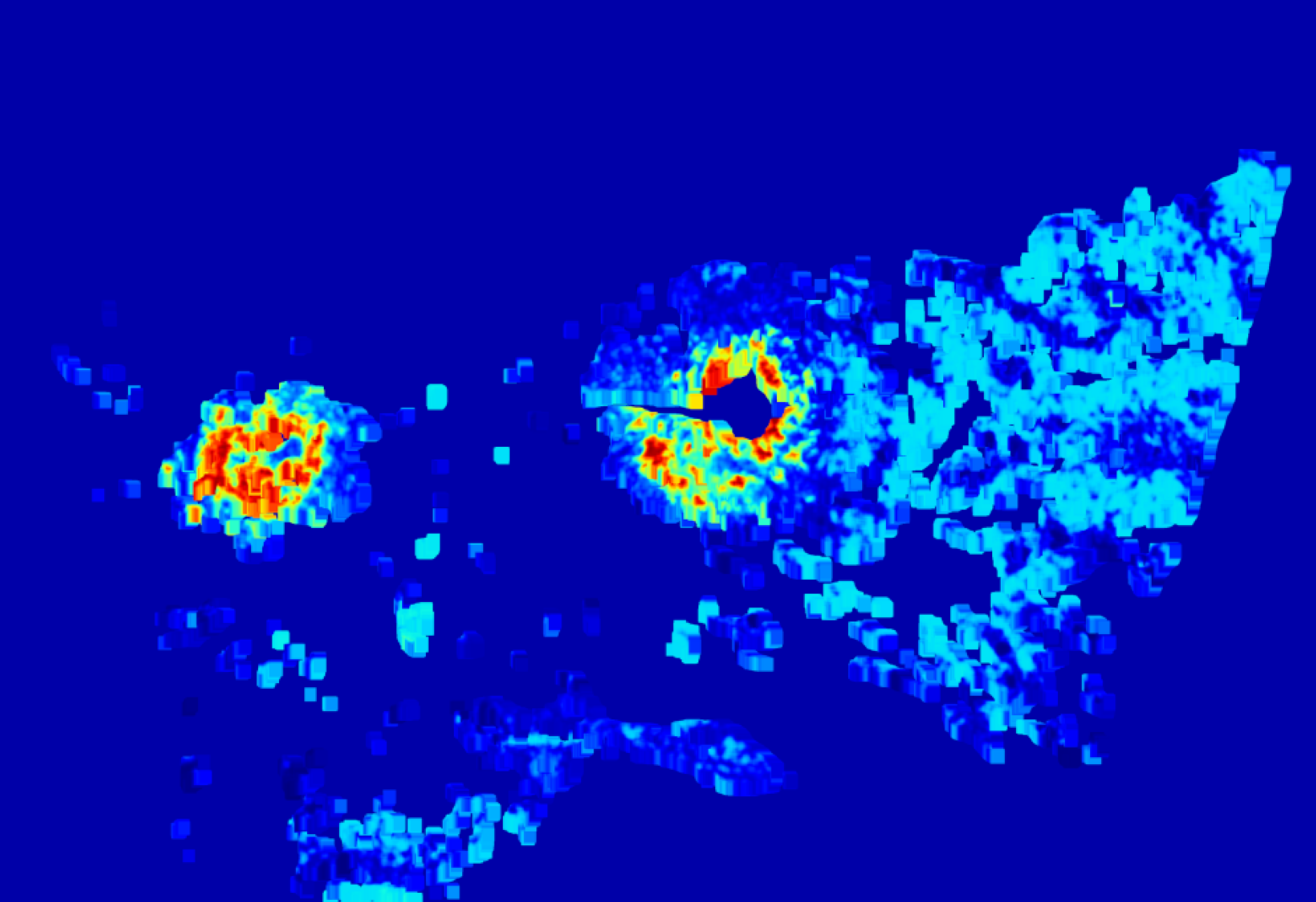}} 
    \centerline{(h)}\medskip    
    \end{minipage}
\hfill
	\begin{minipage}{0.1\linewidth}
    \centerline{\includegraphics[width=2.1cm, height=2.1cm]{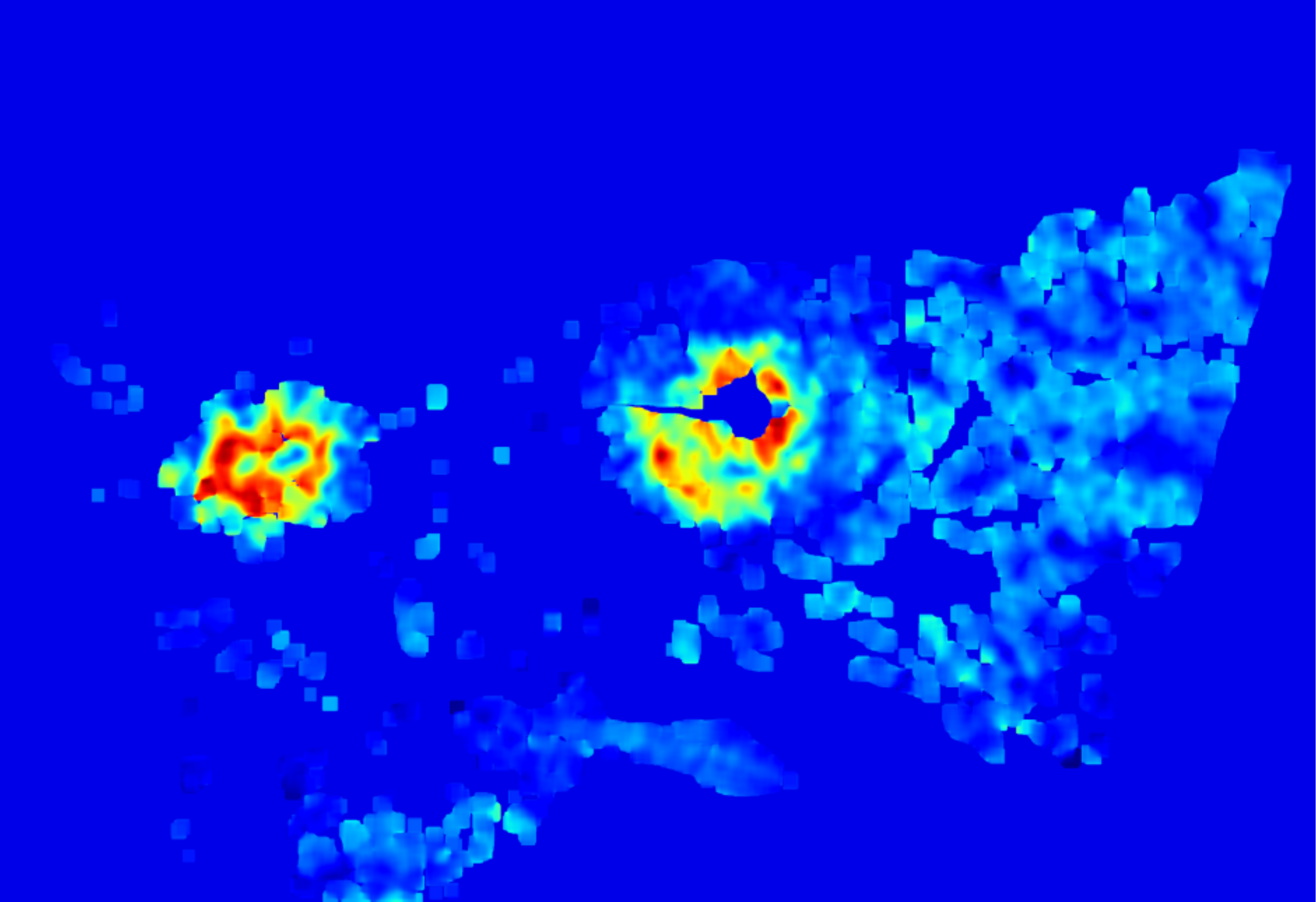}} 
    \centerline{(i)}\medskip    
    \end{minipage}
\hfill
	\begin{minipage}{\linewidth}
    \centerline{\includegraphics[scale=0.60, angle=270,origin=c]{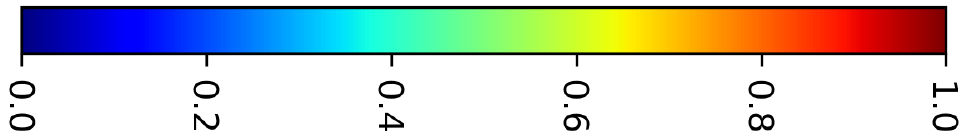}} 
    \end{minipage}
    \vspace{-2.7cm}
\caption{TC scores produced on WSIs of the BreastPathQ test set for 10\% labeled data. (a) Original WSI overlaid with ground truth mask (annotation labels with pink square boxes denote 0\% cellularity and green square boxes indicate 100\% cellularity); (b) -- (e) corresponds to TC score produced by random (supervised), VAE, MoCo, and RSP approach, respectively; (f) -- (i) corresponds to TC score produced by random+CR (supervised), VAE+CR, MoCo+CR, and RSP+CR methods, respectively. The color blue denotes healthy (0\% TC), and red denotes malignant (100\% TC).}
\label{Fig:Heat-maps on BreastPathQ}
\end{figure}
\end{landscape}
}


\afterpage{
\clearpage
\begin{landscape}
\thispagestyle{lscape}
\pagestyle{lscape}

\begin{figure}
 	\begin{minipage}{0.095\linewidth}
 	\centerline{\textbf{Image}}\medskip
    \centerline{\includegraphics[width=1.9cm, height=2cm]{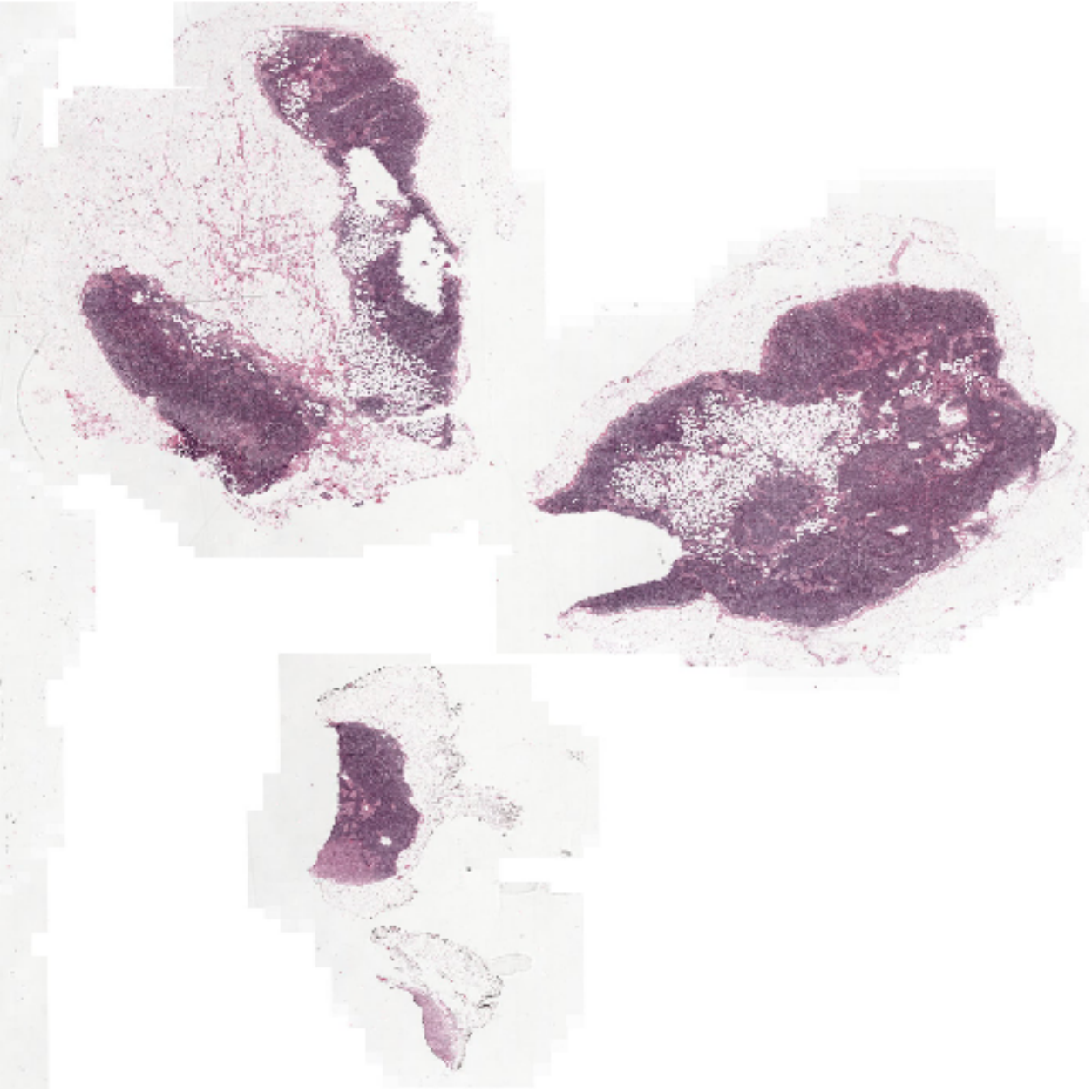}} 
    \end{minipage}
\hfill
	\begin{minipage}{0.095\linewidth}
	\centerline{\textbf{Ground truth}}\medskip
    \centerline{\includegraphics[width=1.9cm, height=2cm]{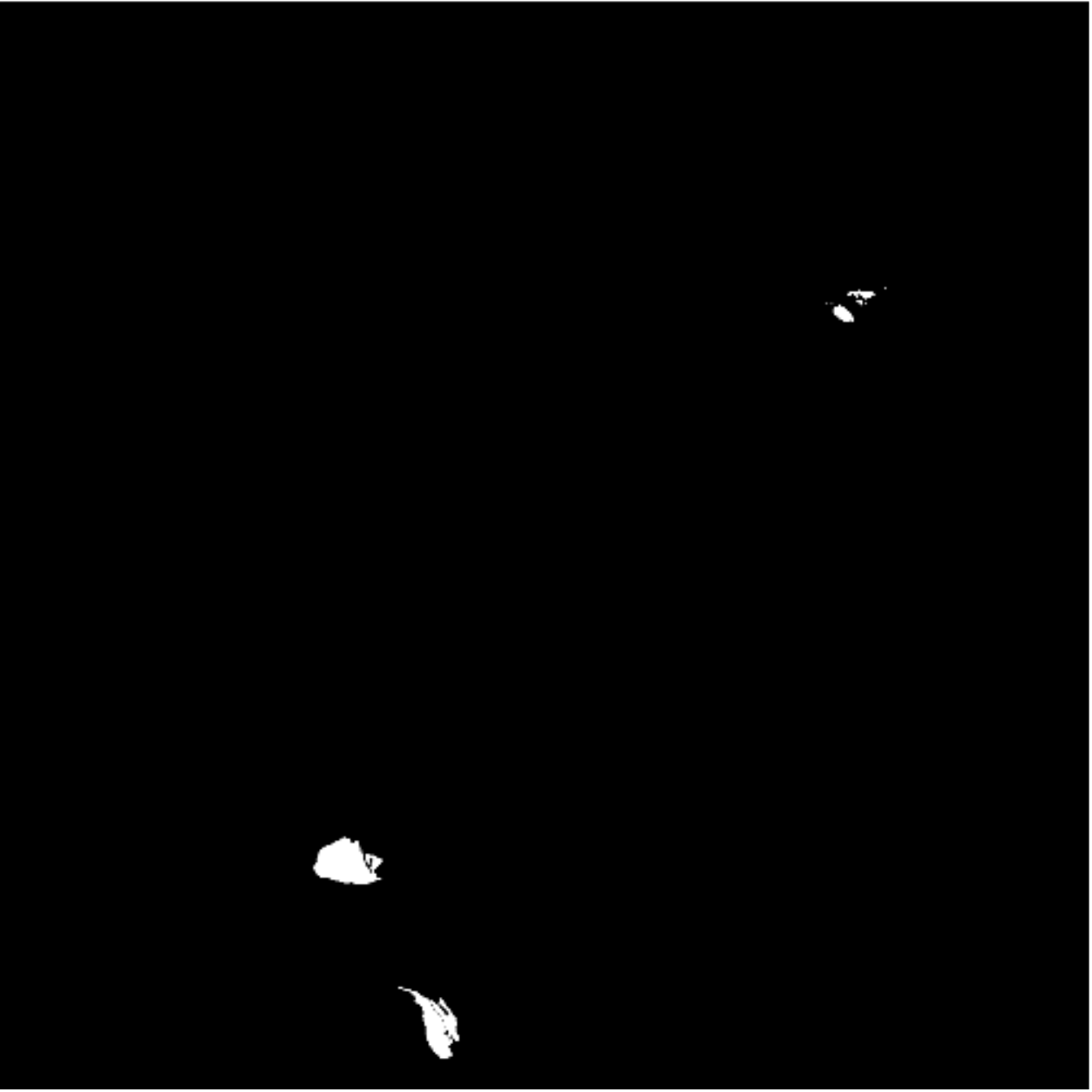}} 
    \end{minipage}    
\hfill
	\begin{minipage}{0.095\linewidth}
	\centerline{\textbf{Random}}\medskip
    \centerline{\includegraphics[width=1.9cm, height=2cm]{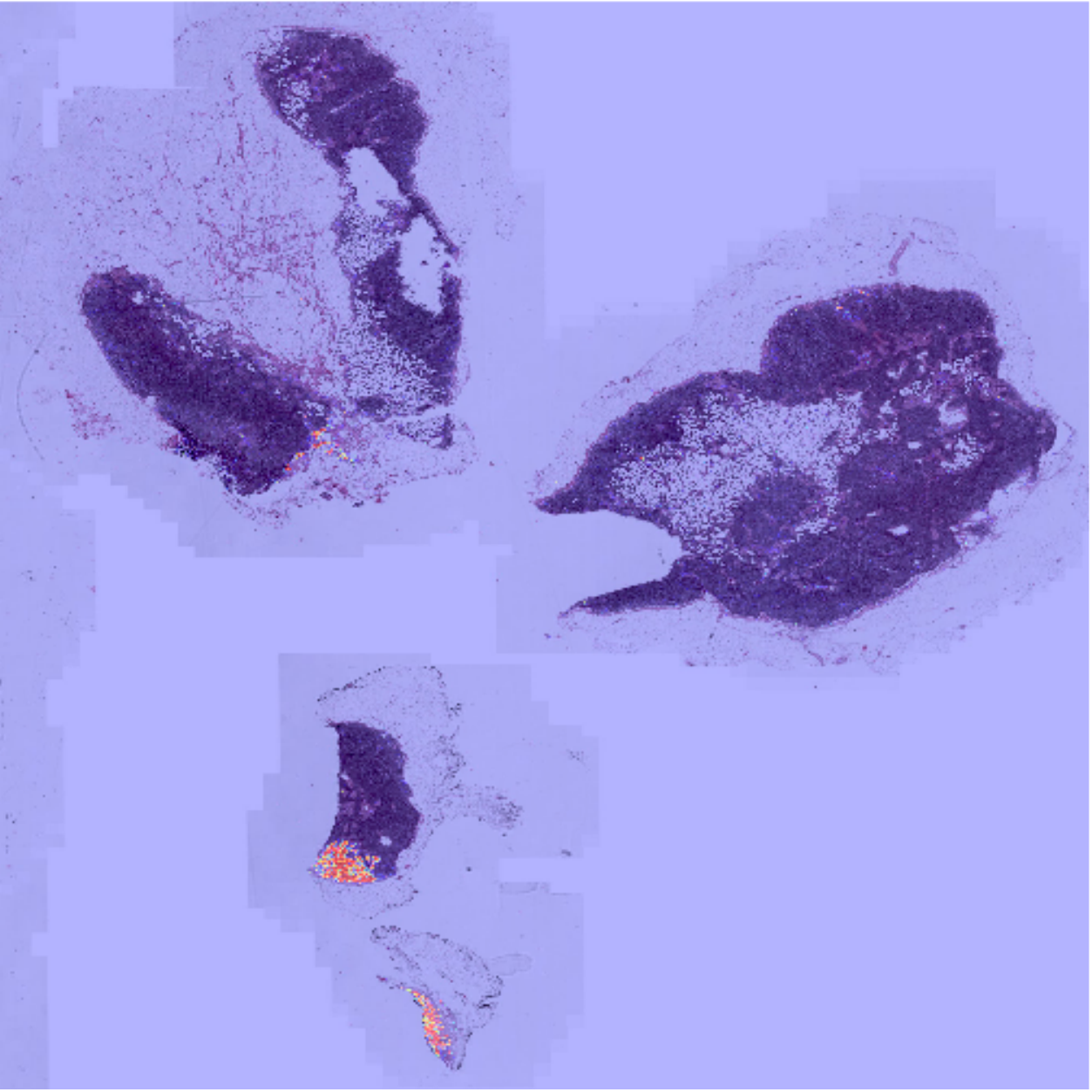}} 
    \end{minipage}
\hfill
	\begin{minipage}{0.095\linewidth}
	\centerline{\textbf{VAE}}\medskip
    \centerline{\includegraphics[width=1.9cm, height=2cm]{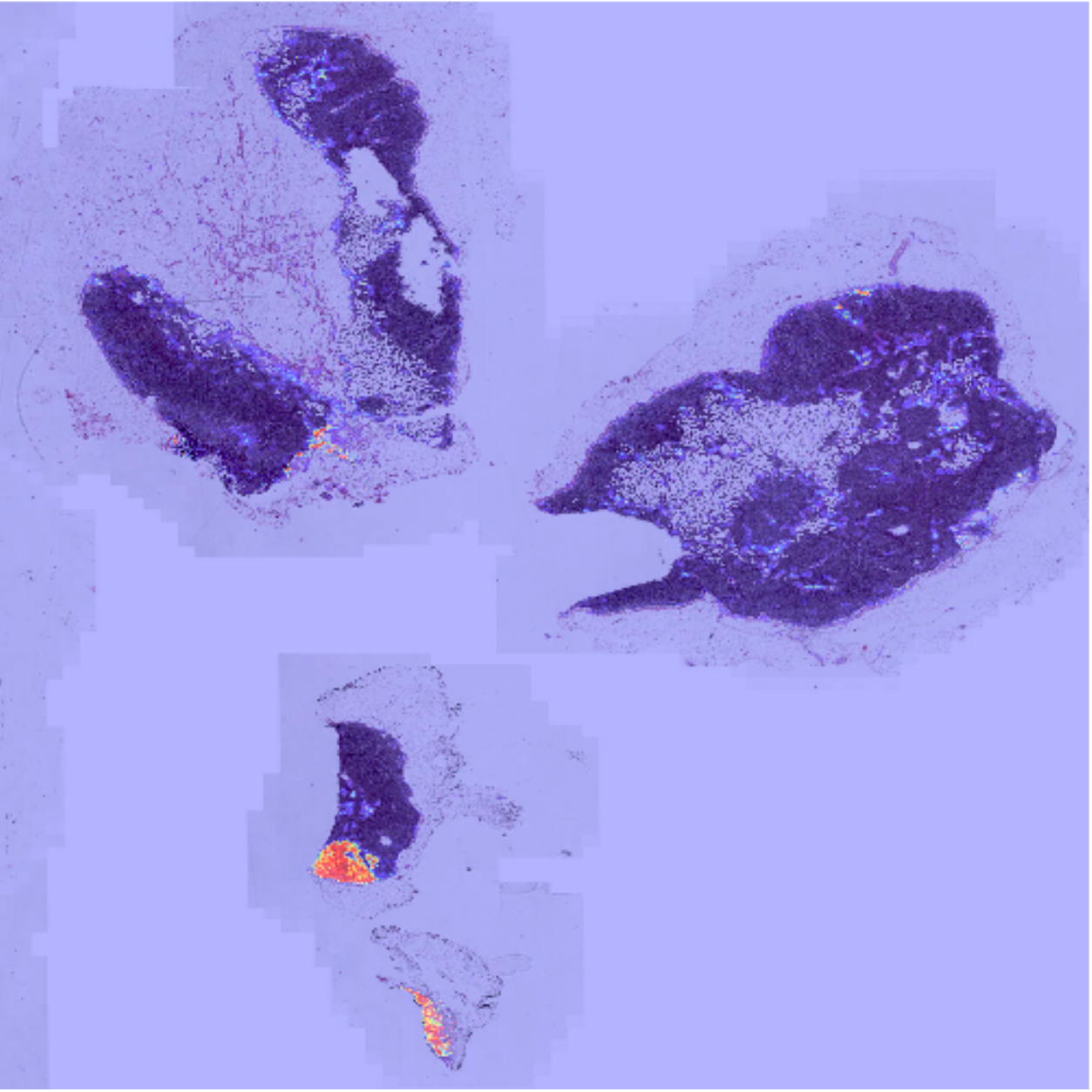}} 
    \end{minipage}
\hfill
	\begin{minipage}{0.095\linewidth}
	\centerline{\textbf{MoCo}}\medskip
    \centerline{\includegraphics[width=1.9cm, height=2cm]{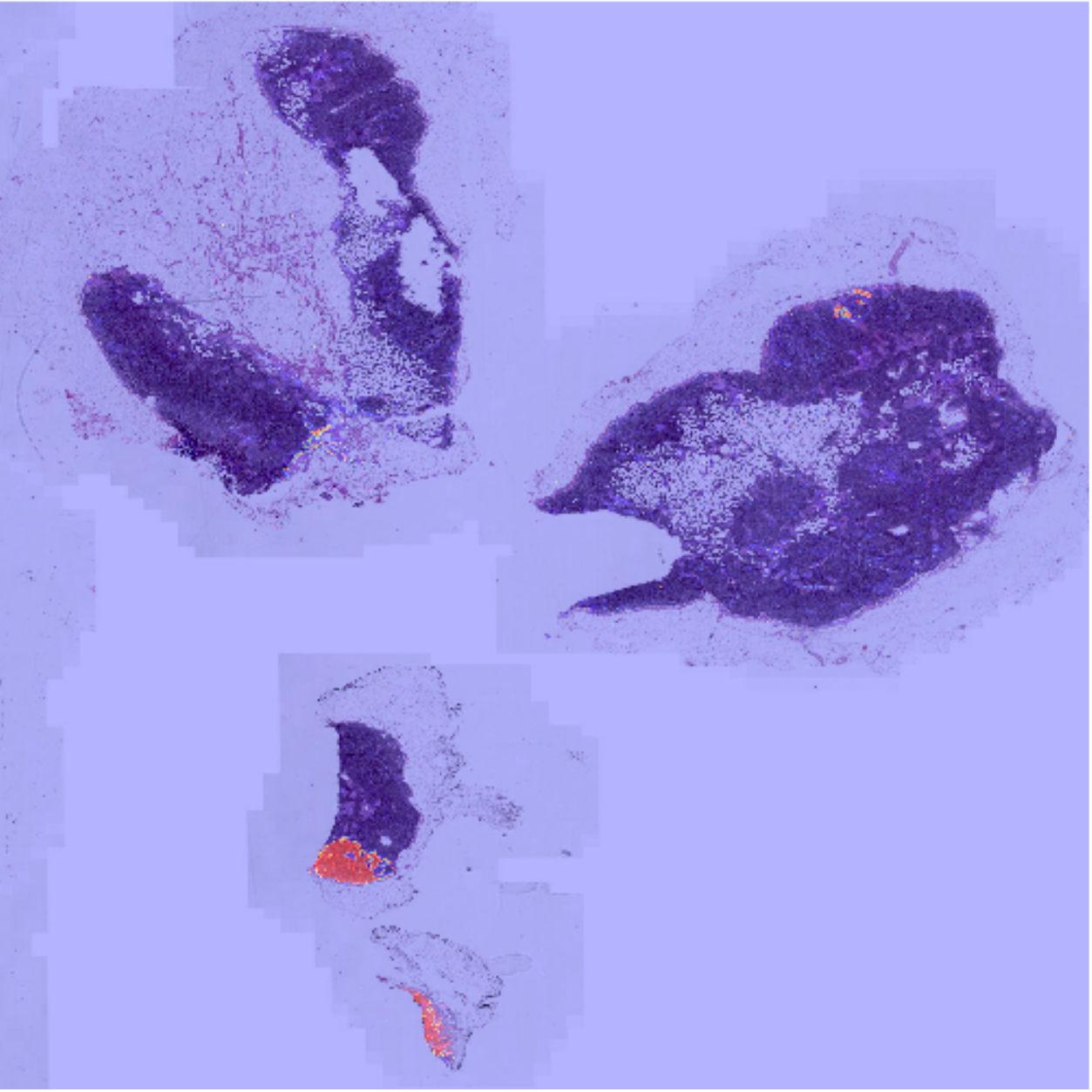}} 
    \end{minipage}
\hfill
	\begin{minipage}{0.095\linewidth}
	\centerline{\textbf{RSP}}\medskip
    \centerline{\includegraphics[width=1.9cm, height=2cm]{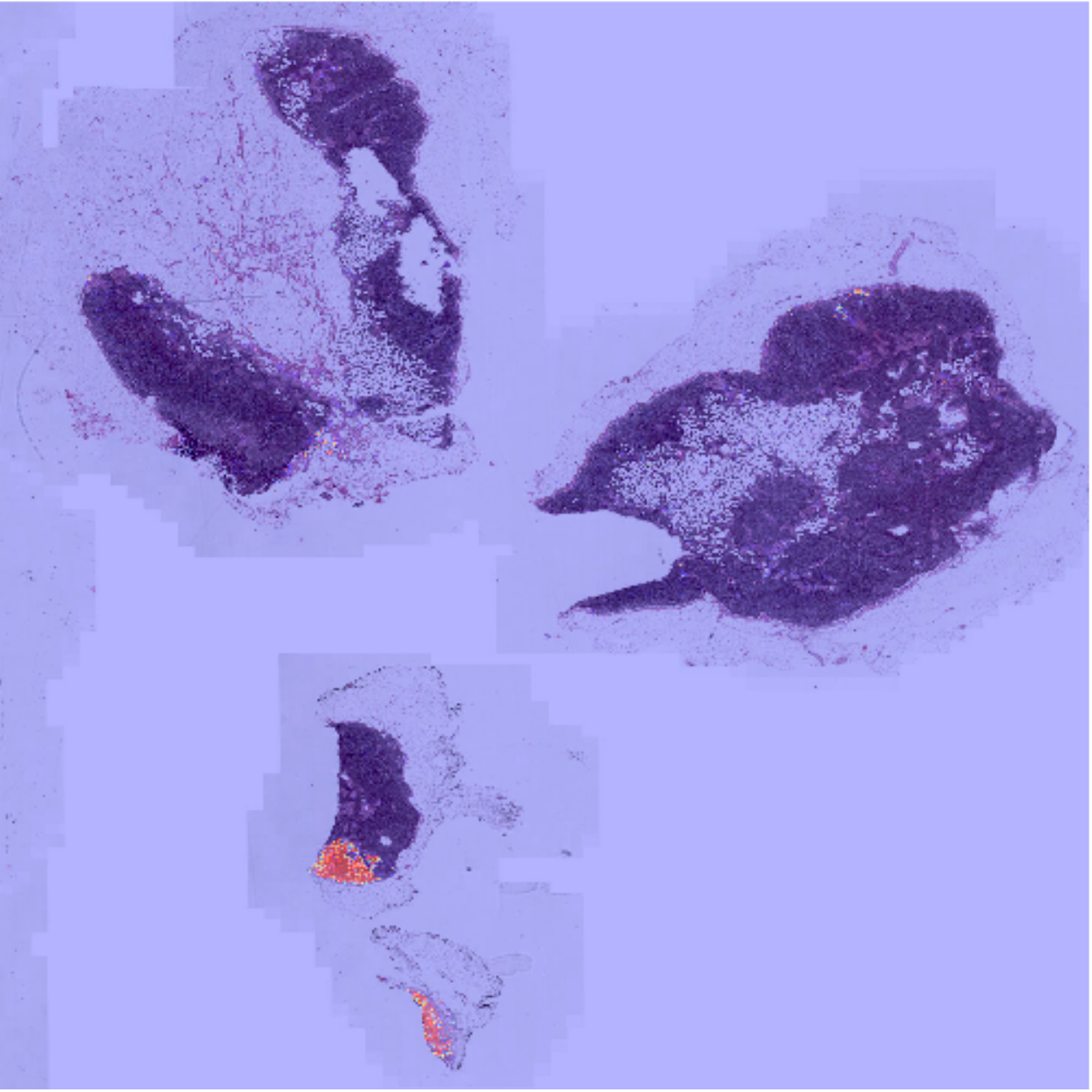}} 
    \end{minipage}    
\hfill
	\begin{minipage}{0.095\linewidth}
	\centerline{\textbf{Random+CR}}\medskip
    \centerline{\includegraphics[width=1.9cm, height=2cm]{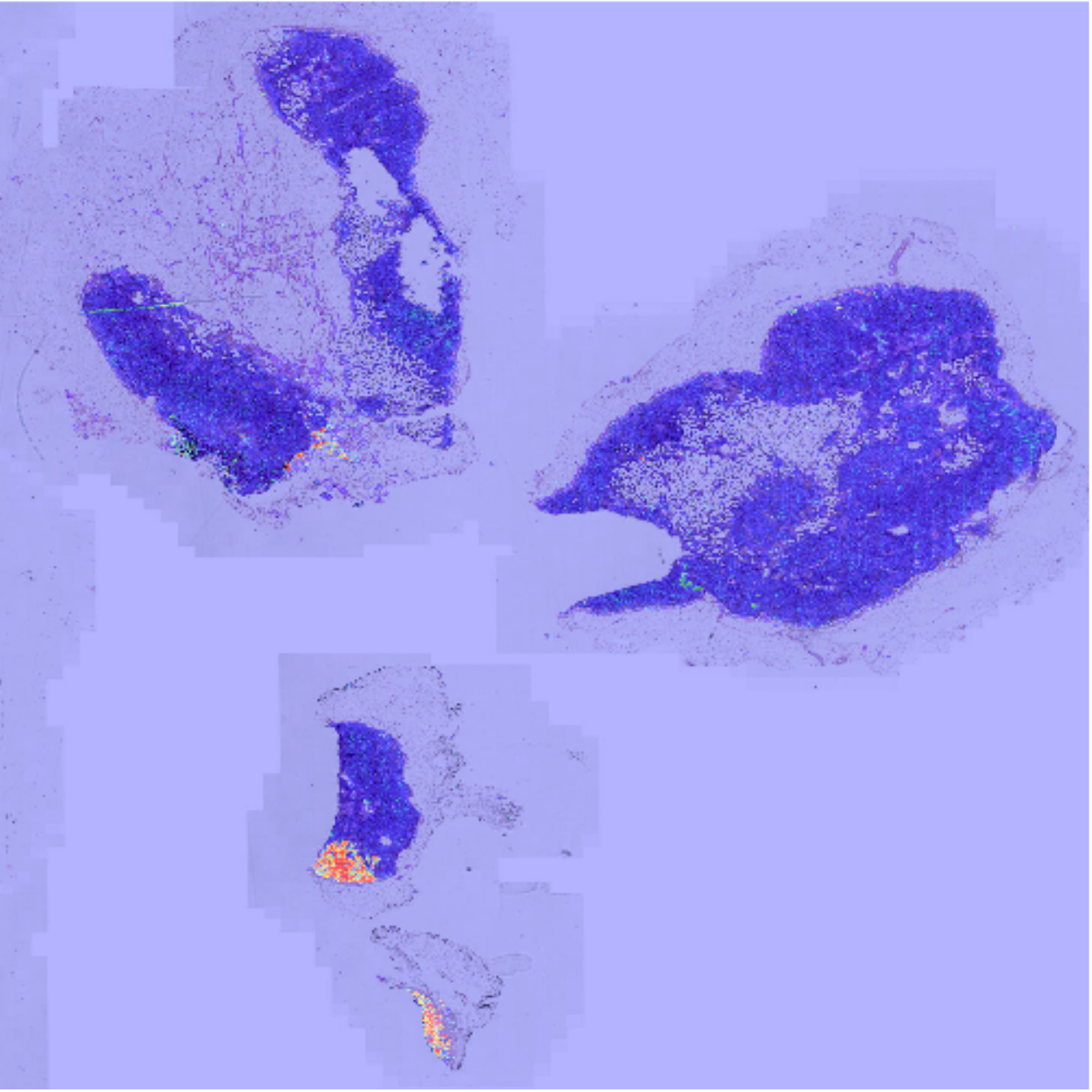}} 
    \end{minipage}
\hfill
	\begin{minipage}{0.095\linewidth}
	\centerline{\textbf{VAE+CR}}\medskip
    \centerline{\includegraphics[width=1.9cm, height=2cm]{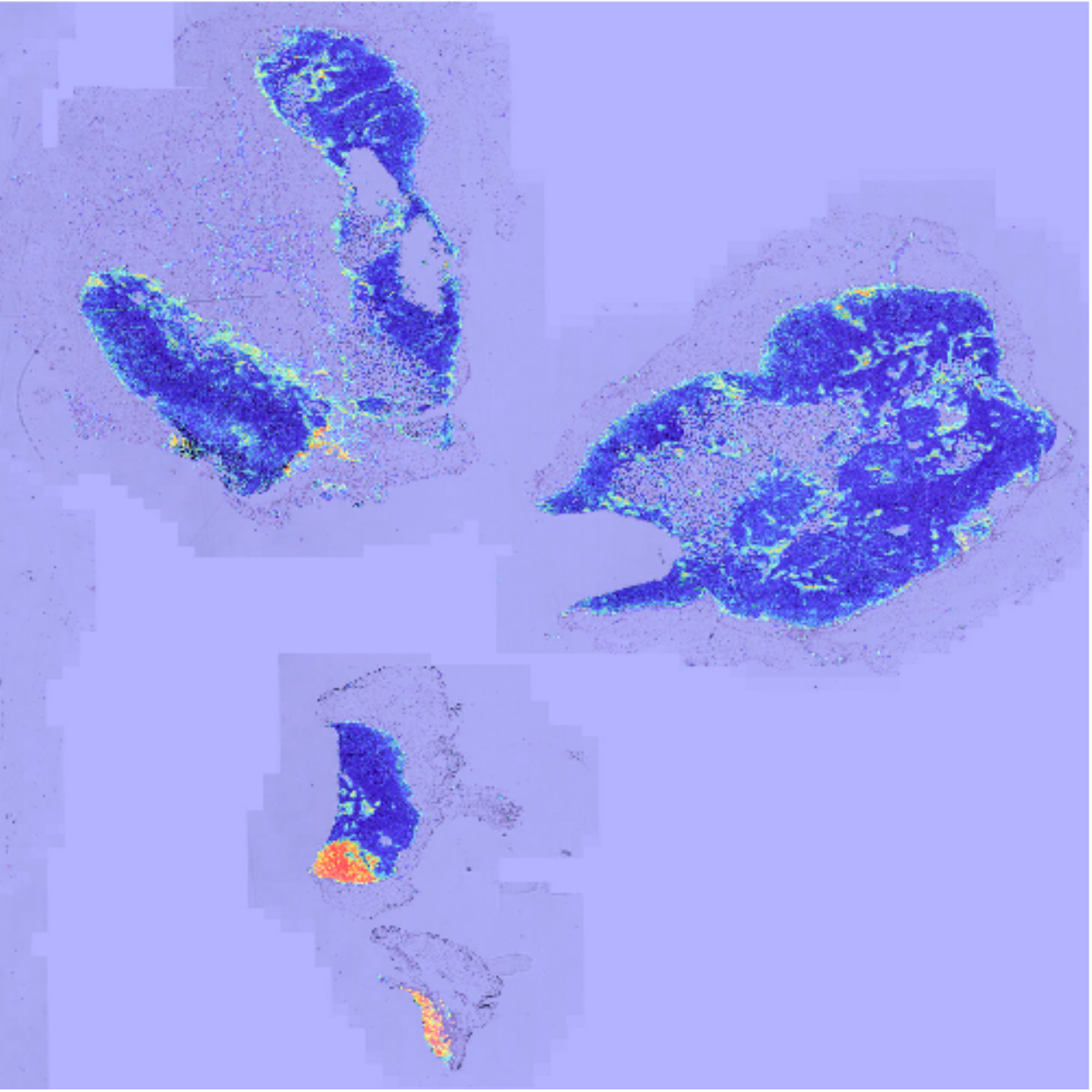}} 
    \end{minipage}
\hfill
	\begin{minipage}{0.095\linewidth}
	\centerline{\textbf{MoCo+CR}}\medskip
    \centerline{\includegraphics[width=1.9cm, height=2cm]{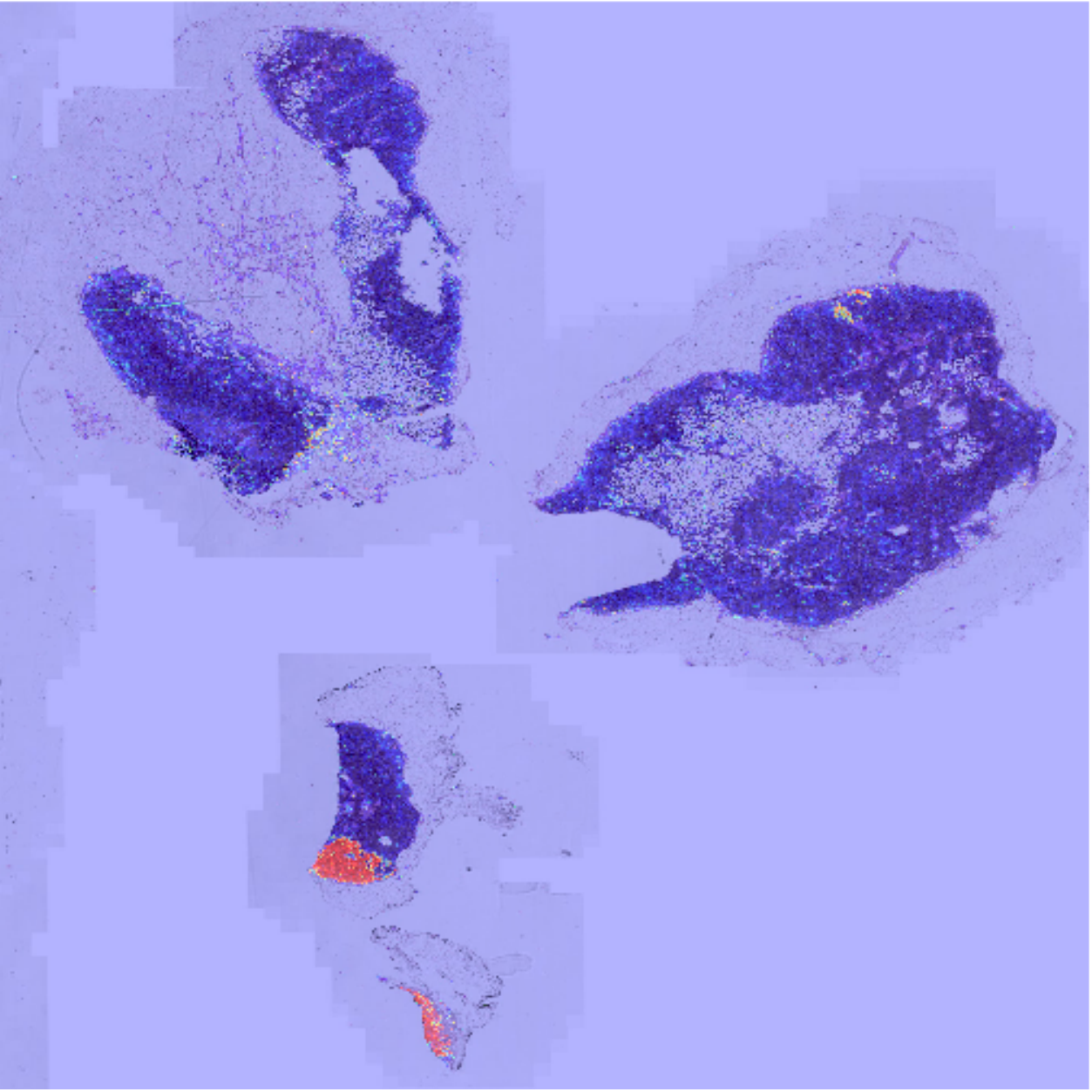}} 
    \end{minipage}
\hfill
	\begin{minipage}{0.095\linewidth}
	\centerline{\textbf{RSP+CR}}\medskip
    \centerline{\includegraphics[width=1.9cm, height=2cm]{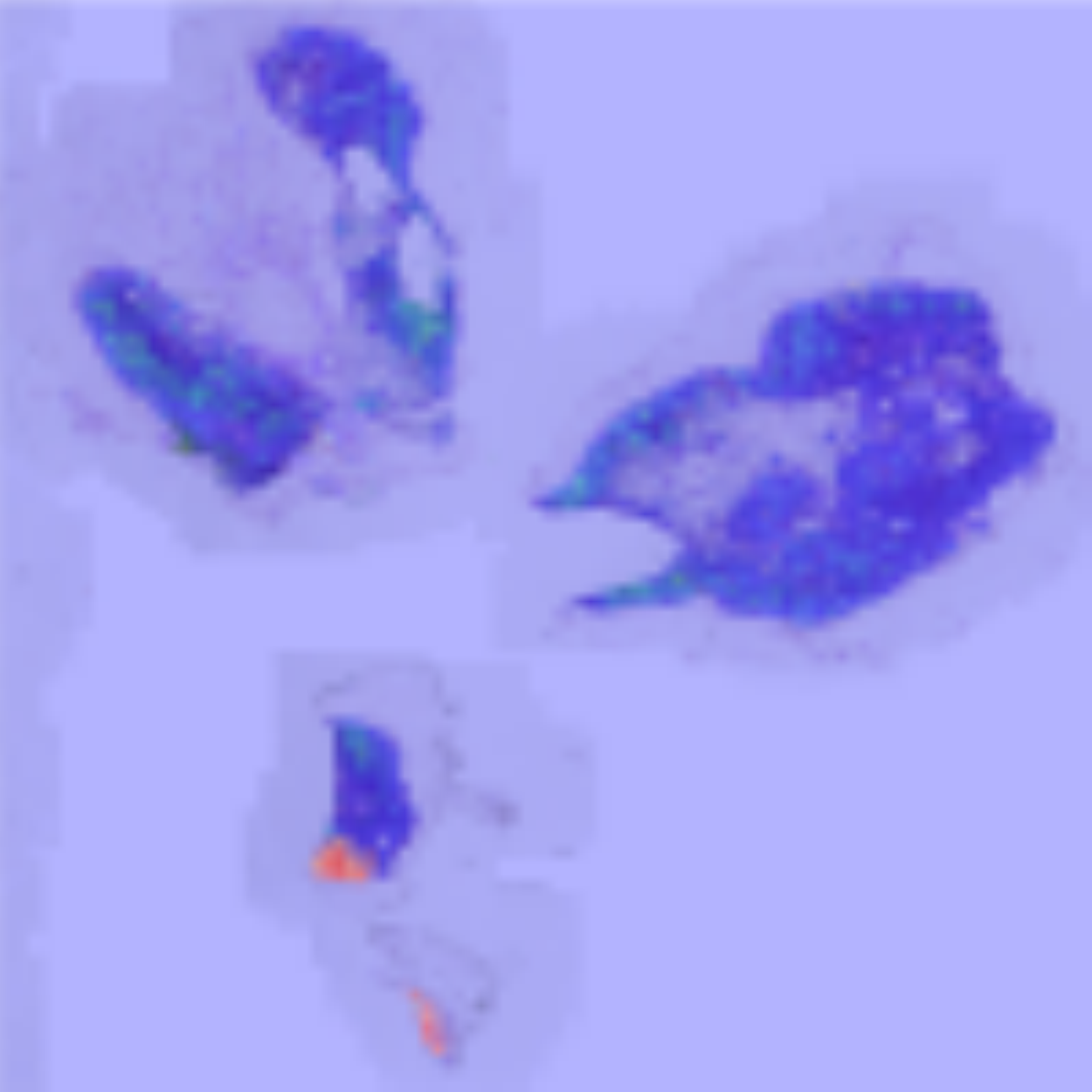}} 
    \end{minipage}   
    
\vfill	

 	\begin{minipage}{0.095\linewidth}
    \centerline{\includegraphics[width=1.9cm, height=2cm]{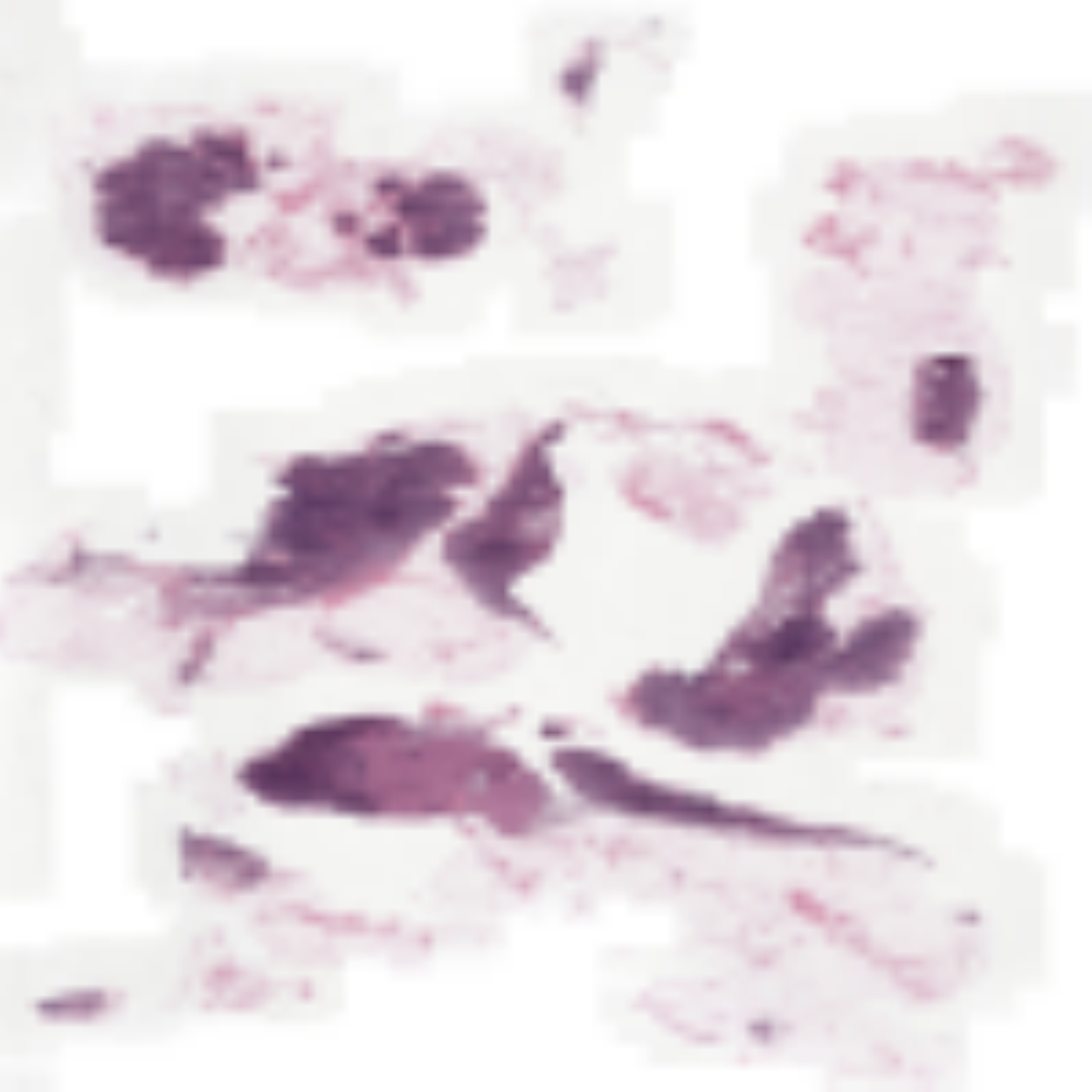}} 
    \end{minipage}
\hfill
	\begin{minipage}{0.095\linewidth}
    \centerline{\includegraphics[width=1.9cm, height=2cm]{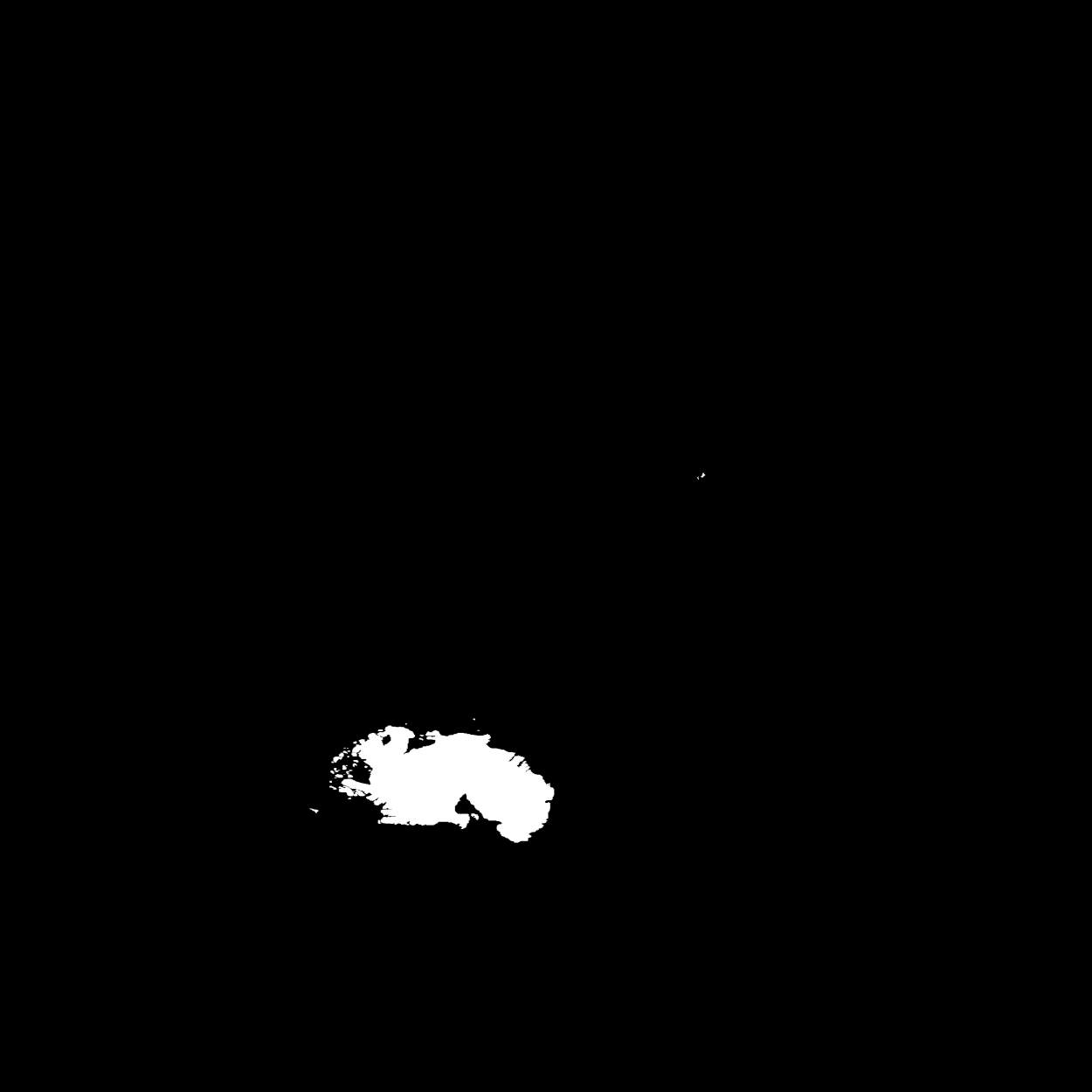}} 
    \end{minipage}    
\hfill
	\begin{minipage}{0.095\linewidth}
    \centerline{\includegraphics[width=1.9cm, height=2cm]{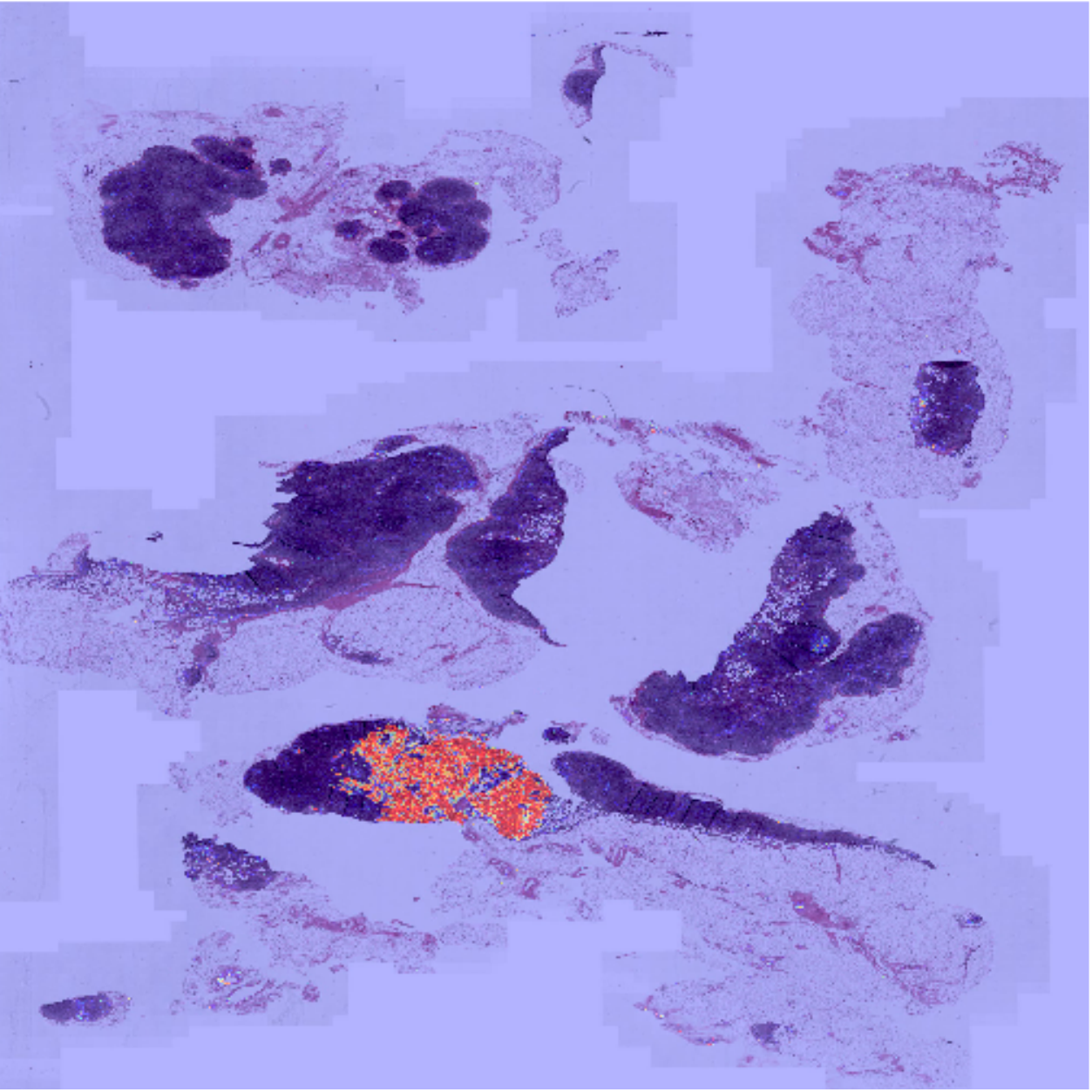}} 
    \end{minipage}
\hfill
	\begin{minipage}{0.095\linewidth}
    \centerline{\includegraphics[width=1.9cm, height=2cm]{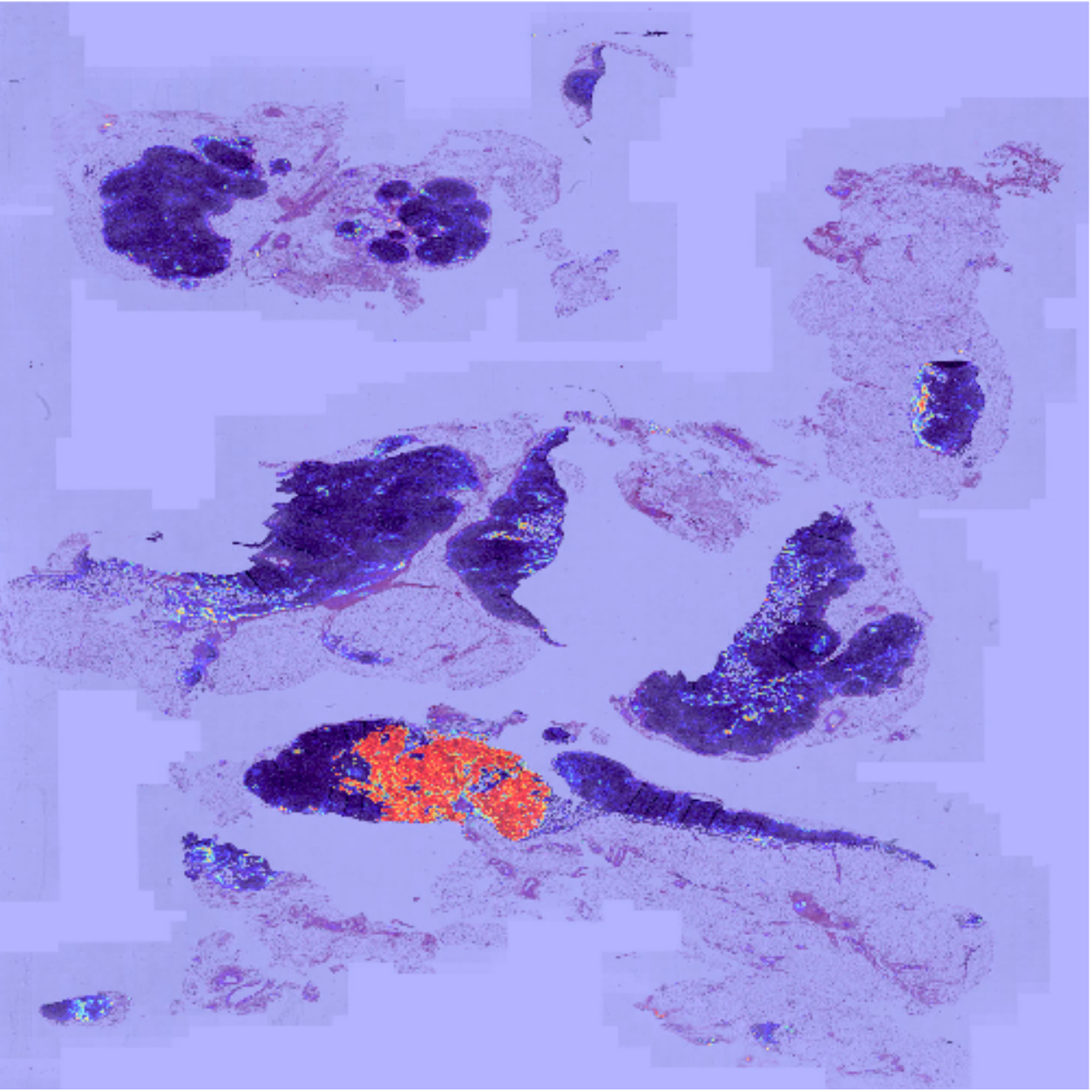}} 
    \end{minipage}
\hfill
	\begin{minipage}{0.095\linewidth}
    \centerline{\includegraphics[width=1.9cm, height=2cm]{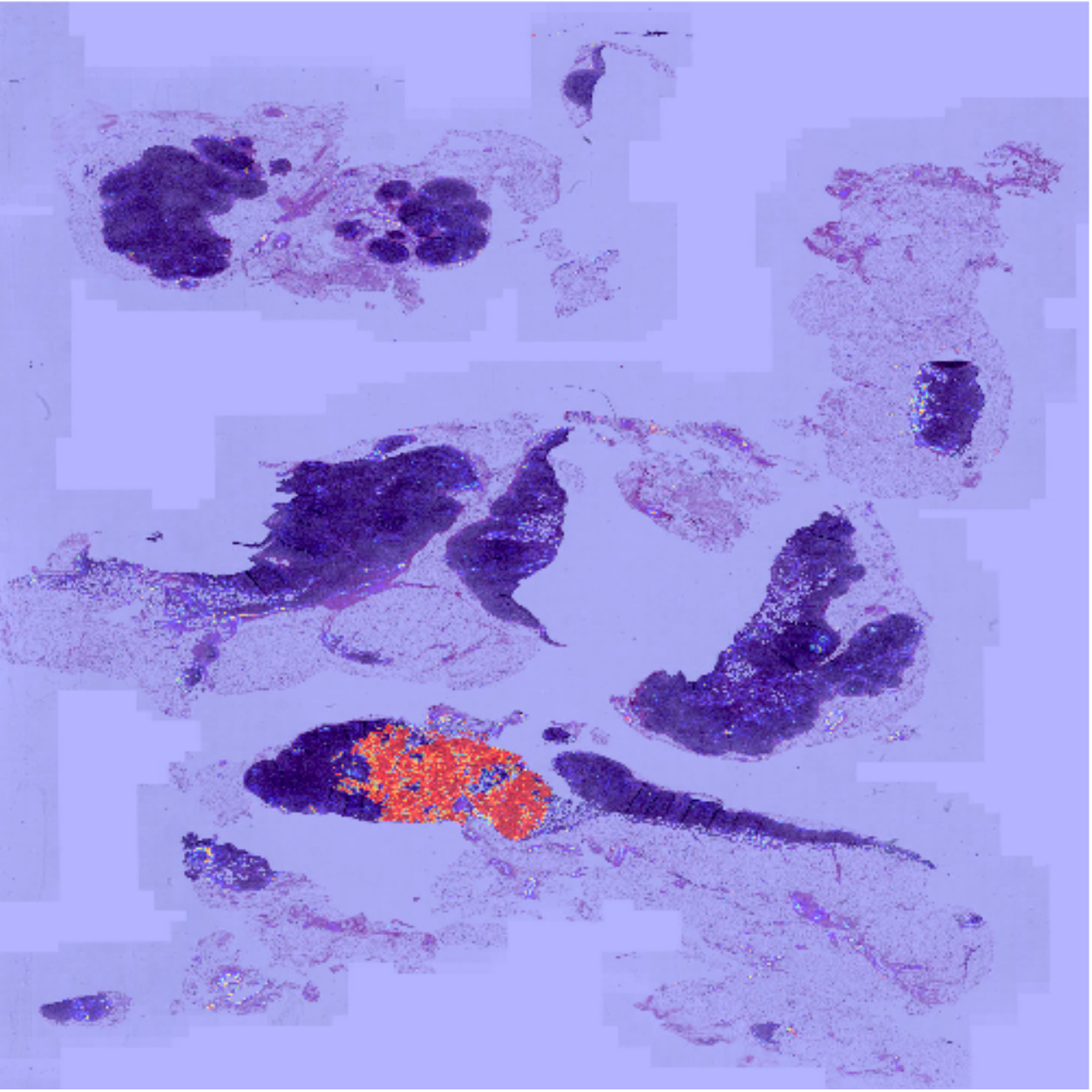}} 
    \end{minipage}
\hfill
	\begin{minipage}{0.095\linewidth}
    \centerline{\includegraphics[width=1.9cm, height=2cm]{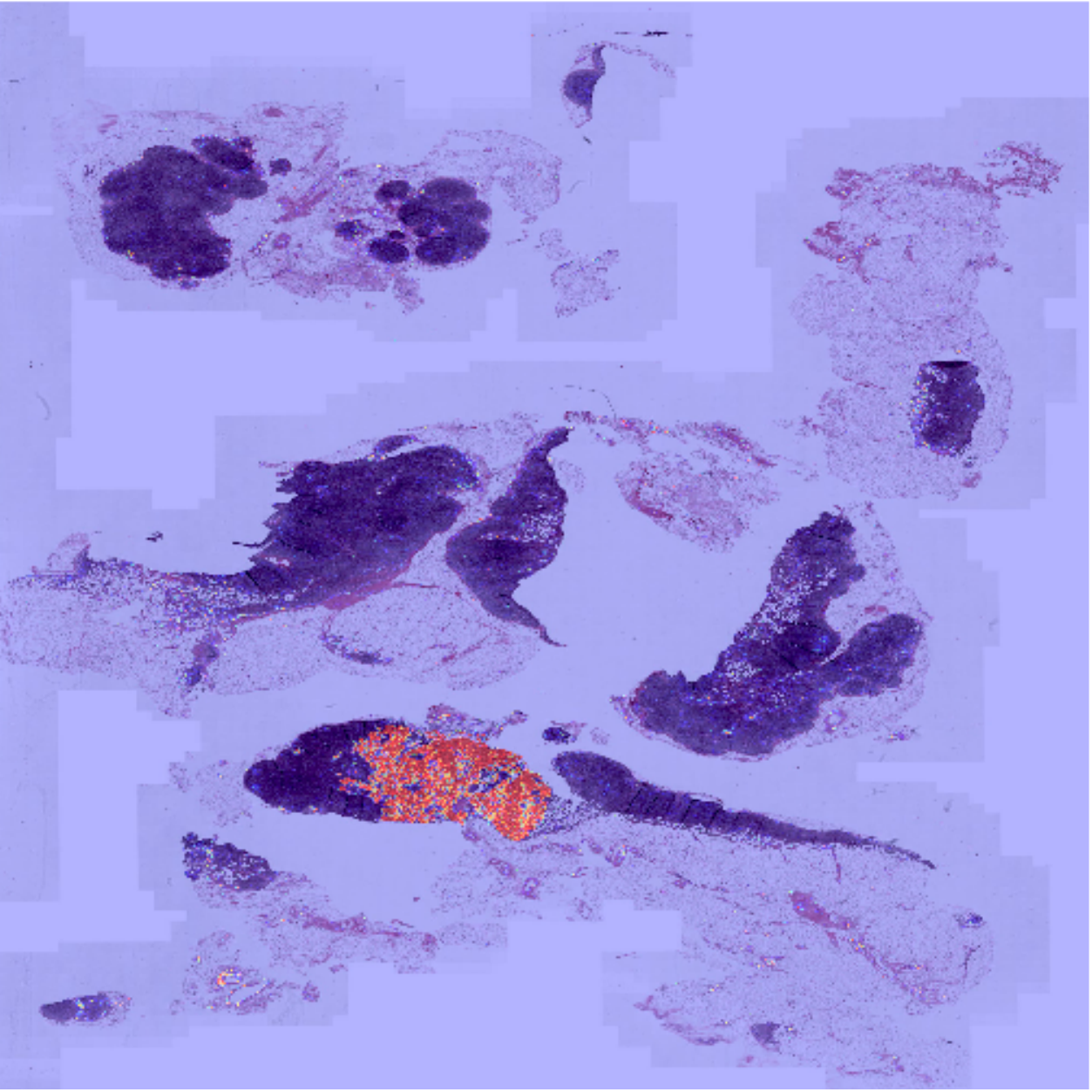}} 
    \end{minipage}    
\hfill
	\begin{minipage}{0.095\linewidth}
    \centerline{\includegraphics[width=1.9cm, height=2cm]{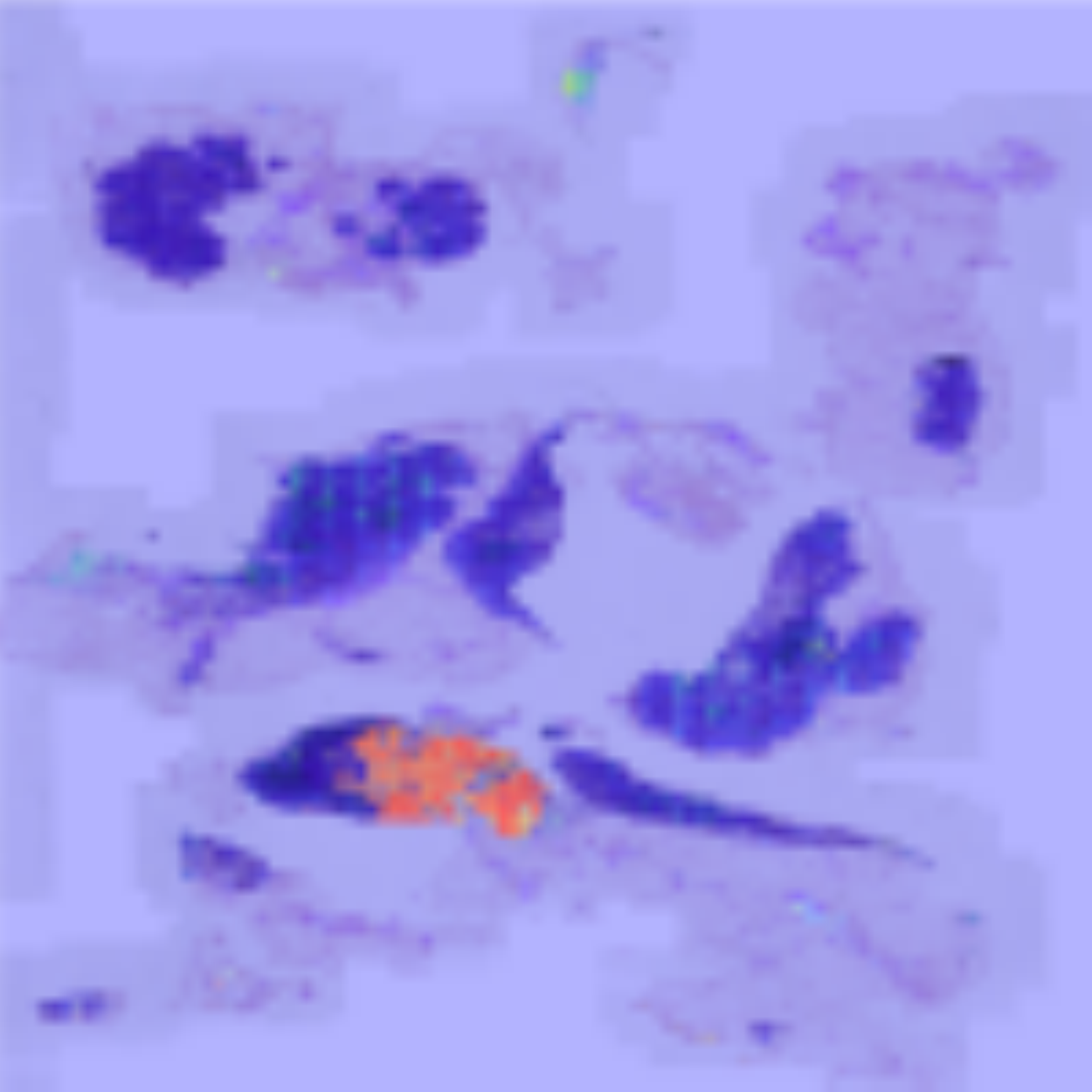}} 
    \end{minipage}
\hfill
	\begin{minipage}{0.095\linewidth}
    \centerline{\includegraphics[width=1.9cm, height=2cm]{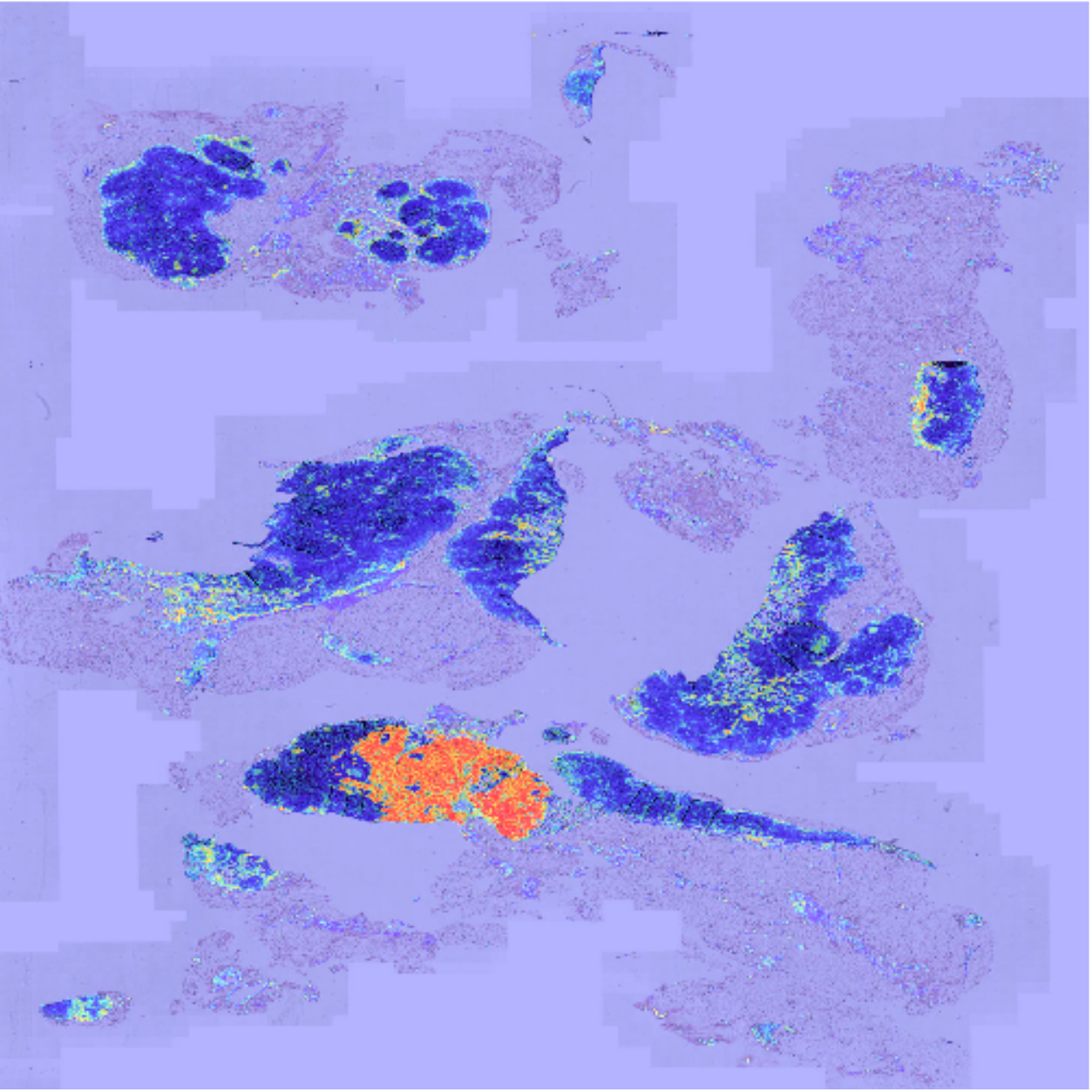}} 
    \end{minipage}
\hfill
	\begin{minipage}{0.095\linewidth}
    \centerline{\includegraphics[width=1.9cm, height=2cm]{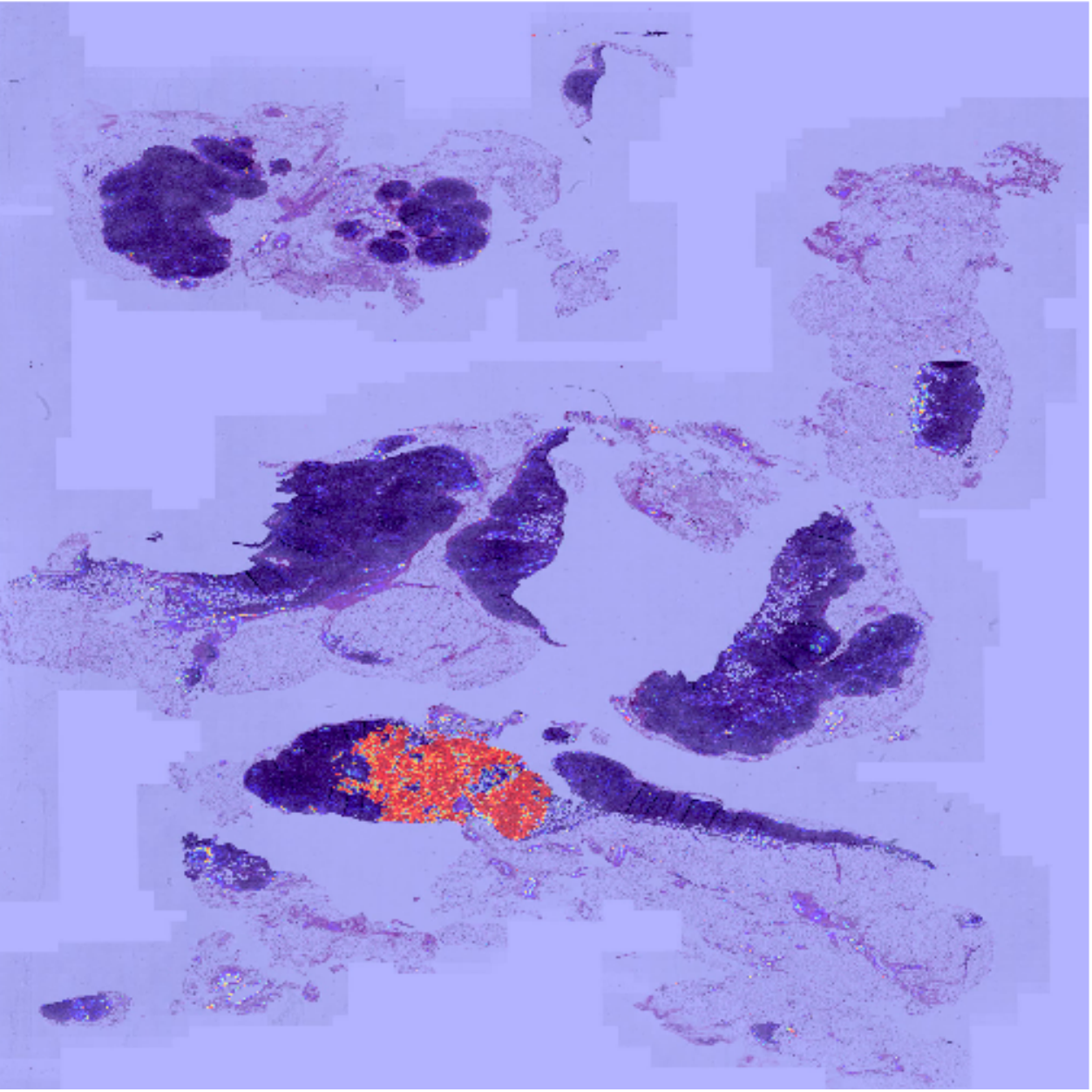}} 
    \end{minipage}
\hfill
	\begin{minipage}{0.095\linewidth}
    \centerline{\includegraphics[width=1.9cm, height=2cm]{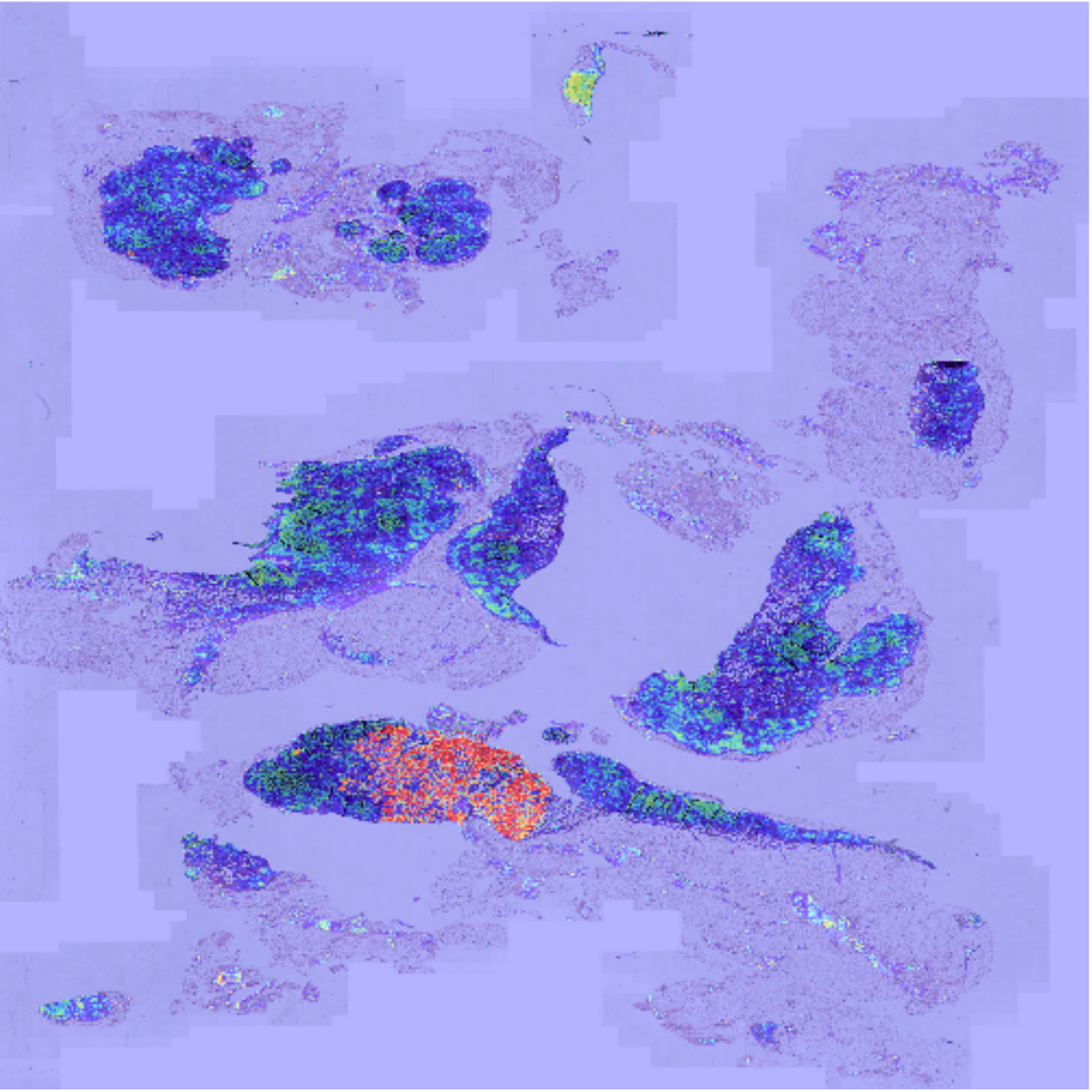}} 
    \end{minipage}    
    
\vfill	

 	\begin{minipage}{0.095\linewidth}
    \centerline{\includegraphics[width=1.9cm, height=2cm]{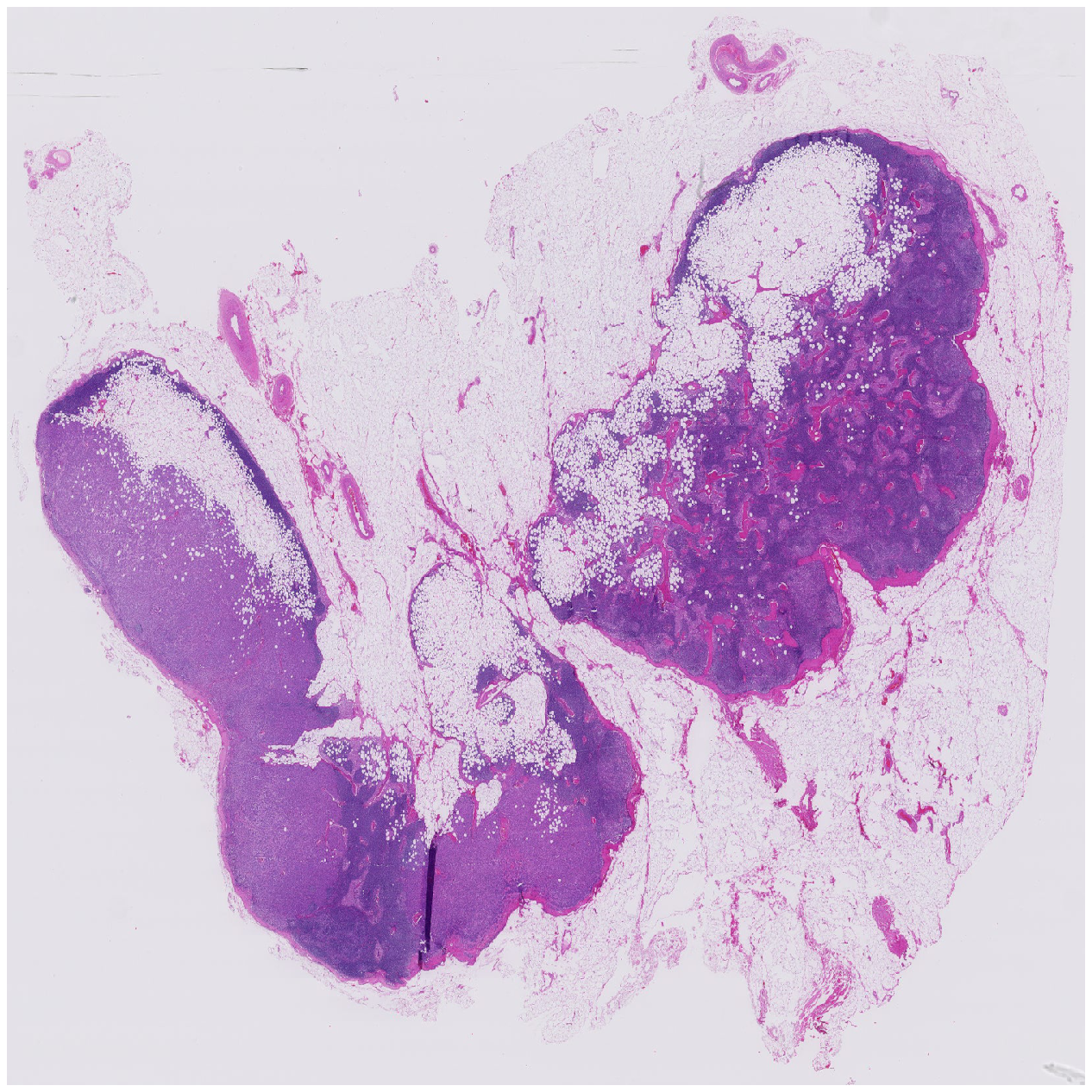}} 
    \end{minipage}
\hfill
	\begin{minipage}{0.095\linewidth}
    \centerline{\includegraphics[width=1.9cm, height=2cm]{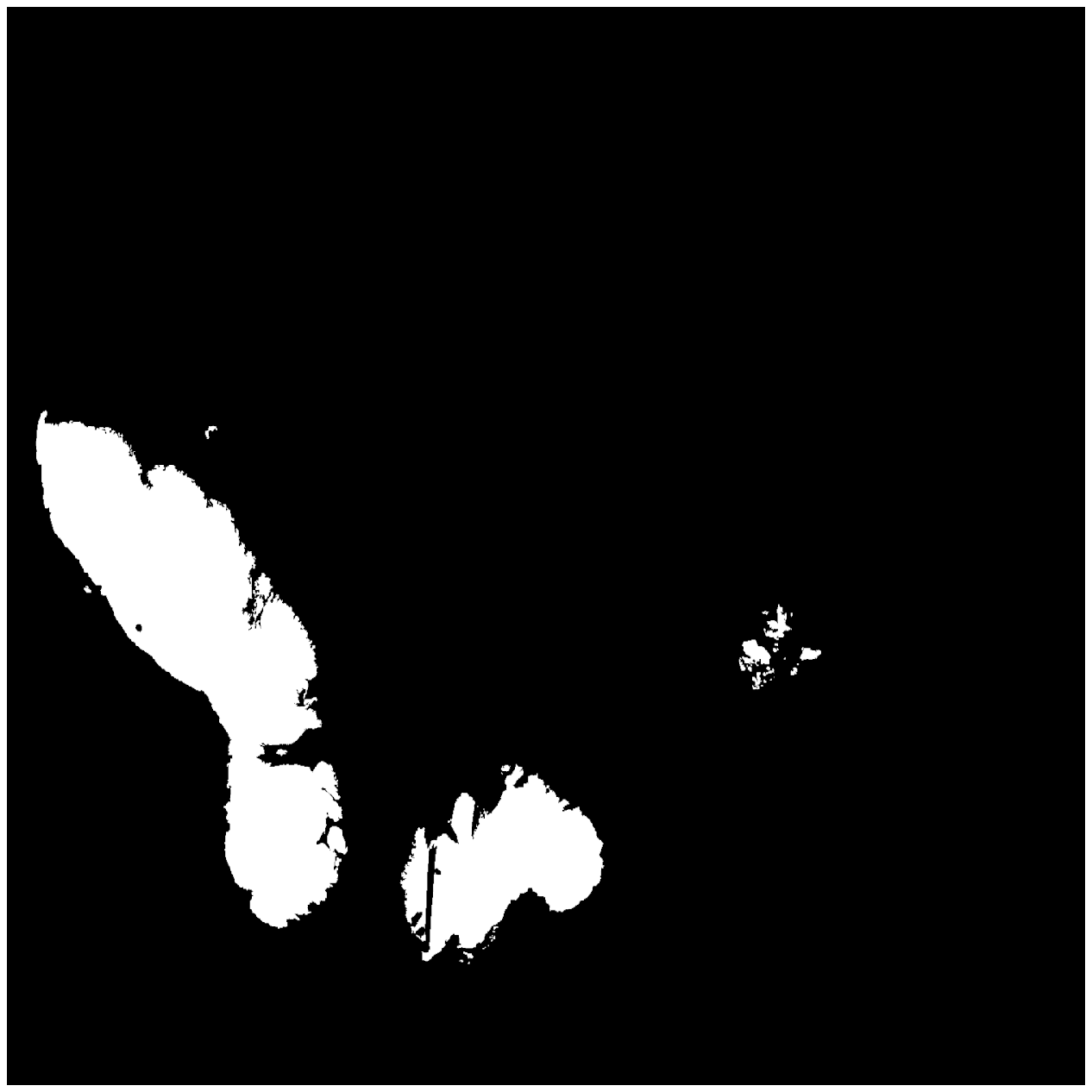}} 
    \end{minipage}    
\hfill
	\begin{minipage}{0.095\linewidth}
    \centerline{\includegraphics[width=1.9cm, height=2cm]{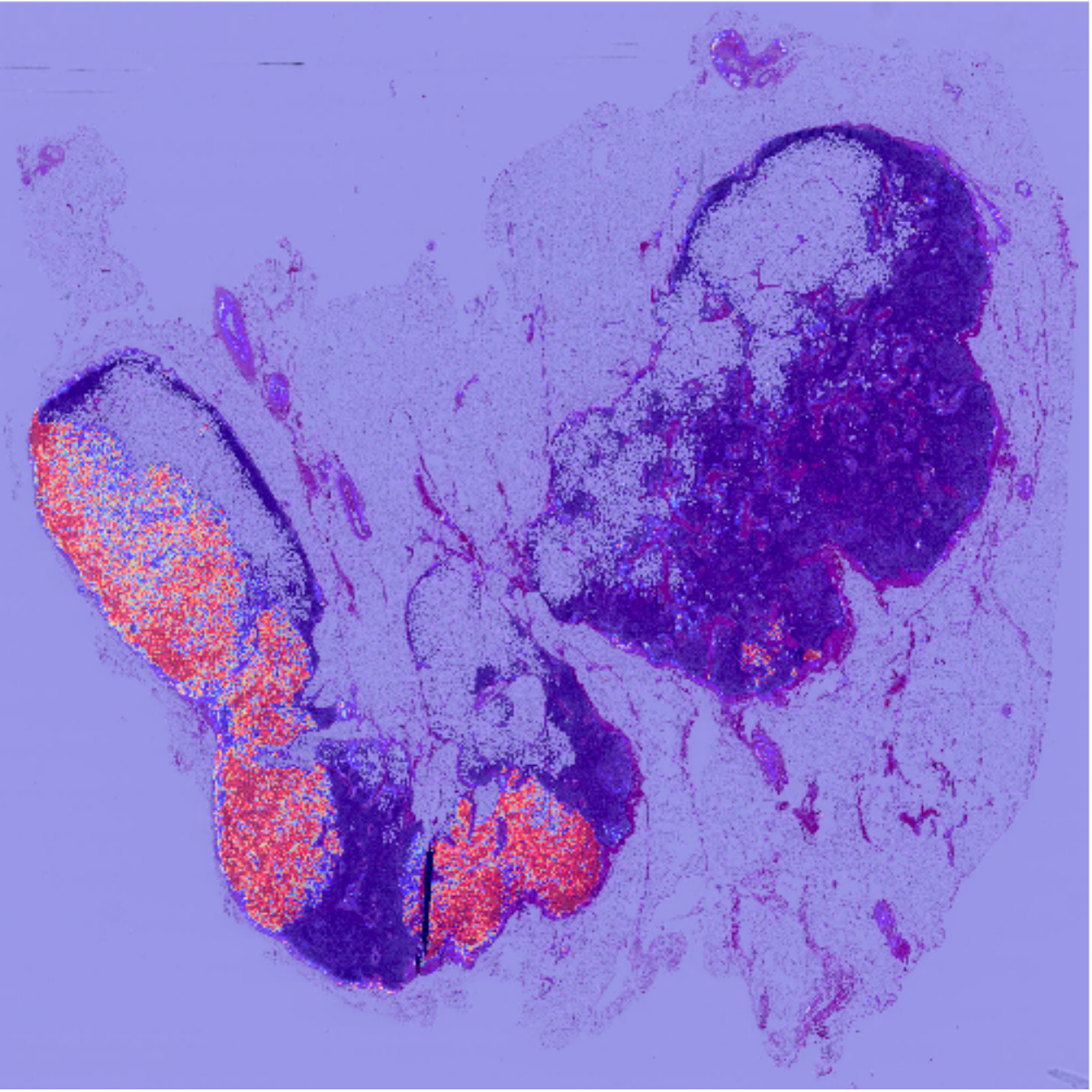}} 
    \end{minipage}
\hfill
	\begin{minipage}{0.095\linewidth}
    \centerline{\includegraphics[width=1.9cm, height=2cm]{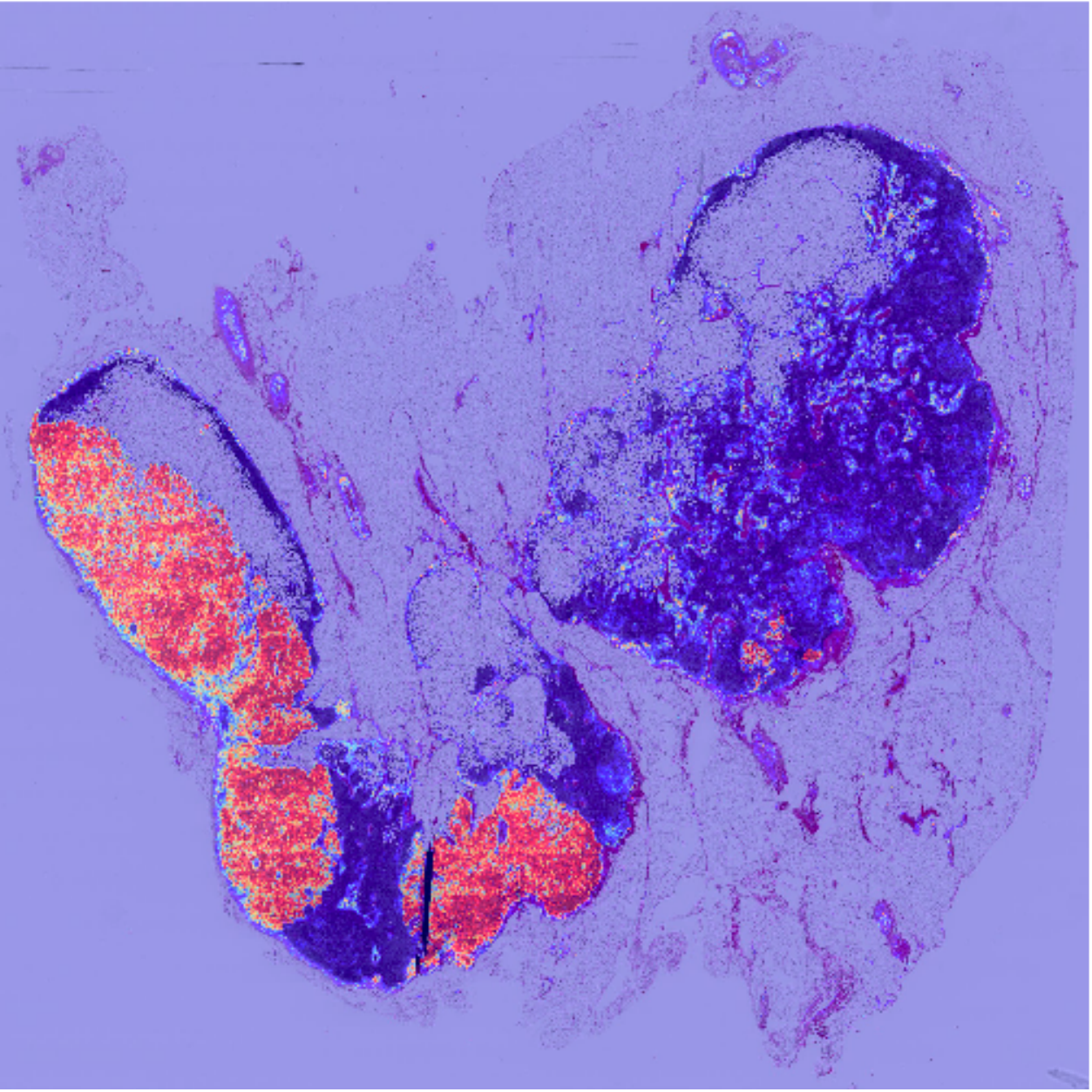}} 
    \end{minipage}
\hfill
	\begin{minipage}{0.095\linewidth}
    \centerline{\includegraphics[width=1.9cm, height=2cm]{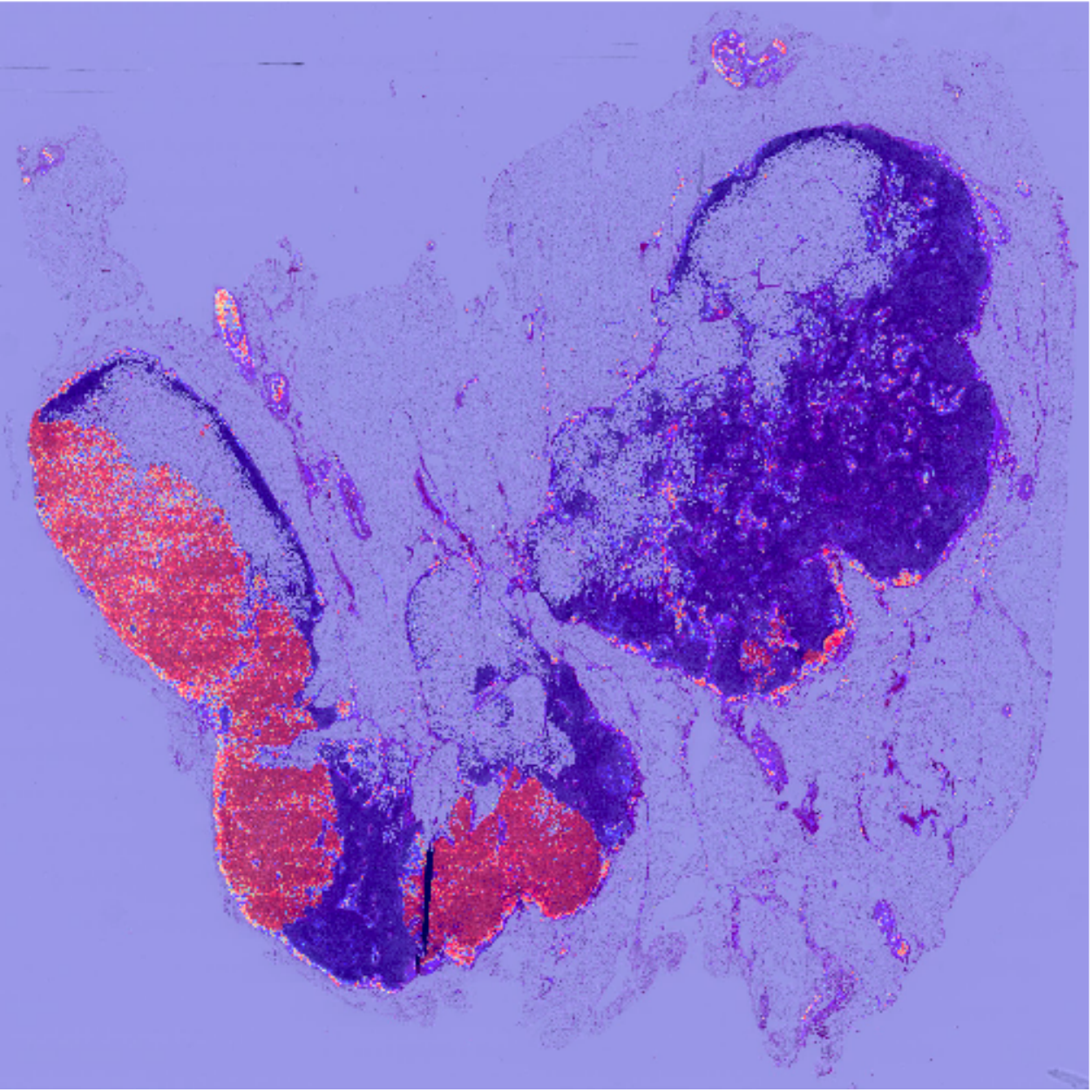}} 
    \end{minipage}
\hfill
	\begin{minipage}{0.095\linewidth}
    \centerline{\includegraphics[width=1.9cm, height=2cm]{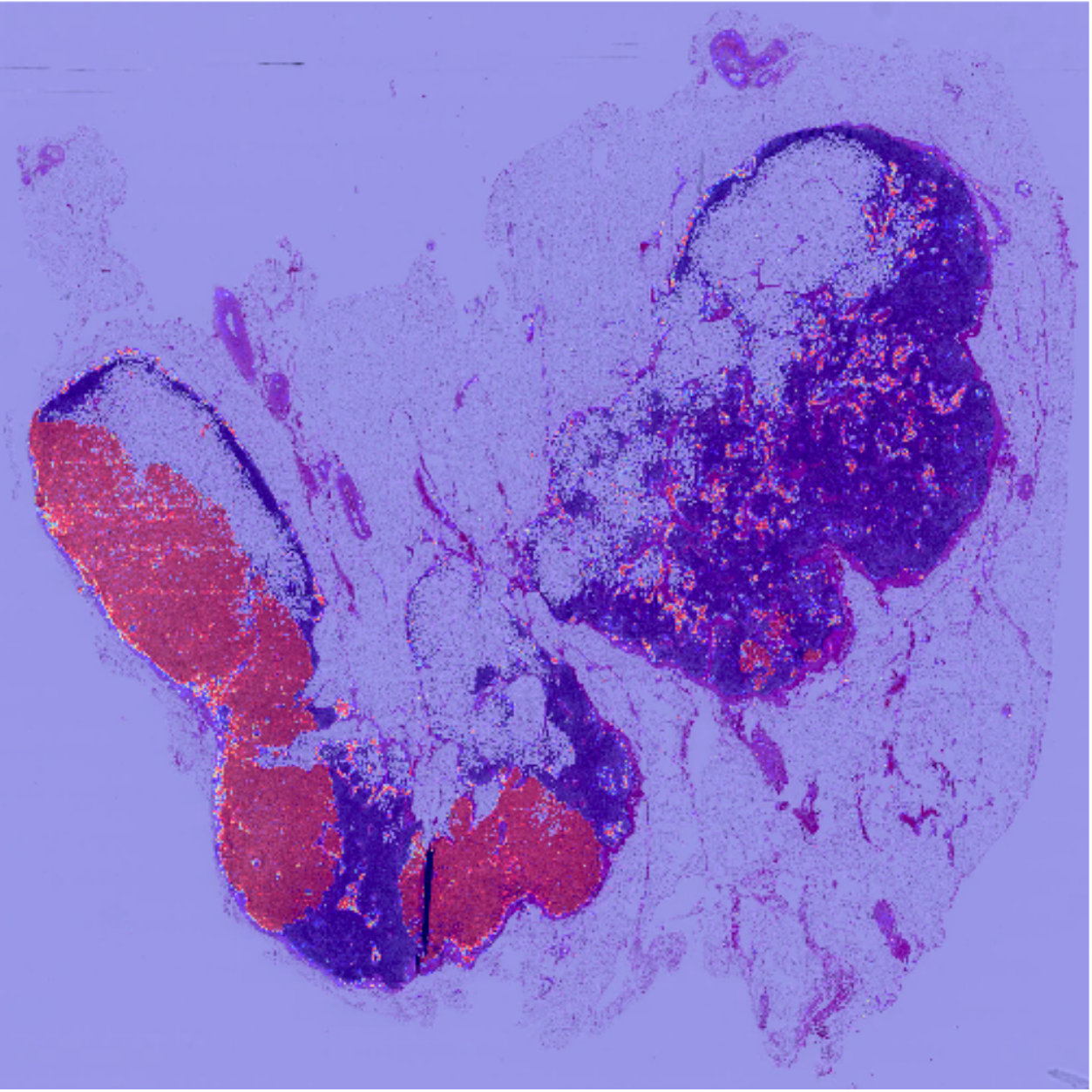}} 
    \end{minipage}    
\hfill
	\begin{minipage}{0.095\linewidth}
    \centerline{\includegraphics[width=1.9cm, height=2cm]{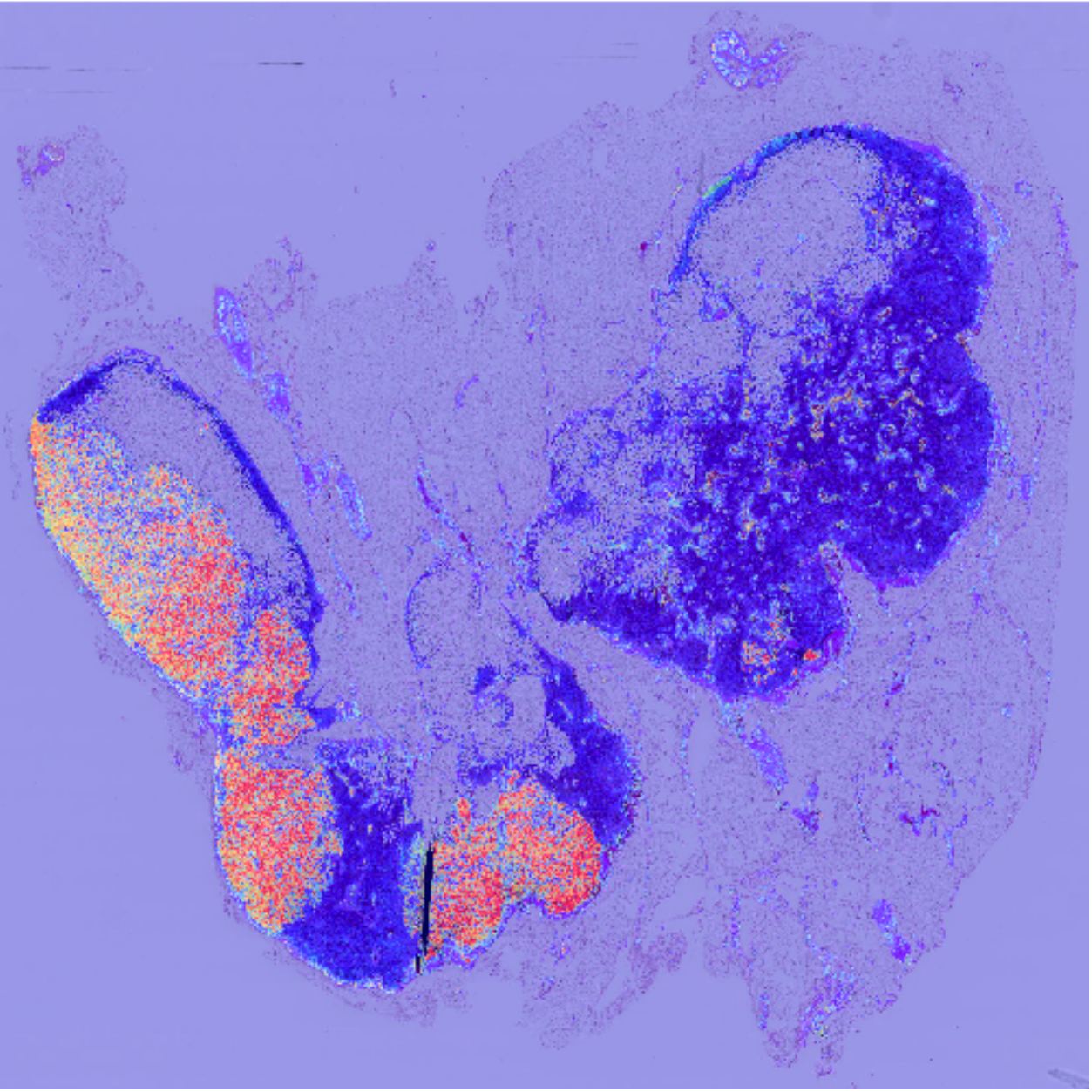}} 
    \end{minipage}
\hfill
	\begin{minipage}{0.095\linewidth}
    \centerline{\includegraphics[width=1.9cm, height=2cm]{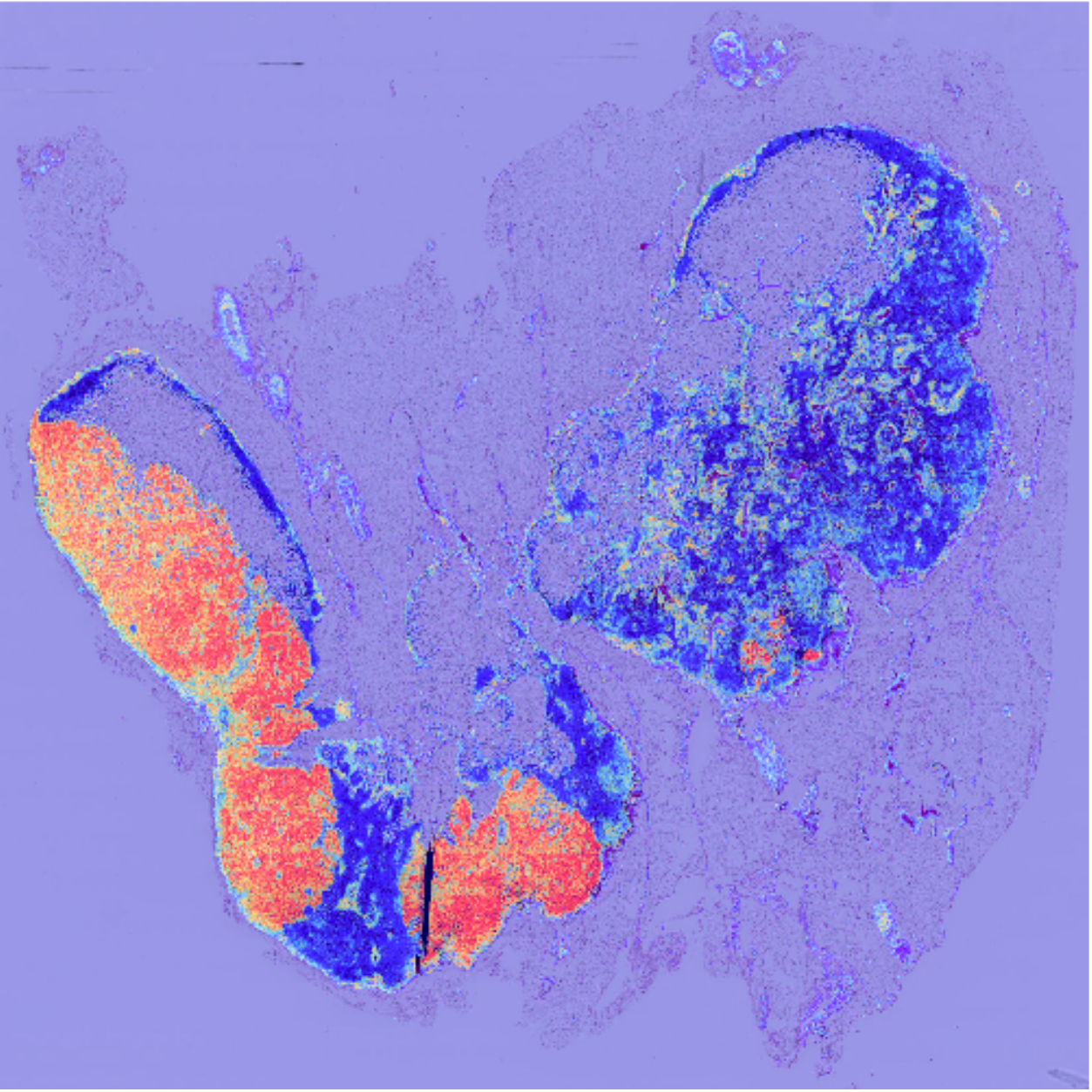}} 
    \end{minipage}
\hfill
	\begin{minipage}{0.095\linewidth}
    \centerline{\includegraphics[width=1.9cm, height=2cm]{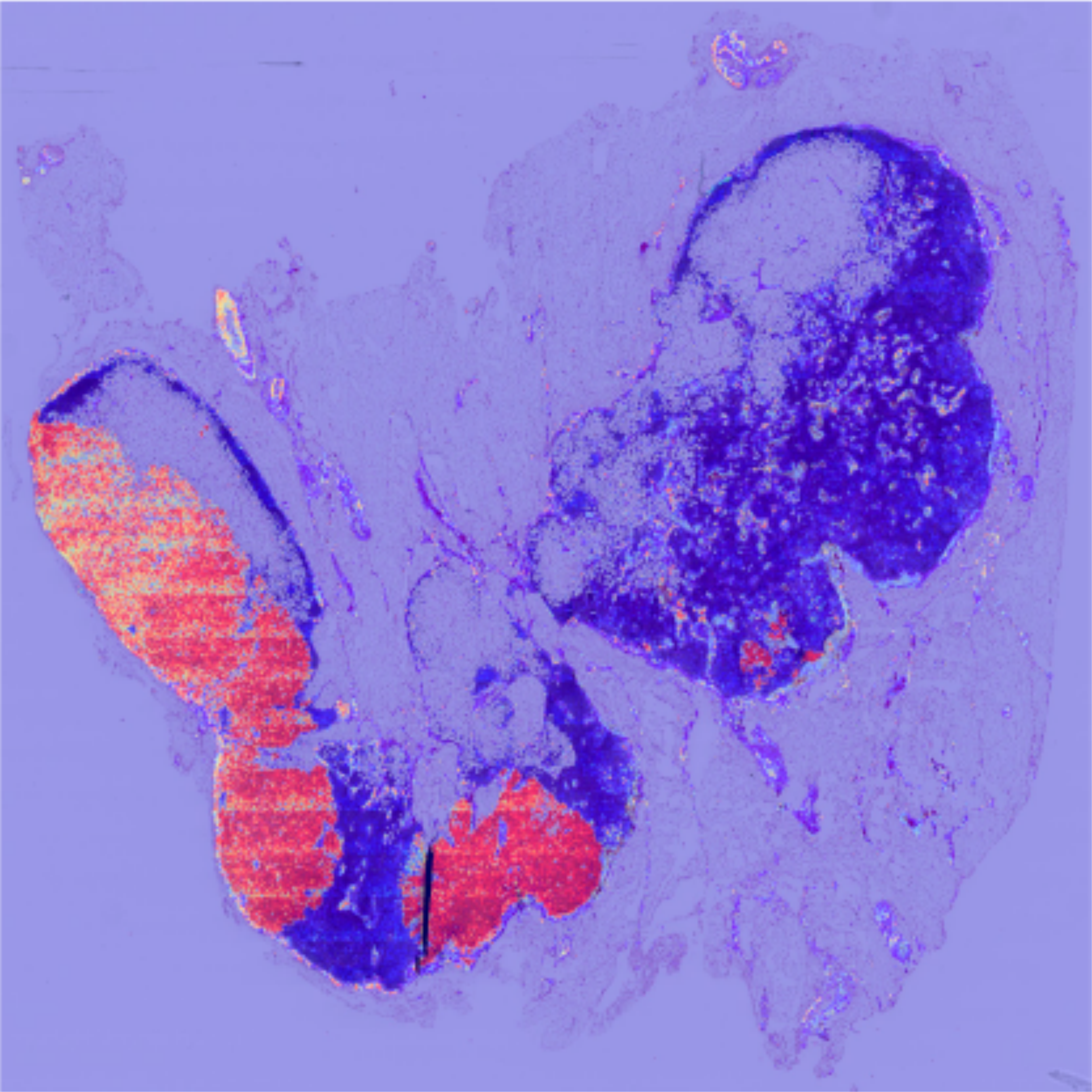}} 
    \end{minipage}
\hfill
	\begin{minipage}{0.095\linewidth}
    \centerline{\includegraphics[width=1.9cm, height=2cm]{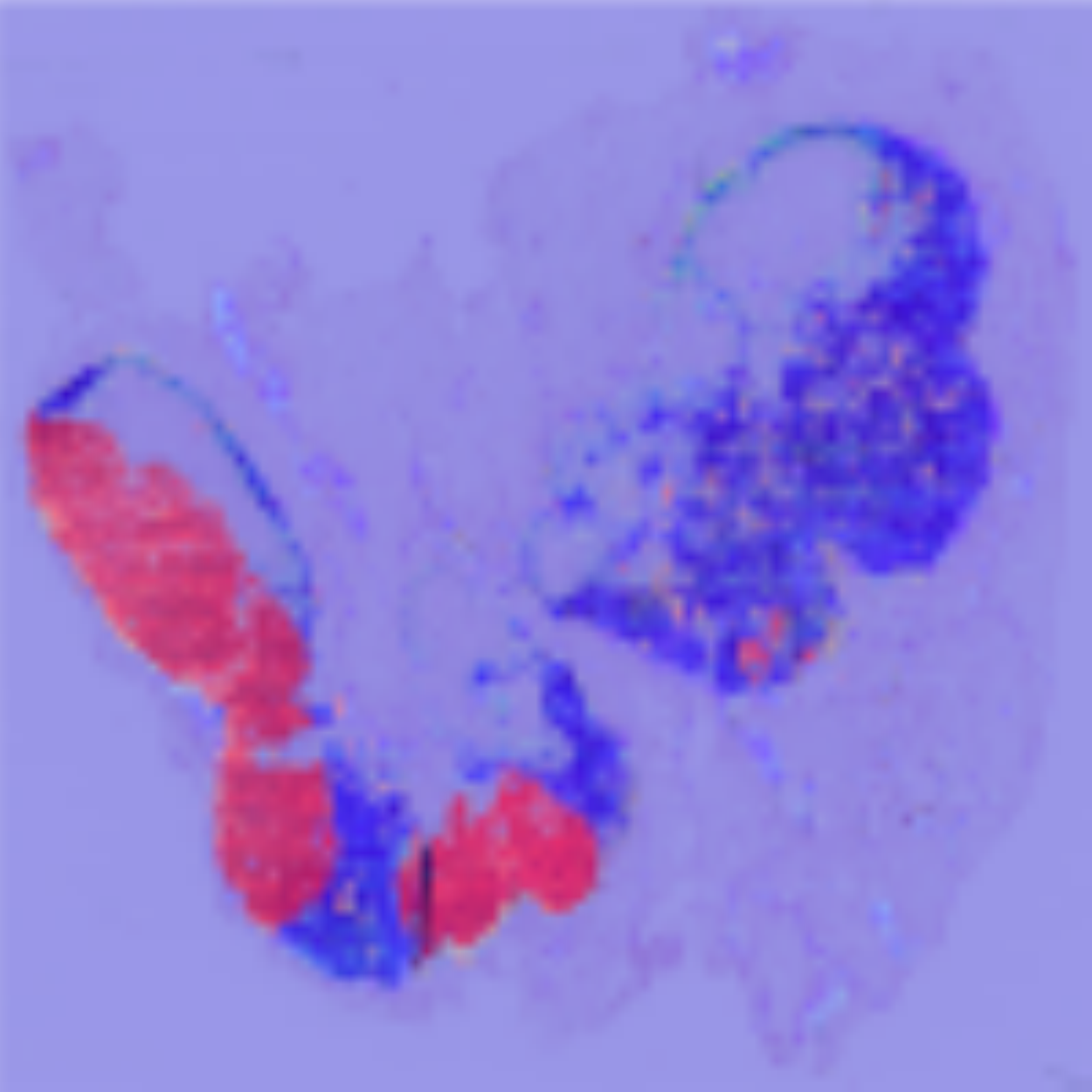}} 
    \end{minipage}    


    
\vfill	

 	\begin{minipage}{0.095\linewidth}
    \centerline{\includegraphics[width=1.9cm, height=2cm]{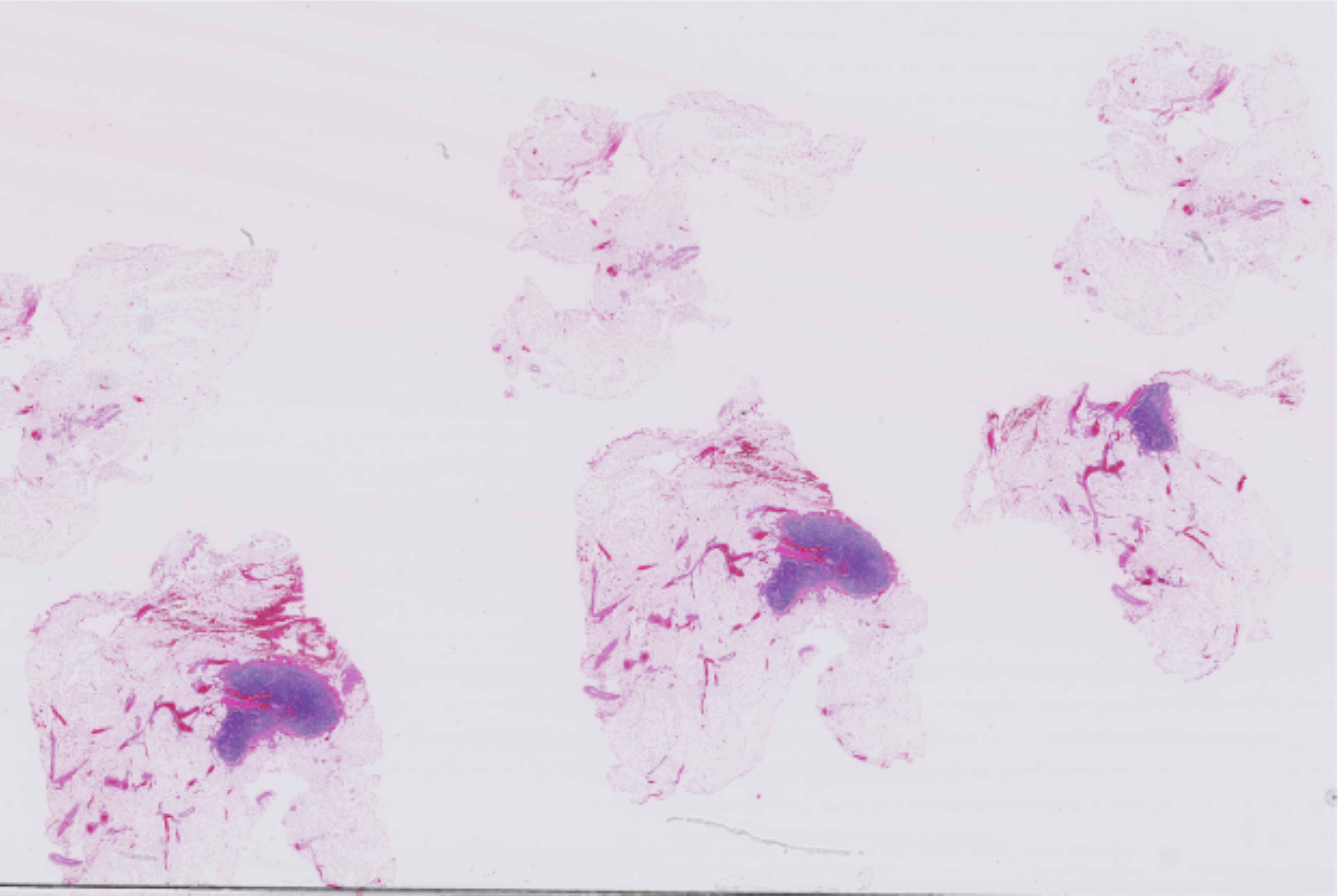}} 
    \centerline{(a)}\medskip
    \end{minipage}
\hfill
	\begin{minipage}{0.095\linewidth}
    \centerline{\includegraphics[width=1.9cm, height=2cm]{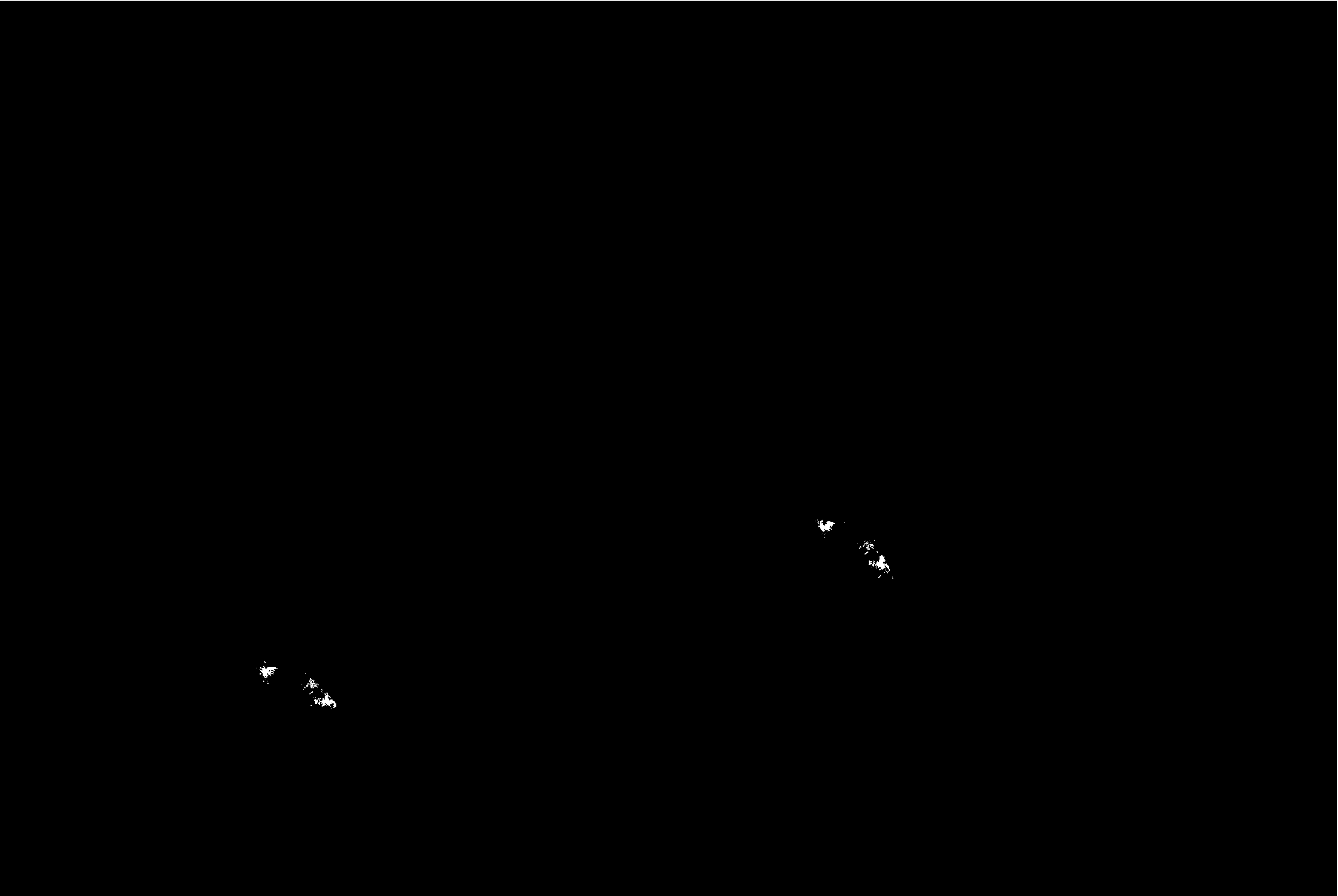}} 
    \centerline{(b)}\medskip
    \end{minipage}    
\hfill
	\begin{minipage}{0.095\linewidth}
    \centerline{\includegraphics[width=1.9cm, height=2cm]{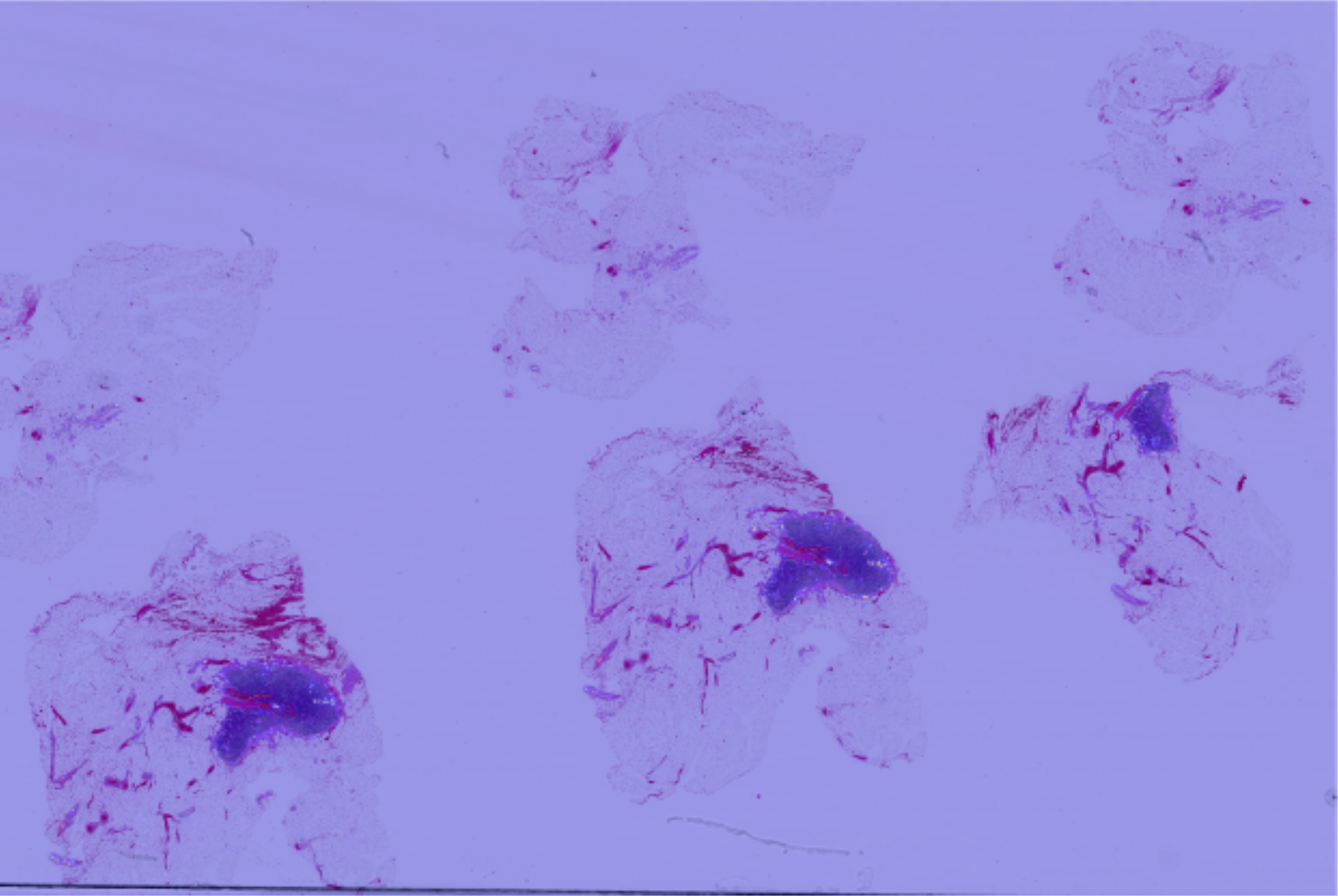}} 
    \centerline{(c)}\medskip
    \end{minipage}
\hfill
	\begin{minipage}{0.095\linewidth}
    \centerline{\includegraphics[width=1.9cm, height=2cm]{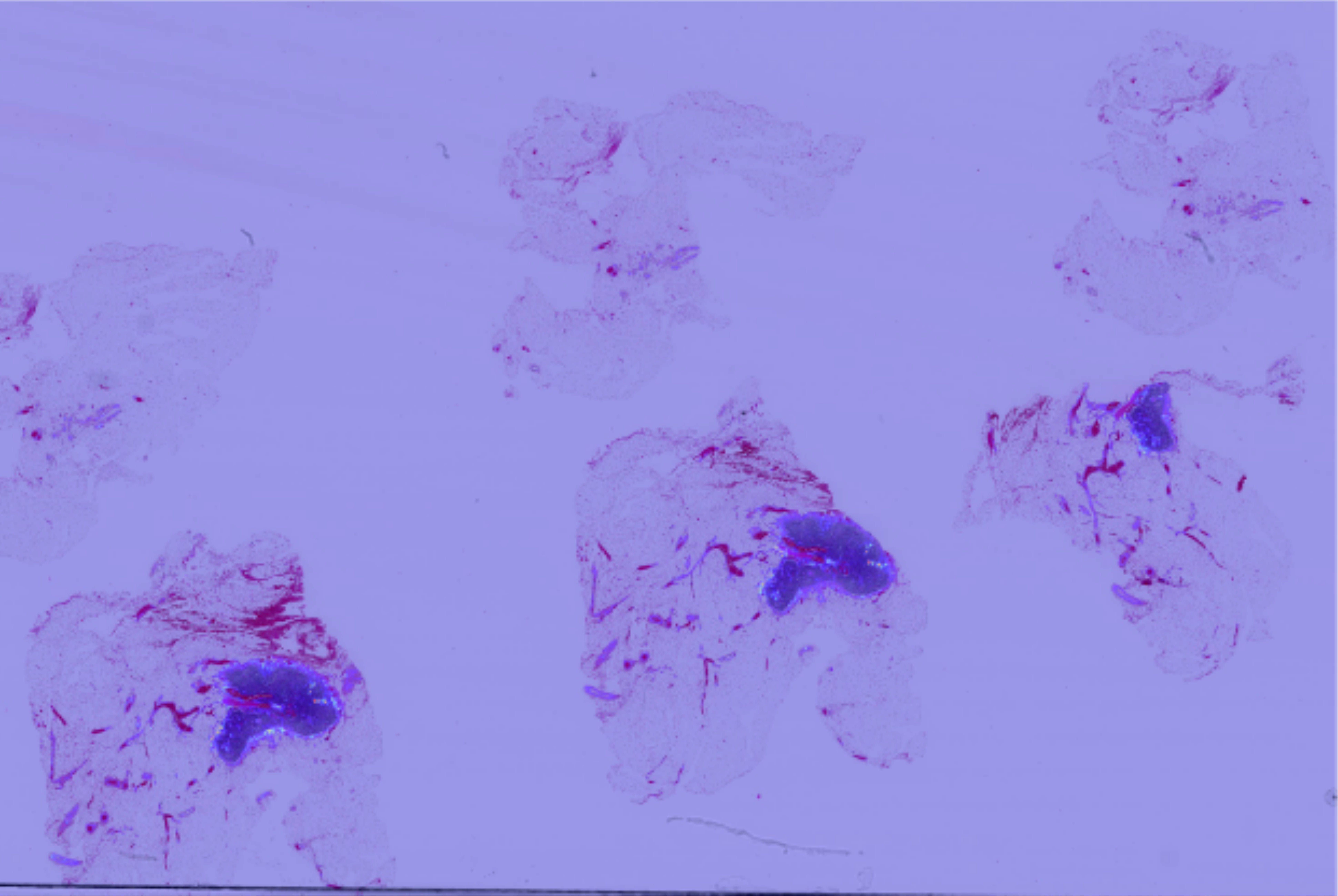}} 
    \centerline{(d)}\medskip
    \end{minipage}
\hfill
	\begin{minipage}{0.095\linewidth}
    \centerline{\includegraphics[width=1.9cm, height=2cm]{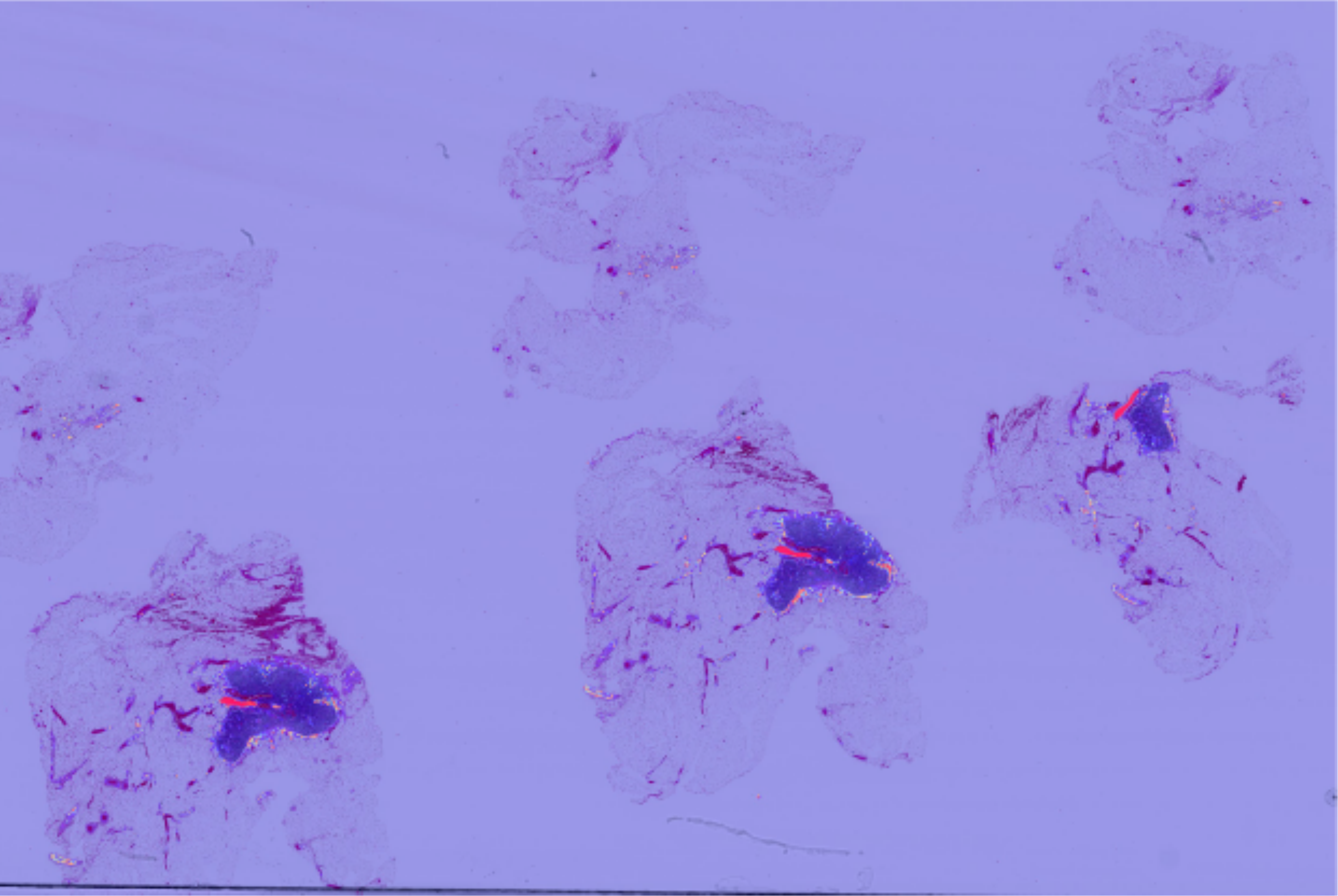}} 
    \centerline{(e)}\medskip
    \end{minipage}
\hfill
	\begin{minipage}{0.095\linewidth}
    \centerline{\includegraphics[width=1.9cm, height=2cm]{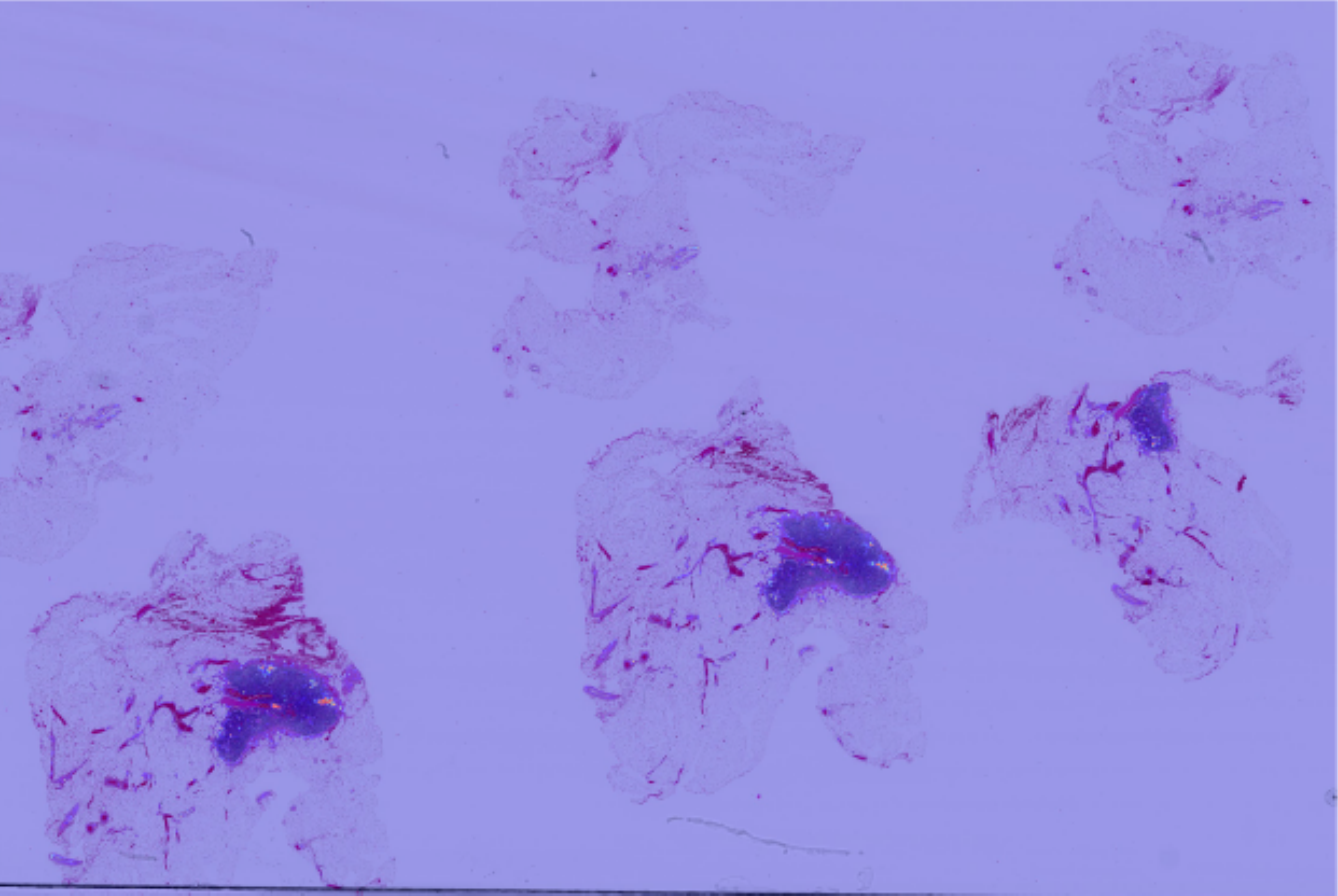}} 
    \centerline{(f)}\medskip
    \end{minipage}    
\hfill
	\begin{minipage}{0.095\linewidth}
    \centerline{\includegraphics[width=1.9cm, height=2cm]{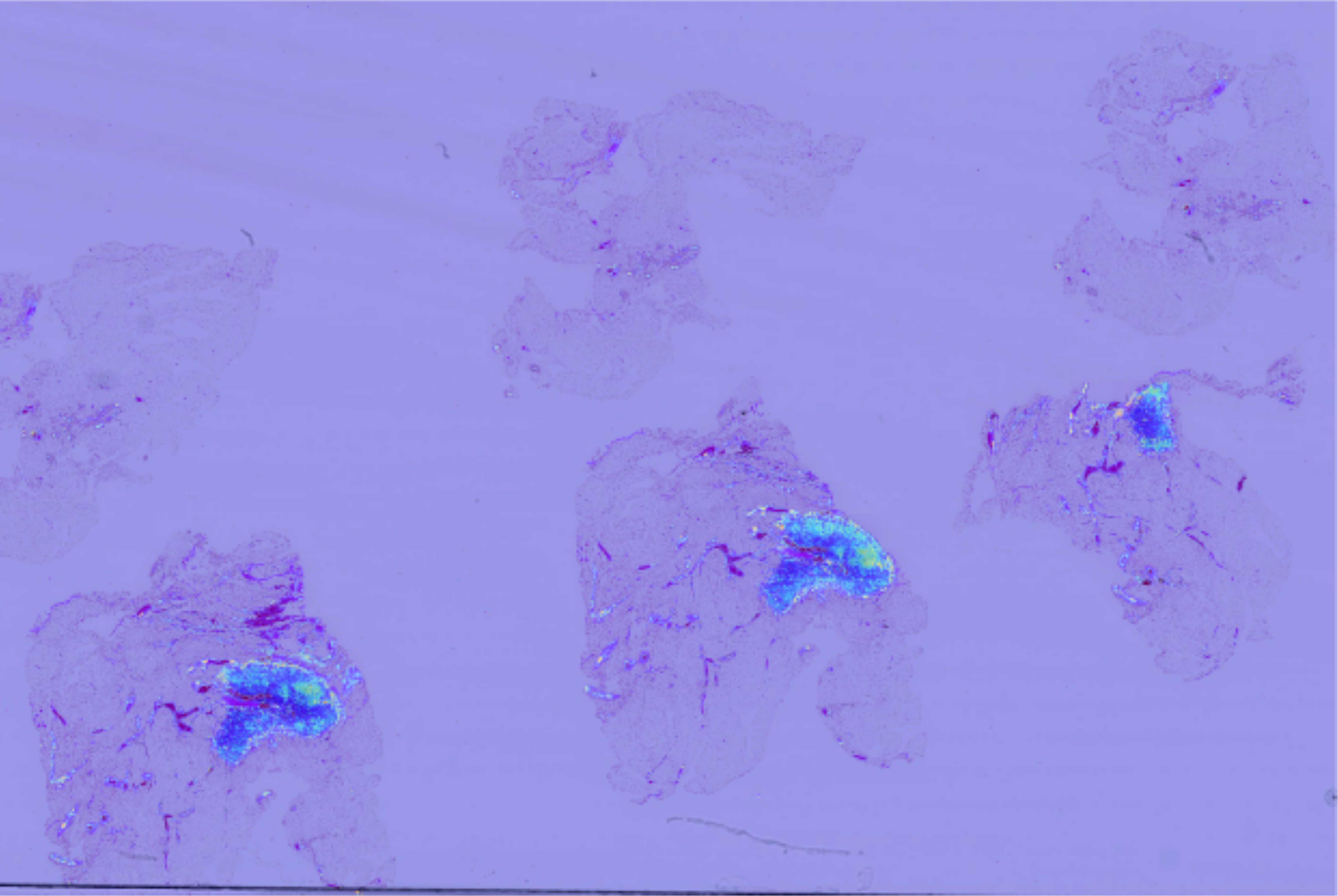}} 
    \centerline{(g)}\medskip
    \end{minipage}
\hfill
	\begin{minipage}{0.095\linewidth}
    \centerline{\includegraphics[width=1.9cm, height=2cm]{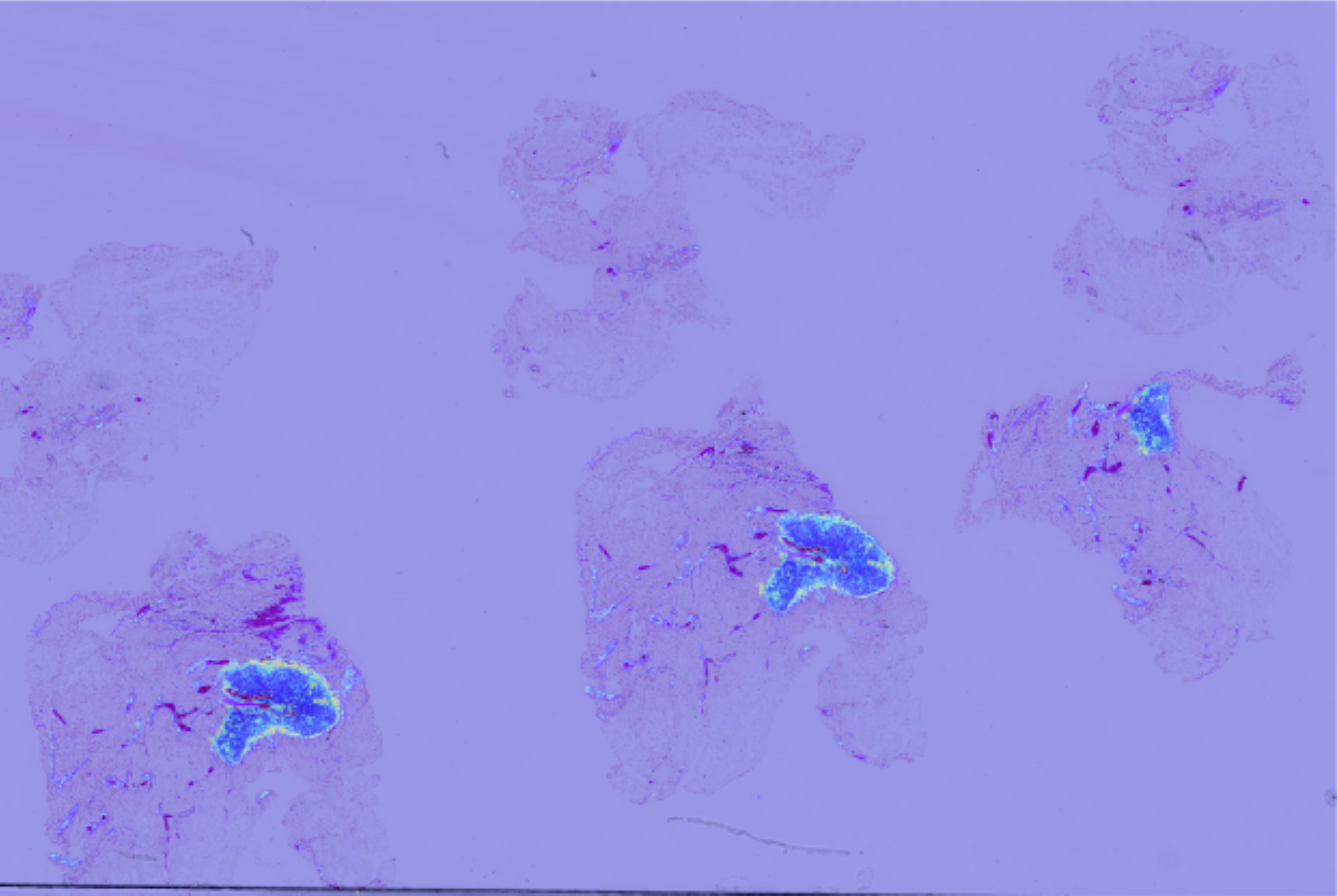}} 
    \centerline{(h)}\medskip
    \end{minipage}
\hfill
	\begin{minipage}{0.095\linewidth}
    \centerline{\includegraphics[width=1.9cm, height=2cm]{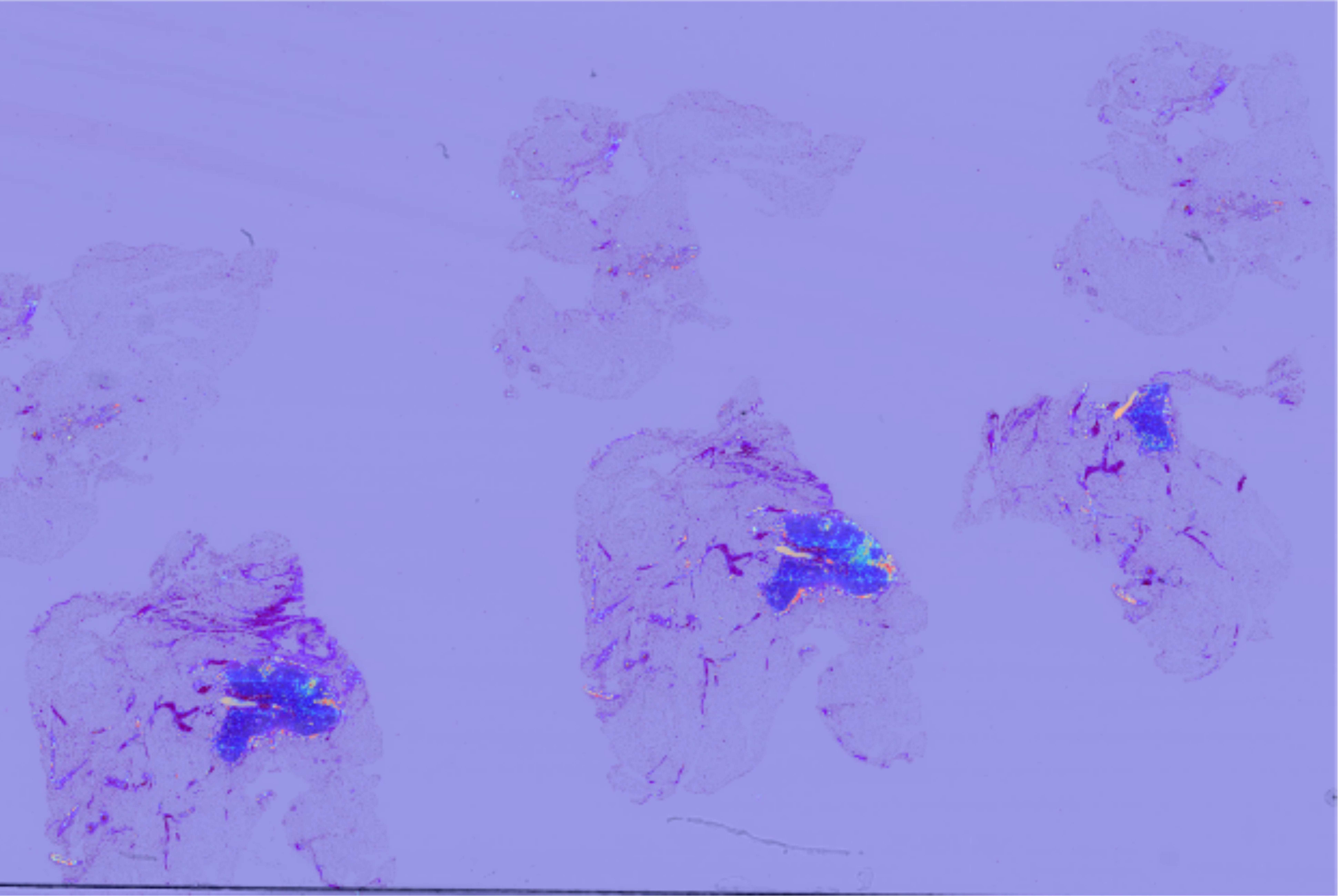}} 
    \centerline{(i)}\medskip
    \end{minipage}
\hfill
	\begin{minipage}{0.095\linewidth}
    \centerline{\includegraphics[width=1.9cm, height=2cm]{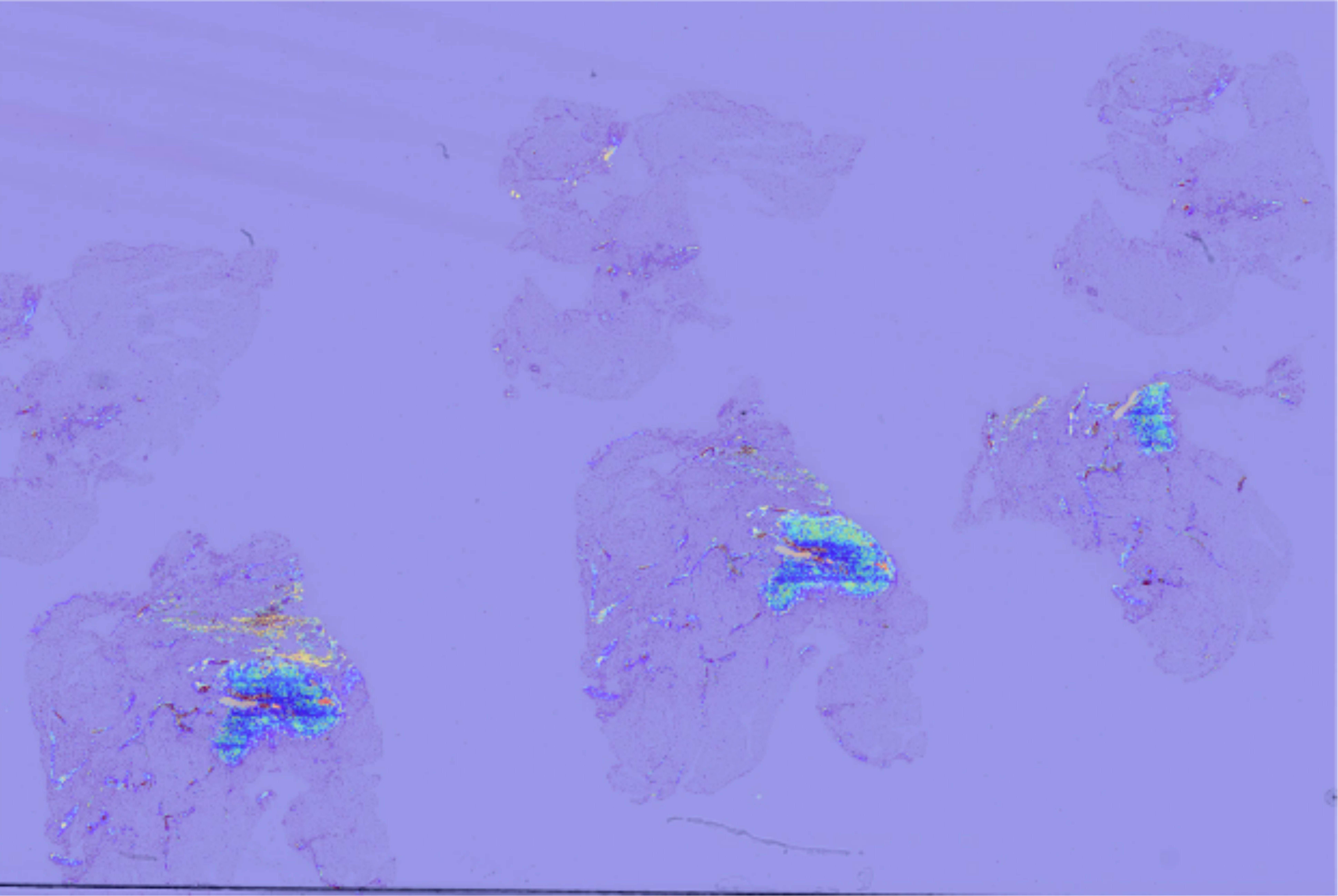}} 
    \centerline{(j)}\medskip
    \end{minipage}   
\hfill
	\begin{minipage}{\linewidth}
    \centerline{\includegraphics[scale=0.60, angle=270,origin=c]{color_bar}} 
    \end{minipage}
\vspace{-2.7cm}    
\caption{Tumor probability heat-maps overlaid on original WSIs from Camelyon16 test set predicted from 10\% labeled data. (a) Original WSI; (b) Ground truth annotation mask; (c) -- (f) corresponds to tumor probability produced by random (supervised), VAE, MoCo, and RSP approach, respectively; (g) -- (i) corresponds to tumor probability produced by random+CR (supervised), VAE+CR, MoCo+CR, and RSP+CR methods, respectively. The first three rows correspond to examples of macro-metastases (tumor cell cluster diameter $\geq 2 mm$), while the last row corresponds to micro-metastases (tumor cell cluster diameter from $>0.2mm $ to $<2mm$). The color blue denotes healthy regions, and red denotes tumor regions.}
\label{Fig:Heat-maps on Camelyon16}
\end{figure}
\end{landscape}
}
\bibliographystyle{model2-names.bst}\biboptions{authoryear}  
\bibliography{refs}

\end{document}